\documentclass[preprint,10pt,authoryear]{elsarticle}

\usepackage{graphicx}
\usepackage{url}
\usepackage{enumitem}
\usepackage{amsfonts}
\usepackage{amsmath}
\usepackage{amssymb}
\usepackage{booktabs}
\usepackage{enumitem}
\usepackage{boldline}
\usepackage{threeparttable}
\usepackage{array}
\usepackage{float}
\usepackage{textcomp}
\usepackage{mdwmath}
\usepackage{mdwtab}
\usepackage{subfig}
\usepackage{amsmath}
\usepackage{color}
\usepackage{tikz}
\usepackage{multirow}
\usepackage{bbding}
\usepackage{caption}
\usepackage{bm}
\usepackage{amsthm}
\usepackage{dblfloatfix}
\usepackage{hyperref}
\usepackage{algorithm,algpseudocode}
\usepackage{lipsum}
\usepackage{etoolbox}
\usepackage{lineno}
\usepackage{tikz}
\usepackage[margin=2cm]{geometry}

\definecolor{navy}{RGB}{0, 0, 128}

\makeatletter
\newenvironment{breakablealgorithm}
  {
   \begin{center}
     \refstepcounter{algorithm}
     \hrule height.8pt depth0pt \kern2pt
     \renewcommand{\caption}[2][\relax]{
       {\raggedright\textbf{\ALG@name~\thealgorithm:} ##2\par}%
       \ifx\relax##1\relax 
         \addcontentsline{loa}{algorithm}{\protect\numberline{\thealgorithm}##2}%
       \else 
         \addcontentsline{loa}{algorithm}{\protect\numberline{\thealgorithm}##1}%
       \fi
       \kern2pt\hrule\kern2pt
     }
  }{
     \kern2pt\hrule\relax
   \end{center}
  }
\makeatother

\algnewcommand\algorithmicinput{\textbf{Input:}}
\algnewcommand\Input{\item[\algorithmicinput]}
\algnewcommand\algorithmicoutput{\textbf{Output:}}
\algnewcommand\Output{\item[\algorithmicoutput]}
\algnewcommand\algorithmicParameter{\textbf{Parameters:}}
\algnewcommand\Parameters{\item[\algorithmicParameter]}
\newcommand{\Break}{\State \textbf{break} }
\newcommand{\Continue}{\State \textbf{continue} }

\journal{Engineering Applications of Artificial Intelligence}

\biboptions{authoryear}

\begin{document}

\begin{frontmatter}

\captionsetup[figure]{labelfont={bf},labelformat={default},labelsep=period,name={Fig.}}
\captionsetup[table]{labelfont={bf},labelformat={default},labelsep=period,name={Table}}

\title{Human-Centric Traffic Signal Control for Equity: A Multi-Agent Action Branching Deep Reinforcement Learning Approach}


\author[mymainaddress]{Xiaocai Zhang\corref{mycorrespondingauthor}}
\ead{xiaocai.zhang@unimelb.edu.au}

\author[mymainaddress]{Neema Nassir}
\ead{neema.nassir@unimelb.edu.au}

\author[mymainaddress]{Lok Sang Chan}
\ead{loksangc@student.unimelb.edu.au}

\author[mymainaddress]{Milad Haghani}
\ead{milad.haghani@unimelb.edu.au}

\address[mymainaddress]{Department of Infrastructure Engineering, Faculty of Engineering and Information Technology, The University of Melbourne, VIC 3010, Australia}

\begin{abstract}
Coordinating traffic signals along multimodal corridors is challenging because many multi-agent deep reinforcement learning (DRL) approaches remain vehicle-centric and struggle with high-dimensional discrete action spaces. We propose MA2B-DDQN, a human-centric multi-agent action-branching double Deep Q-Network (DQN) framework that explicitly optimizes traveler-level equity. Our key contribution is an action-branching discrete control formulation that decomposes corridor control into (i) local, per-intersection actions that allocate green time between the next two phases and (ii) a single global action that selects the total duration of those phases. This decomposition enables scalable coordination under discrete control while reducing the effective complexity of joint decision-making. We also design a human-centric reward that penalizes the number of delayed individuals in the corridor, accounting for pedestrians, vehicle occupants, and transit passengers. Extensive evaluations across seven realistic traffic scenarios in Melbourne, Australia, demonstrate that our approach significantly reduces the number of impacted travelers, outperforming existing DRL and baseline methods. Experiments confirm the robustness of our model, showing minimal variance across diverse settings. This framework not only advocates for a fairer traffic signal system but also provides a scalable solution adaptable to varied urban traffic conditions.
\end{abstract}


\end{frontmatter}

\section{Introduction}
Driven by population growth and urbanization, the demand for efficient and diverse transportation options is steadily increasing in many cities around the world \citep{chu2019multi}. This leads to the development of multimodal transport systems, which integrate various modes of transport, such as walking, cycling, public transit, and private vehicles, to provide seamless, accessible and sustainable mobility solutions.
Fairness and equity are important principles of a multimodal transportation system, which encompasses all users of transportation, not just those in cars. By shifting the focus from a car-centric model to an individual-centric solution, this system promotes sustainability while ensuring fair access to transportation resources for all road users, regardless of their mode of travel.
Despite the importance of fairness, only a few works have translated fairness into the urban transportation sector, with a focus on transportation planning rather than traffic engineering. The concept of fairness has received limited attention from scholars and practitioners alike in the context of traffic engineering \citep{riehl2024towards}. Ensuring fairness in transportation systems means making them accessible to all segments of the community, regardless of economic status, disability, or geographic location. This is important for reducing social inequalities and promoting fair access to public services, ultimately leading to more accessible and sustainable transportation.

Traffic signals coordinate movements at intersections to enhance the efficiency of transportation networks. Traffic Signal Control (TSC) remains a challenging but active research problem within transportation due to its complex, dynamic, and stochastic characteristics \citep{wei2019presslight,chan2026multi}. Traditional ﬁxed-time traffic light control algorithms become inefficient in managing congestion within highly dynamic trafﬁc scenarios. Although manual traffic control can be effective, it often results in inefficient use of time and manpower. Early-stage adaptive TSC methods tackle optimization challenges to develop efficient coordination and control strategies. Popular TSC solutions, including Split Cycle Offset Optimization Technique (SCOOT) \citep{hunt1982scoot} and Sydney Coordinated Adaptive Traffic System (SCATS) \citep{stevanovic2009scoot}, have been implemented in many cities across the world \citep{chu2019multi}. However, these adaptive TSC systems, like SCATS, mainly address vehicle traffic and often require further enhancements to accommodate pedestrians or cyclists. For SCATS to efficiently cater to integrated transportation systems, more innovative adaptations are necessary \citep{mccannfeet}.

In order to elevate the efficiency of TSC, the paradigm of Deep Reinforcement Learning (DRL) has been extensively leveraged to handle the complex and fluctuating environmental dynamics in traffic and transportation systems. DRL is particularly effective in TSC due to its ability to autonomously discover optimal strategies through a process of trial and error, effectively adapt to evolving conditions, and make informed decisions grounded in intricate and variable traffic patterns. This approach has been shown to surpass conventional transportation approaches in managing traffic flows, as evidenced by several recent studies \citep{wei2018intellilight,yazdani2023intelligent}. These studies highlight the substantial benefits of DRL, showcasing its potential to revolutionize traffic management by enhancing decision-making processes and response times to real-time traffic situations.

In its early stages, the applications of DRL in TSC were primarily concentrated on individual intersections. Subsequently, techniques like Q-learning \citep{prabuchandran2014multi}, DQN \citep{ge2019cooperative}, DDDQN \citep{gong2019decentralized}, and hybrid methods that integrate fuzzy control with DRL \citep{kumar2020fuzzy} were gradually extended to optimize the signal control of multiple intersections. Effectively coordinating traffic signals is essential to ensure that all traffic lights within a network collaborate to enhance overall traffic flow efficiency. This paper aims to develop advanced DRL methodologies for TSC at multiple intersections within the same corridor. However, previous research on DRL for multi-intersection signal coordination primarily targets enhancing vehicle throughput and reducing waiting time or congestion at intersections. This has prompted a need for exploring more humanized objectives, such as equity and fairness, within DRL frameworks for traffic and social management.
Moreover, effectively managing both vehicle and pedestrian traffic is challenging due to the contradiction in their optimization scale required: pedestrian traffic needs focused optimization at individual intersections, while vehicle traffic optimization needs to be extended at the corridor-wide level. This paper tackles this challenge by exploring multimodal transportation coordination that encompasses both vehicles and pedestrians.

\begin{figure}[htbp]
\begin{center}
\includegraphics[width=0.86\textwidth]{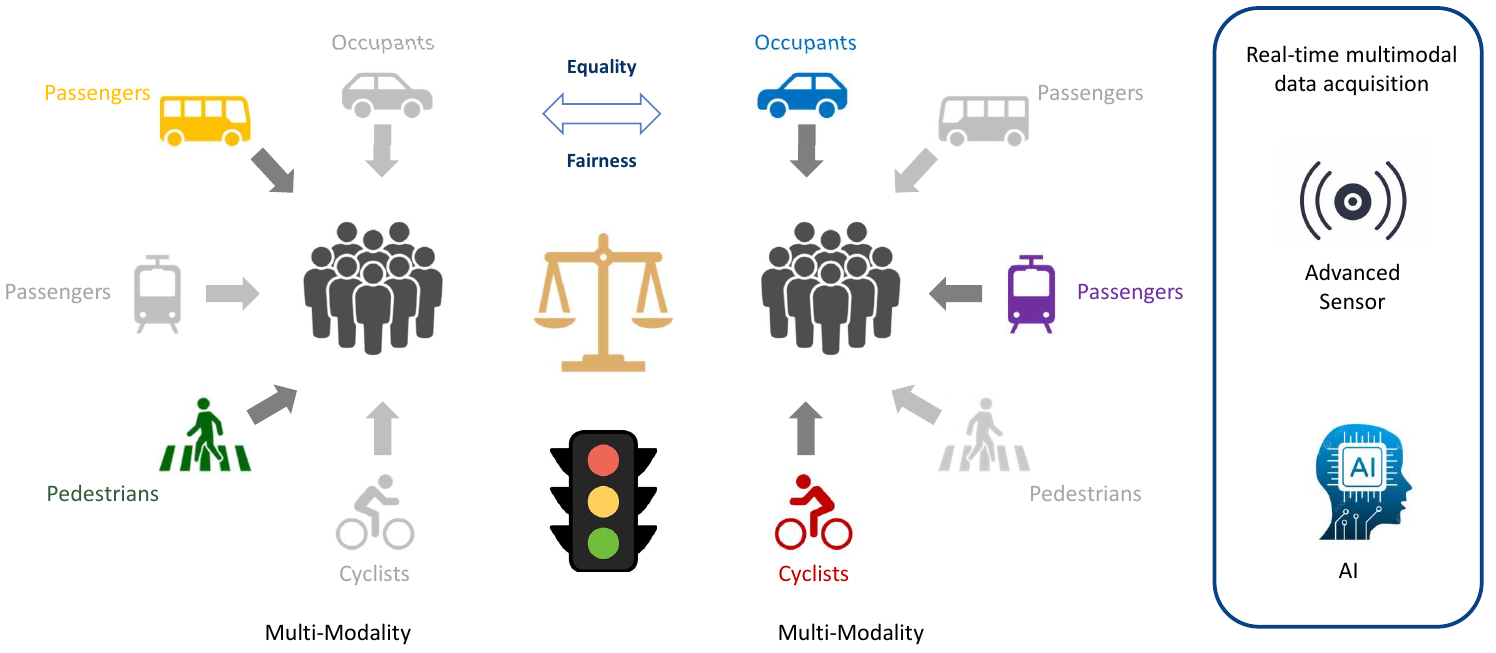}
\caption{Conceptual overview of human-centric multimodal traffic signal control: travelers include private-vehicle occupants, public-transit passengers, cyclists, and pedestrians, with advanced sensing/AI enabling real-time multimodal user data acquisition and pre-processing.}
\label{tsc_source}
\end{center}
\end{figure}

In this paper, we introduce a novel DRL framework tailored for multi-agent environments, aimed at optimizing traffic signals across a network of multiple intersections. This cooperative learning model uses a periodic synchronization strategy, where all agents periodically update their policies based on globally shared information. The updating interval is also an adaptive variable, determined by the DRL model to promote exploration of actions and states for potentially higher rewards.
The method is fundamentally built on the Double Deep-Q Network (DDQN) framework, incorporating an action branch architecture \citep{tavakoli2018action} extending for multi-agent optimization.
The model is termed as \textbf{M}ulti-\textbf{A}gent \textbf{A}ction-\textbf{B}ranching \textbf{DDQN} (\textbf{MA2B-DDQN}).
This approach can be extended to a large-scale network with more intersections (agents) included. The proposed method aims to find the optimal timing of the traffic signals for the upcoming few phases at all intersections within this road network, by treating every participant in the transportation network equally and fairly. As shown by Figure \ref{tsc_source}, users of the transportation services can come from various modalities, such as private cars, pedestrians, cyclists, and public transit options like buses or trams. The information such as the numbers of users for different modalities of transportation is useful for human-centric optimization. 
On the other hand, the development of advanced sensor and Artificial Intelligence (AI) technologies has facilitated the process of information sensing, data collection, data fusion \citep{zhang2023prediction}, and data pre-processing \citep{liang2024survey}. For example, the number of occupants in a car can be estimated \citep{jiang2023pa,papakis2021convolutional}, the number of pedestrians can be counted using Bluetooth low energy scanners \citep{gong2022using}, and on-board passenger counts for public transportation could be precisely predicted using advanced AI and deep learning models \citep{roncoli2023estimating,zhang2024dynamic}.

The main \textbf{contributions} of this study are summarized as follows:
\begin{enumerate}
\item We introduced the concept of forward-looking human-centric TSC at multiple intersections considering fairness and equality by treating every participant in the transport network equally;
\item We presented a cooperative multi-agent DRL framework that adaptively determines a global action using globally shared information, while simultaneously optimizing traffic signal timings through local actions taken by individual agents;
\item We first proposed the action branching architecture for multi-agent DRL in modeling the TSC problem to handle the issues brought by discrete action space and multi-agent setting.
\end{enumerate}

The structure of the subsequent sections of this paper is organized as follows: Section \ref{relatedwork} provides an in-depth review of existing research on multi-agent DRL approaches for TSC. Section \ref{method} details the proposed MA2B-DDQN framework, including a preliminary overview and crucial elements of DRL such as the environment and state definition, action space, reward mechanism, network architecture, and learning algorithm. Section \ref{experiment} discusses the experimental results and analyses. Finally, Section \ref{conclusion} concludes the study and outlines directions for future research.

\section{Related work}
\label{relatedwork}
We conducted an extensive literature review of 44 recent peer-reviewed papers on multi-agent DRL in TSC across multiple intersections. These papers were published in prestigious journals and conferences, including Transportation Research Part C: Emerging Technologies, IEEE Transactions on Intelligent Transportation Systems, Neurocomputing, the ACM SIGKDD Conference on Knowledge Discovery and Data Mining, and the International Joint Conference on Artificial Intelligence (IJCAI). In the following sections, we review existing TSC methods that utilize DRL, examining them from three perspectives: traffic simulation environment, reward mechanism, and multi-agent DRL framework.

\subsection{Traffic simulation environment}
Microscopic traffic simulation is widely used to replicate realistic traffic environments for multi-agent traffic signal control research based on DRL. Five popular simulation tools are commonly employed in this field: SUMO\footnote[1]{\url{https://eclipse.dev/sumo/}}, CityFlow\footnote[2]{\url{https://cityflow-project.github.io/}}, VISSIM\footnote[3]{\url{https://www.ptvgroup.com/en/products/ptv-vissim}}, Aimsun\footnote[4]{\url{https://www.aimsun.com/}}, and Paramics\footnote[5]{\url{https://www.systra.com/digital/solutions/transport-planning/paramics/}}. Table \ref{table_simulation} lists the traffic simulation tools used by each study. The majority of existing studies, around 56.82\% (25 out of 44), use SUMO for environment simulation, probably due to its free availability, which is advantageous for budget-limited projects. CityFlow ranks as the second most popular choice, utilized in 10 studies. PTV VISSIM is third, used in 5 studies, Aimsun has been applied in 2 studies for multi-agent traffic signal control. While Paramics has been used in only one study. SUMO and CityFlow are freely available, while VISSIM, Aimsun, and Paramics are commercial software, which can be costly. While SUMO is functional, its visualization is less advanced compared to commercial software. CityFlow is specifically designed for reinforcement learning applications in traffic signal control, making it less versatile than other simulators. Paramics offers detailed microscopic simulation and real-time traffic management capabilities, making it particularly suitable for testing adaptive signal control and ITS applications. VISSIM excels in modeling complex vehicle interactions and mixed traffic, whereas Aimsun supports both microscopic and mesoscopic simulations and can handle large networks. Additionally, VISSIM, Aimsun, and Paramics can be integrated with other software and databases, enhancing their functionality.

\begin{table}[htbp]
\centering
\caption{Summary of traffic simulation environments}
\label{table_simulation}
\begin{threeparttable}
\begin{tabular}{p{3.0cm}p{9.5cm}p{2.0cm}}
\hline
Traffic Simulation & References & Quantity \\
\hline
SUMO & \cite{wang2021adaptive,li2021network,su2023emvlight,song2024cooperative,wang2024large,bie2024multi,yang2023hierarchical,yang2021ihg,ren2024two,xu2024graph,zhang2022distributed,hu2024multi,tan2020multi,chu2019multi,wu2020multi1,bokade2023multi,ge2021multi,luo2024reinforcement,liu2023traffic,nie2025cmrm,jia2025multi,fereidooni2025multi,shen2025hierarchical,ma2020feudal,tao2023network} & 25\\
\hline
CityFlow & \cite{zhu2023multi,liu2023multiple,wu2022distributed,liu2023gplight,zhang2022neighborhood,wang2020stmarl,liu2025globallight,lai2025llmlight,satheesh2025constrained,wang2025towards} & 10\\
\hline
VISSIM & \cite{jiang2021distributed,wang2021gan,wu2020multi2,lee2019reinforcement,zhang2025towards} & 5\\
\hline
Aimsun & \cite{abdoos2020cooperative,yoon2025decentralized} & 2\\
\hline
Paramics & \cite{el2013multiagent} & 1\\
\hline
NM & \cite{wang2020large} & 1\\
\hline
\end{tabular}
        \textit{Notes:} "\textit{NM}" means "not mentioned in the paper".
\end{threeparttable}
\end{table}

\subsection{Reward mechanism}
The reward mechanism is a key component in developing DRL models, as it guides the learning process of an agent by providing feedback during interactions with its environment. Most of the studies reviewed primarily define rewards from the perspective of the vehicle, with only 2 studies out of 44 also considering multimodal transport, like pedestrians, in the reward function design.
For example, \cite{wu2020multi1} developed a comprehensive reward function based on a weighted value incorporating six factors: total vehicle queue length, cumulative vehicle waiting time, vehicle delay, total vehicle throughput, traffic light status, and pedestrian waiting times at intersections. \cite{zhang2025towards} developed a fairness-based traffic signal control method that uses travelers affected by delay as the reward function, which includes travelers from private vehicles, public transport and pedestrians.
In studies focused solely on vehicles, reward functions are typically based on factors like vehicle delay, queue length (or the number of vehicles waiting at intersections), waiting time, throughput, and speed, either individually or in combination. Table \ref{table_reward} provides a detailed summary of reward definitions across studies. Queue length is a commonly used metric, with 11 studies relying exclusively on it as an indicator, as shown in the first row of Table \ref{table_reward}.
Additionally, some studies combine queue length with other indicators. For example, 5 studies use both queue length and vehicle waiting time, making this the second most popular reward mechanism. \cite{su2023emvlight} introduced the concept of "pressure", defined as the difference between traffic density on incoming and outgoing lanes. Other research works used changes between consecutive time steps as rewards; for instance, \cite{jiang2021distributed} and \cite{luo2024reinforcement} defined rewards as the change in vehicle queue length between timesteps, capturing evolving dynamics during learning. Similarly, \cite{li2021network} leveraged the change in vehicle delay between consecutive times as the reward function.
The most comprehensive reward structure was constructed by \cite{hu2024multi}, incorporating six indicators: vehicle throughput, queue length, phase switch penalty, minor street phase penalty, lane delay, and waiting time. Also, \cite{shen2025hierarchical} constructed a reward based on six indicators: total delay, queue length, waiting time, throughput, average travel time and average speed.

\begin{table}[htbp]
\footnotesize
\centering
\caption{Summary of reward mechanism components for each reference (\romannumeral1).}
\label{table_reward}
\begin{threeparttable}
\begin{tabular}{p{6.0cm}p{8.3cm}p{1.5cm}}
\hline
Reward Components & References & Quantity \\
\hline
Queue length (veh) & \cite{zhu2023multi,liu2023multiple,wang2020large,liu2023gplight,tan2020multi,bokade2023multi,zhang2022neighborhood,wang2020stmarl,fereidooni2025multi,liu2025globallight,lai2025llmlight,wang2025towards} & 12 \\
\hline
Waiting time (veh)& \cite{wang2024large,yang2023hierarchical,yang2021ihg}; & \multirow{2}{*}{5} \\
Queue length (veh) & \cite{chu2019multi,liu2023traffic} & \multirow{2}{*}{} \\
\hline
Difference of queue lengths between two consecutive timesteps (veh) & \cite{jiang2021distributed,luo2024reinforcement} & 2 \\
\hline
Waiting time (veh) & \cite{wu2022distributed,xu2024graph} & 2 \\
\hline
Throughput (veh) & \multirow{6}{*}{\cite{hu2024multi}} & \multirow{6}{*}{1} \\
Queue length (veh) & \multirow{6}{*}{} & \multirow{6}{*}{} \\
Phase switch penalty & \multirow{6}{*}{} & \multirow{6}{*}{} \\
Minor street phase penalty & \multirow{6}{*}{} & \multirow{6}{*}{} \\
Lane delay/speed loss (veh) & \multirow{6}{*}{} & \multirow{6}{*}{} \\
Waiting time (veh) & \multirow{6}{*}{} & \multirow{6}{*}{} \\
\hline
Total delay (veh) & \multirow{6}{*}{\cite{shen2025hierarchical}} & \multirow{6}{*}{1} \\
Queue length (veh) & \multirow{6}{*}{} & \multirow{6}{*}{} \\
Waiting time (veh) & \multirow{6}{*}{} & \multirow{6}{*}{} \\
Throughput (veh) & \multirow{6}{*}{} & \multirow{6}{*}{} \\
Average travel time (veh) & \multirow{6}{*}{} & \multirow{6}{*}{} \\
Average speed (veh) & \multirow{6}{*}{} & \multirow{6}{*}{} \\
\hline
Waiting time (veh) & \multirow{5}{*}{\cite{wu2020multi1}} & \multirow{5}{*}{1} \\
Waiting time (ped) & \multirow{5}{*}{} & \multirow{5}{*}{} \\
Queue length (veh) & \multirow{5}{*}{} & \multirow{5}{*}{} \\
Throughput (veh) & \multirow{5}{*}{} & \multirow{5}{*}{} \\
Traffic light blinking condition & \multirow{5}{*}{} & \multirow{5}{*}{} \\
\hline
Waiting time (veh) & \multirow{4}{*}{\cite{ge2021multi}} & \multirow{4}{*}{1} \\
Queue length (veh) & \multirow{4}{*}{} & \multirow{4}{*}{} \\
Total number of vehicles & \multirow{4}{*}{} & \multirow{4}{*}{} \\
Speed (veh) & \multirow{4}{*}{} & \multirow{4}{*}{} \\
\hline
Number of approaching vehicles & \multirow{4}{*}{\cite{ma2020feudal}} & \multirow{4}{*}{1} \\
Cumulative waiting time (veh) & \multirow{4}{*}{} & \multirow{4}{*}{} \\
Number of vehicles reaching destination & \multirow{4}{*}{} & \multirow{4}{*}{} \\
Traffic flow liquidity & \multirow{4}{*}{} & \multirow{4}{*}{} \\
\hline
Traffic density (veh) & \multirow{3}{*}{\cite{wang2021adaptive}} & \multirow{3}{*}{1} \\
Speed (veh) & \multirow{3}{*}{} & \multirow{3}{*}{} \\
Distance between intersections & \multirow{3}{*}{} & \multirow{3}{*}{} \\
\hline
\end{tabular}
        \textit{Notes:} "veh" means vehicle and "ped" stands for pedestrian.
\end{threeparttable}
\end{table}

\begin{table}[htbp]
\footnotesize
\centering
\caption{Summary of reward mechanism components for each reference (\romannumeral2).}
\label{table_reward}
\begin{threeparttable}
\begin{tabular}{p{6.0cm}p{8.3cm}p{1.5cm}}
\hline
Reward Components & References & Quantity \\
\hline
Travel time delay (veh) & \multirow{3}{*}{\cite{song2024cooperative}} & \multirow{3}{*}{1} \\
Queue length (veh) & \multirow{3}{*}{} & \multirow{3}{*}{} \\
Throughput (veh) & \multirow{3}{*}{} & \multirow{3}{*}{} \\
\hline
Waiting time (veh) & \multirow{3}{*}{\cite{bie2024multi}} & \multirow{3}{*}{1} \\
Queue length (veh) & \multirow{3}{*}{} & \multirow{3}{*}{} \\
Throughput (veh) & \multirow{3}{*}{} & \multirow{3}{*}{} \\
\hline
Queue length (veh) & \multirow{3}{*}{\cite{zhang2022distributed}} & \multirow{3}{*}{1} \\
Traffic density (veh) & \multirow{3}{*}{} & \multirow{3}{*}{} \\
Cumulative stop delay (veh) & \multirow{3}{*}{} & \multirow{3}{*}{} \\
\hline
Waiting time (veh) & \multirow{3}{*}{\cite{nie2025cmrm}} & \multirow{3}{*}{1} \\
Emission (veh) & \multirow{3}{*}{} & \multirow{3}{*}{} \\
Cumulative collision risk index (veh) & \multirow{3}{*}{} & \multirow{3}{*}{} \\
\hline
Waiting time (veh) & \multirow{3}{*}{\cite{jia2025multi}} & \multirow{3}{*}{1} \\
Throughput (veh) & \multirow{3}{*}{} & \multirow{3}{*}{} \\
Variance of green time across phases & \multirow{3}{*}{} & \multirow{3}{*}{} \\
\hline
Total number of vehicles & \multirow{2}{*}{\cite{lee2019reinforcement}} & \multirow{2}{*}{1} \\
Speed (veh) & \multirow{2}{*}{} & \multirow{2}{*}{} \\
\hline
Queue length (veh) & \multirow{2}{*}{\cite{yoon2025decentralized}} & \multirow{2}{*}{1} \\
Throughput (veh) & \multirow{2}{*}{} & \multirow{2}{*}{} \\
\hline
Waiting time (veh) & \multirow{2}{*}{\cite{ren2024two}} & \multirow{2}{*}{1} \\
Emission (veh) & \multirow{2}{*}{} & \multirow{2}{*}{} \\
\hline
Queue length (veh) & \multirow{2}{*}{\cite{tao2023network}} & \multirow{2}{*}{1} \\
Average speed (veh) & \multirow{2}{*}{} & \multirow{2}{*}{} \\
\hline
Number of moving vehicles & \multirow{2}{*}{\cite{satheesh2025constrained}} & \multirow{2}{*}{1} \\
Number of waiting vehicles & \multirow{2}{*}{} & \multirow{2}{*}{} \\
\hline
Difference of delays between two consecutive timesteps (veh) & \cite{li2021network} & 1 \\
\hline
Average delay time (veh) & \cite{abdoos2020cooperative} & 1 \\
\hline
Cumulative delay (veh) & \cite{el2013multiagent} & 1 \\
\hline
Pressure (veh) & \cite{su2023emvlight} & 1 \\
\hline
Throughput (veh) & \cite{wu2020multi2} & 1 \\
\hline
Speed (veh) & \cite{wang2021gan} & 1 \\
\hline
Number of delayed travellers & \cite{zhang2025towards} & 1 \\
\hline
\end{tabular}
        \textit{Notes:} "veh" means vehicle and "ped" stands for pedestrian.
\end{threeparttable}
\end{table}

\subsection{Multi-agent DRL framework}
The fundamental DRL framework for multi-agent TSC can be categorized into three types based on the learning process: value-based DRL methods, policy-based DRL methods, and hybrid DRL methods. Each of these will be explained in the following sections.

\subsubsection{Value-based DRL methods}
Value-based DRL focuses on learning the value of each action in a given state without explicitly learning the policy. The policy is derived by selecting actions that maximize the estimated values \citep{mckenzie2022modern}. For multi-agent TSC, value-based DRL methods primarily include Q-learning and deep Q-learning variants, such as Deep Q-Network (DQN), Double Deep Q-Network (DDQN), Double Dueling Deep Q-Learning Network (DDDQN), and QMIX. Table \ref{value_based} summarizes the value-based DRL approaches from the reviewed papers.

Traditional Q-learning methods have been employed in two studies. One study used a broad learning system instead of a deep learning framework for modeling \citep{zhu2023multi}, while the other applied the standard Q-learning algorithm for decision-making \citep{abdoos2020cooperative}.
In particular, \cite{zhu2023multi} proposed a multi-agent DRL approach using broad learning architectures \citep{gong2021research}, which significantly reduces model complexity and, consequently, training time, without sacrificing performance. Q-learning was utilized within this broader framework.
In contrast, \cite{abdoos2020cooperative} introduced an innovative TSC method that integrates game theory with DRL. This method features a two-mode agent architecture in which each intersection is controlled by an agent that operates independently during normal traffic conditions and cooperatively when traffic becomes congested. Each agent utilizes Q-learning to manage its intersection under normal conditions. However, in times of congestion, the agents shift to a cooperative mode, using game theory to dynamically coordinate traffic signals with adjacent agents.
\cite{el2013multiagent} proposed MARLIN-ATSC, a decentralized multi-agent RL system for network-wide adaptive signal control that coordinates neighboring intersections via modular Q-learning and learned neighbor-policy models.

Multi-agent DRL based on DQN is the most popular method, employed by a total of 12 studies.
\cite{wang2021adaptive} proposed a cooperative, group-based multi-agent DRL framework that uses k-Nearest-Neighbor (kNN) for state representation to enhance coordination among agents. They adapted the DQN framework to a multi-agent environment for Q-value approximation.
\cite{wang2024large} introduced a multi-agent DRL approach tailored for complex, large-scale traffic networks, where each intersection is modeled as a graph with lanes acting as nodes and traffic relationships forming the edges. Graph Attention Network (GAT) is utilized to learn lane embeddings that encapsulate spatial and geometric features of intersections. These embeddings are then fed into DQN for the evaluation of Q-values.
\cite{bie2024multi} presented a multi-agent DRL framework that tackles intersection heterogeneity by employing Graph Attention Networks (GATs) to capture interactions and dependencies between intersections. The core element of this DRL approach is DQN, which aims to select optimal actions by maximizing Q-values.
\cite{liu2023multiple} employed a decentralized form of DQN, in which each agent operates independently but shares some information with neighboring agents.
\cite{xu2024graph} addressed the issue of missing data in TSC by introducing a Wasserstein Generative Adversarial Network (WGAN) to estimate the state space and maintain data integrity. They also utilized a Graph Neural Network (GNN) to aggregate state features from multiple agents, which is crucial for informed decision-making. Each agent then applied decentralized DQN-based learning to select the optimal action.
In the study by \cite{wang2021gan}, a Generative Adversarial Network (GAN) framework was employed for traffic data recovery, utilizing a decentralized model based on DQN.
\cite{tan2020multi} utilized the bootstrapped DQN algorithm, which enhances exploration through an ensemble of behavior policies. This approach demonstrated superior efficiency and robustness compared to standard DQN by managing uncertainty more effectively.
\cite{wu2020multi2} demonstrated a decentralized multi-agent DRL approach based on DQN, applying transfer learning to assess the robustness of the algorithm and traditional control methods under varying traffic scenarios, such as changes in traffic volume, flow patterns, and sensor reliability.
\cite{zhang2022neighborhood} implemented a decentralized multi-agent DRL where each agent learns independently from its own observations as well as those from neighboring intersections, utilizing Hysteretic DQN for strategy optimization. Hysteretic DQN, which is a variant of the standard DQN, introduces a novel update mechanism designed to address the complexities in multi-agent environments.
\cite{liu2023traffic} introduced a multi-agent DRL strategy for optimizing traffic light control in areas affected by epidemics, incorporating a Convolutional Neural Network (CNN) to process state representations within the DQN framework. Meanwhile, \cite{wang2020stmarl} discussed a spatio-temporal multi-agent DRL framework that integrates GNN, RNN, and DQN. RNNs are employed to capture the temporal dynamics of traffic, while GNNs model cooperative structures among traffic lights based on an adjacency graph, enabling each traffic light to make decisions through deep Q-learning.

\cite{wang2020large} addressed the issue of large-scale traffic signal control with a scalable independent double DQN (DDQN) method aimed at reducing the overestimation issues found in standard DQN algorithms. The approach incorporates an upper confidence bound policy to balance the exploration-exploitation trade-off.
\cite{liu2023gplight} created a grouped multi-agent DRL framework for managing large-scale traffic signal control by clustering agents (intersections) based on environmental similarity. A Graph Convolutional Network (GCN) extracts features from the traffic network, modeled as a graph, and QMIX is implemented for policy learning. Agents within the same group share a neural network model and parameters, facilitating efficient learning and decision-making.
\cite{bokade2023multi} demonstrated a centralized training with decentralized execution approach for inter-agent communication in large-scale traffic control. Using QMIX, agents learn to identify which parts of their messages need to be shared with others to achieve effective coordination.
\cite{liu2025globallight} proposed a method that (i) uses a multi-head GAT to extract local multi-dimensional features, (ii) mines global similarities via two representation-space losses to group distant yet similar intersections and share policy parameters, and (iii) integrates these in a QMIX learner.
\cite{hu2024multi} modeled traffic signal coordination using a multi-agent system that incorporates both local and global agents with a DDQN approach. The system comprises seven local agents, one for each intersection, which operate independently but are coordinated by a global agent. This setup allows for localized decision-making while aligning with global objectives.
\cite{jiang2021distributed} proposed a distributed multi-agent DRL approach that utilizes graph decomposition to handle large-scale traffic signal control. The network of intersections is segmented into subgraphs according to connectivity, with each controlled by a collective of agents operating within the DRL framework. They employ the framework of DDDQN to enhance learning efficiency. DDDQN combines the advantages of DDQN to minimize Q-value overestimation. The dueling architecture in DDDQN enables more effective state-value learning. Additionally, the prioritized experience replay makes it concentrate on critical experiences to accelerate the learning process.

\begin{table}[htbp]
\centering
\caption{Summary of value-based DRL frameworks}
\label{value_based}
\begin{tabular}{p{3.0cm}p{10.0cm}p{1.5cm}}
\hline
DRL framework & References & Quantity \\
\hline
Q-learning & \cite{zhu2023multi,abdoos2020cooperative,el2013multiagent} & 3\\
DQN & \cite{wang2021adaptive,wang2024large,bie2024multi,liu2023multiple,xu2024graph,wang2021gan,tan2020multi,wu2020multi2,zhang2022neighborhood,lee2019reinforcement,liu2023traffic,wang2020stmarl,fereidooni2025multi,lai2025llmlight,tao2023network} & 15\\
QMIX & \cite{liu2023gplight,bokade2023multi,liu2025globallight} & 3\\
DDQN & \cite{hu2024multi,wang2020large} & 2\\
DDDQN & \cite{jiang2021distributed} & 1\\
\hline
\end{tabular}
\end{table}

\subsubsection{Policy-based DRL methods}
Policy-based DRL methods focus on directly learning the policy that determines the agent’s actions in a given state. Unlike value-based approaches, which learn a value function and derive a policy from it, policy-based methods optimize the policy itself without the need for an explicit value function \citep{li2017deep}. For multi-agent TSC, policy-based DRL methods primarily include Proximal Policy Optimization (PPO) and Counterfactual Multi-Agent (COMA) Policy Gradient. Table \ref{policy_based} provides a summary of the policy-based DRL methods examined in the reviewed studies.

\cite{ren2024two} put forward a two-tier coordinated DRL framework designed for multi-agent TSC. This framework comprises a local cooperation layer and a global cooperation layer. At the local layer, each intersection functions as an independent agent, making decisions based on its immediate traffic conditions. Meanwhile, the global layer ensures the coordination of actions across the entire intersection network. PPO is utilized as the primary reinforcement learning algorithm for agents in both layers.
Similarly, \cite{zhang2022distributed} employed the PPO algorithm, incorporating parameter sharing among multiple agents, to address multi-agent TSC along an arterial corridor. This method accelerates the learning process and minimizes computational overhead.
\cite{luo2024reinforcement} presented a hybrid action space method for optimizing traffic signal control, integrating both discrete and continuous actions. The discrete component manages the selection of traffic light stages, while the continuous component fine-tunes the duration of green light intervals. To efficiently handle these hybrid action spaces, a DRL framework based on PPO was developed. \cite{nie2025cmrm} proposed CMRM, a collaborative MARL method for multi-objective TSC: it augments independent PPO with a Cooperation Enhancement Module (CEM) based on graph attention to share informative neighbor context. \cite{satheesh2025constrained} proposed a constrained DRL method based on PPO. A multi-agent DRL based on PPO actor–critic augmented with a learned Lagrange Cost Estimator that stabilizes updates of the Lagrange multiplier to boost throughput/reduces delay with fewer constraint violations.

\cite{song2024cooperative} suggested a novel DRL framework based on the COMA Policy Gradient method. COMA, a specialized actor-critic approach for multi-agent systems, employs a centralized critic to compute the advantage function for each agent’s actions relative to a counterfactual baseline. The actors function in a decentralized manner, utilizing both local and globally shared information. Furthermore, the framework integrates an innovative scheduler component to improve information sharing among the decentralized actors.

\begin{table}[htbp]
\centering
\caption{Summary of policy-based DRL frameworks}
\label{policy_based}
\begin{tabular}{p{4.0cm}p{9.0cm}p{1.5cm}}
\hline
DRL framework & References & Quantity \\
\hline
PPO & \cite{ren2024two,zhang2022distributed,luo2024reinforcement,nie2025cmrm,fereidooni2025multi,satheesh2025constrained} & 6\\
COMA Policy Gradient & \cite{song2024cooperative} & 1\\
\hline
\end{tabular}
\end{table}

\subsubsection{Hybrid DRL methods}
Hybrid DRL combines elements from both value-based and policy-based methods, leveraging the strengths of each to mitigate their individual limitations \citep{farazi2021deep}. These hybrid DRL approaches are primarily built on an actor-critic structure. Common frameworks include Advantage Actor-Critic (A2C), Actor-Critic, Deep Deterministic Policy Gradient (DDPG), Asynchronous Advantage Actor-Critic (A3C), and Soft Actor-Critic (SAC). Table \ref{hybrid_based} provides a summary of hybrid DRL methods applied to the multi-agent TSC research problem.

\cite{yang2023hierarchical} proposed a novel approach that combines multi-granularity information fusion with a mutual information optimization framework for urban traffic signal control. Multi-granularity fusion integrates both current and historical traffic signal states across multiple time steps to capture a comprehensive view of traffic dynamics. GNNs are used to encode state information, leveraging spatial relationships among traffic nodes. The approach employs decentralized execution with centralized training using an actor-critic DRL framework.
\cite{yang2021ihg} devised a multi-agent DRL approach that combines inductive learning with a heterogeneous GNN. In this model, the traffic network is represented as a heterogeneous graph where nodes correspond to different elements, such as vehicles, traffic signals, and intersections. The inductive learning framework can generate embeddings for nodes not encountered during training, which is especially useful in dynamic traffic networks with continuously changing vehicles and conditions. An actor-critic framework supports agent learning in this approach.

\cite{su2023emvlight} employed a multi-agent A2C-based DRL framework for managing dynamic routing and traffic signal control of Emergency Vehicles (EMVs). This framework integrates dynamic EMV routing with traffic signal preemption, facilitating efficient EMV movement while minimizing the impact on regular traffic. The proposed DRL framework incorporates a policy-sharing mechanism and a spatial discount factor, allowing each traffic signal, functioning as an agent, to make informed decisions based on both local and network-wide information.
\cite{wu2022distributed} introduced two DRL algorithms enhanced by game theory principles: Nash Advantage Actor–Critic (Nash-A2C) and Nash Asynchronous Advantage Actor–Critic (Nash-A3C). Nash-A2C combines the Nash Equilibrium framework with conventional Actor-Critic methods to alleviate congestion and reduce network delays. Nash-A3C builds upon Nash-A2C by introducing asynchronous operations across multiple agents within the traffic network, utilizing a centralized training process paired with decentralized execution.
\cite{chu2019multi} highlighted a multi-agent DRL approach for TSC utilizing a decentralized A2C framework. The approach incorporates neighborhood fingerprints, enabling each agent to consider the recent policies of neighboring agents. This mechanism reduces uncertainty and enhances decision-making despite incomplete information. Additionally, a spatial discount factor is applied to reduce the impact of distant agents, allowing each agent to prioritize local traffic conditions effectively.

\begin{table}[htbp]
\centering
\caption{Summary of hybrid-based DRL frameworks}
\label{hybrid_based}
\begin{tabular}{p{3.0cm}p{10.0cm}p{1.5cm}}
\hline
DRL framework & References & Quantity \\
\hline
A2C & \cite{su2023emvlight,wu2022distributed,chu2019multi,ma2020feudal} & 4\\
A3C & \cite{wu2022distributed} & 1\\
Actor-Critic & \cite{yang2023hierarchical,yang2021ihg} & 2\\
DDPG & \cite{li2021network,wu2020multi1,jia2025multi,shen2025hierarchical,yoon2025decentralized} & 5\\
SAC & \cite{ge2021multi,zhang2025towards,wang2025towards} & 3\\
\hline
\end{tabular}
\end{table}

\cite{li2021network} outlined a Knowledge-Sharing DDPG (KS-DDPG) approach for multi-agent TSC. Built on the DDPG algorithm, KS-DDPG employs centralized training with decentralized execution. During training, agents learn from the aggregated experiences of all agents in a centralized manner. However, execution is decentralized, allowing each agent to make independent decisions based on private observations and shared knowledge. The knowledge-sharing component accelerates model convergence without significantly increasing computational demands.
\cite{wu2020multi1} explored a multi-agent recurrent DDPG framework combining DDPG with Long Short-Term Memory (LSTM) to improve the handling of temporal dependencies and partial observability in traffic environments. Both the actor and critic networks in DDPG are equipped with LSTM layers. Simulations validate the model's effectiveness, demonstrating its ability to reduce vehicle and pedestrian congestion more effectively than other methods, particularly in complex urban settings with multiple intersections. \cite{jia2025multi} proposed a hybrid MADRL traffic-signal framework that fuses spatio-temporal attention (GAT for spatial deps + LSTM for temporal trends) with hierarchical control (sub-region agents + a global coordinator) under CTDE using multi-agent DDPG learning framework. The results reports 25\% lower average waiting time and the highest throughput versus fixed-time, actuated, Max-Pressure, DQN, PPO, and vanilla Multi-Agent DDPG (MADDPG) baselines. 

\cite{ge2021multi} designed a multi-agent DRL approach based on SAC, tailored for multi-agent environments with entropy maximization to maintain a balance between exploration and exploitation. The approach incorporates a multi-view encoder to process state inputs from diverse perspectives: a 1D vector represents local intersection data, while a 2D state representation, processed through a CNN, captures spatial traffic dynamics. Furthermore, a transfer learning mechanism with cooperative guidance enables agents to develop generalizable skills while adapting to the unique characteristics of specific intersections. \cite{zhang2025towards} proposed an M$^{2}$SAC, a cooperative masked SAC for corridor-level signal control, where a hybrid actor network generate a Gaussian-sampled phase mask that can defer the last phase to the next cycle and then picks green-time splits via a masked softmax function. Ablation studies show the introduced mask mechanism yields extra gains.

\subsection{Summary}

Overall, prior multi-agent DRL studies have demonstrated strong potential for adaptive traffic signal control, especially in complex and stochastic settings where rule-based coordination is difficult. Value-based approaches are typically easier to implement for discrete phase decisions and can be relatively stable, while policy-based methods such as PPO offer direct policy optimization and can better accommodate richer action parameterizations. Hybrid methods further benefit from combining policy learning with value estimation, enabling more flexible modeling and improved exploration. These strengths have contributed to consistent reductions in congestion-related metrics across diverse simulation studies. However, existing work also exhibits clear limitations. First, most evaluations rely heavily on a small set of simulators (mainly SUMO/CityFlow), which may constrain realism and transferability to practice. Second, reward design is predominantly vehicle-centric (delay/queue/throughput), and the literature largely overlooks fairness, equity, or individual-level experience, particularly for vulnerable or non-motorized users. Third, scalability and coordination remain challenging in large networks due to non-stationarity, communication/coordination overhead, and sensitivity to state representation choices, even when graph-based modeling is introduced. These gaps highlight the need for human-centric multi-agent DRL that explicitly encodes equity objectives and remains robust under diverse corridor conditions.

\section{Methodology}
\label{method}
In this section, we provide a detailed outline of the methodology for human-centric multi-agent TSC. We begin with an overview of our approach, accompanied by an illustration of the entire workflow. Following this, we elaborate on the specifics of the proposed method, focusing on five key components: environment and state representation, agent and action space, reward mechanism, network architecture, and the learning algorithm used for agent training.

\begin{figure}[htbp]
\begin{center}
\includegraphics[width=0.95\textwidth]{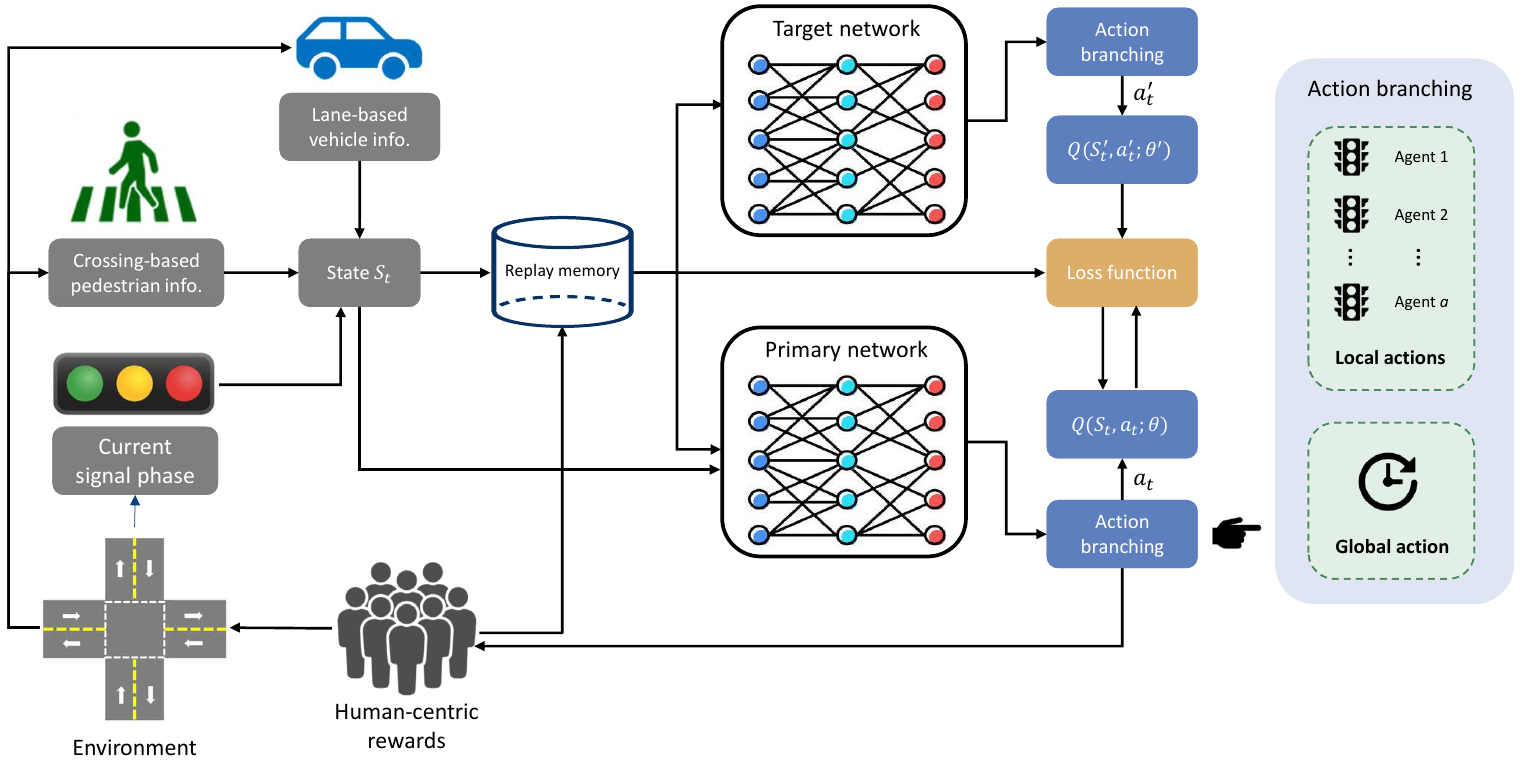}
\caption{Workflow of the proposed MA2B-DDQN for human-centric multi-agent traffic signal control.}
\label{workflow}
\end{center}
\end{figure}

\subsection{Overview}
MA2B-DDQN employs centralized training to efficiently model this multi-agent TSC research problem. The workflow of our approach is depicted in Figure \ref{workflow}. The state is a global and dynamic representation including vehicle, pedestrian, and signal phase information, which are all gathered from the traffic environment. The definition of state will be further explained in Section \ref{state}.
MA2B-DDQN is fundamentally built on a double DQN framework, integrating both a primary and a target network to mitigate the overestimation of Q-values.
The architecture for both the primary and target networks is identical, utilizing neural networks; details of the architecture will be provided in Section \ref{architecture}.
Our approach introduces an innovative action branching that includes both local and global action spaces to model discrete action spaces and to counter the curse of dimensionality in multi-agent optimization environments. Local actions target individual agents, while a global action manages all agents simultaneously. The computation of loss incorporates the rewards and output Q-values from both the primary and target networks.
The reward mechanism is designed with a human-centric concept, prioritizing fairness and the equitable consideration of all transportation service users’ needs. Details about this reward mechanism will be disclosed in Section \ref{rewards}.

\subsection{Environment and state}
\label{state}

\begin{figure}[htbp]
\begin{center}
\includegraphics[width=0.93\textwidth]{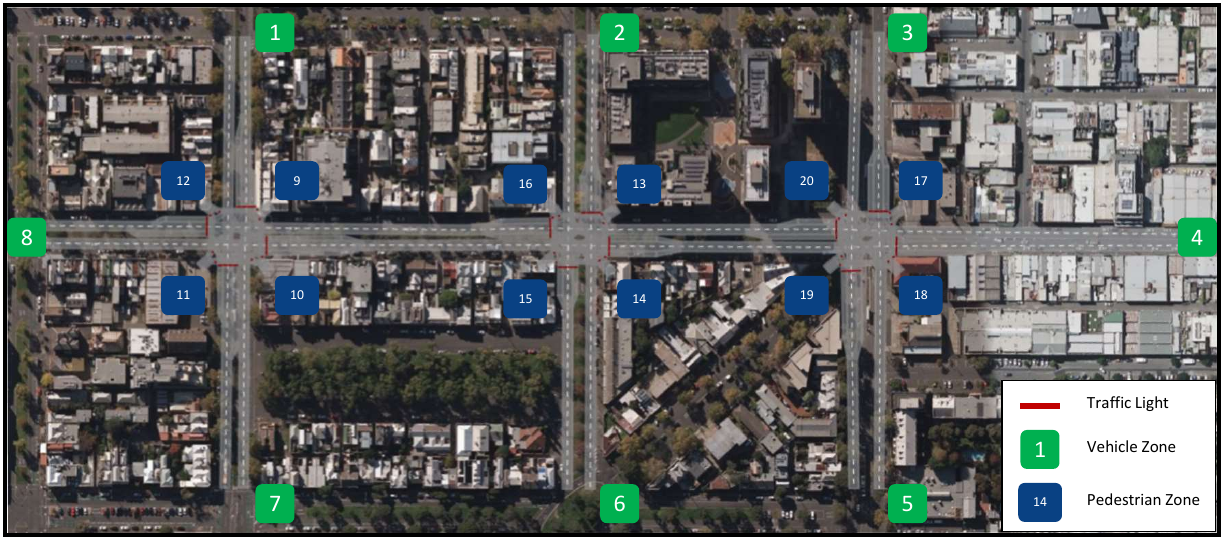}
\caption{The corridor selected for experiments features three intersections and is equipped with 8 entry/exit zones (indexed from 1 to 8) for vehicle simulations. Additionally, there are 12 entry/exit zones (indexed from 9 to 20) for pedestrian simulations, which do not account for the direction of pedestrian movements.}
\label{road_network}
\end{center}
\end{figure}

\begin{figure}[htbp]
\begin{center}
\includegraphics[width=0.76\textwidth]{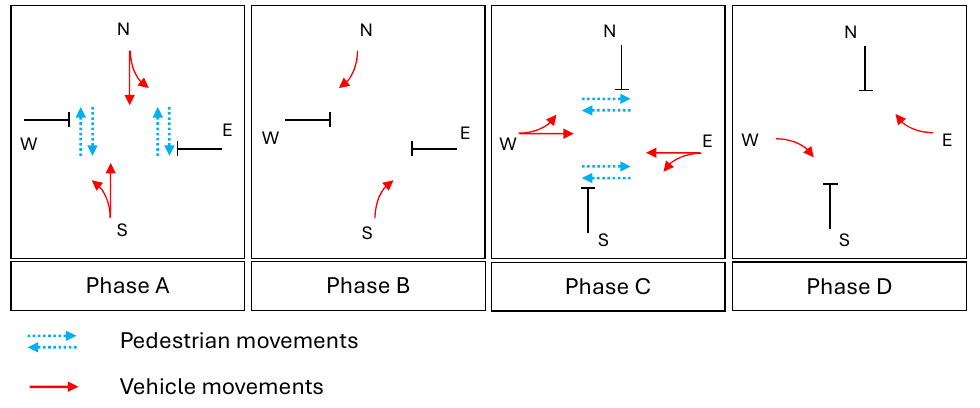}
\caption{Phase and stage diagram of the traffic signal system.}
\label{phases}
\end{center}
\end{figure}

The simulation environment utilizes the PTV VISSIM traffic simulation software, modeling a naturalistic corridor featuring three busy intersections within the Australian Integrated Multimodal EcoSystems (AIMES) testbed, which was established by The University of Melbourne. This simulation integrates various modes of transportation, including cars, buses, and pedestrians. It takes place on a specific corridor located on Elgin Street in Carlton, Victoria, Australia, which is illustrated in Figure \ref{road_network}. This corridor is configured with a traffic light system that operates on a four-phase signal plan, as detailed in Figure \ref{phases}. The traffic signal system is designed with lagging right turns, where protected right turns (phases B and D) occur subsequent to the through and left turn movements. The through and left turns are consolidated into the same phases (phase A and phase C) as they do not conflict with each other.

The state representation includes three components: traffic condition $C$, pedestrian information $N$, and signal state $G$. The state is represented as
\begin{equation}\label{eq1}
S=\left [ C,N,G \right ].
\end{equation}
For representing the traffic condition $C$, the incoming lanes at intersections are divided into cells with equal length (e.g., 6 meters to fit a typical car), and each cell is assigned an index. Critical traffic information, including vehicle occupancy ($O$), velocity ($V$), acceleration ($A$), and number of occupants/passengers ($P$), is stored for these cells correspondingly. Vehicle occupancy is a binary variable, where 1 indicates occupied and 0 indicates unoccupied. In contrast, velocity, acceleration, and number of occupants/passengers are all continuous scalar variables. Traffic condition $C$ is denoted as
\begin{equation}\label{eq2}
C=\left[ C_{1}, \cdots, C_{i}, \cdots ,C_{a} \right], \qquad C_{i}=\left[ O_{i}, V_{i}, A_{i}, P_{i} \right],
\end{equation}
where $a$ denotes the number of signalised intersections (agents); in this study, $a=3$. The representation of $O$ is provided as an example, as illustrated by
\begin{equation}
O_i \in \mathbb{R}^{16\times l}, \qquad r(a,b)=4(a-1)+b,\;\; a,b\in\{1,2,3,4\},
\end{equation}
and
\begin{equation}
\big[O_i\big]_{r(a,b),\,t}=o_{i,a,b,t}, \qquad t\in\{1,\ldots,l\}.
\end{equation}
where the first subscript $i$ identifies the intersection, the second subscript $a$ represents the index of approach (4 approaches in total), the third subscript $b$ means the index of lane (4 lanes in total), $l$ stands for the total number of cells for each lane, and the fourth subscript $t$ indicates the index of cell.
Likewise, we can determine $V_{i}$, $A_{i}$, and $P_{i}$, though they are not explicitly defined here. 

For representing the pedestrian information $N$, we have
\begin{equation}\label{eq4}
N=\left[ N_{1},N_{2},\cdots ,N_{a} \right], \qquad N_{i}=\left[ n_{i,1}, n_{i,2}, \cdots , n_{i,8} \right],
\end{equation}
where $n_{i,1}$ stands for the number of pedestrians waiting at the 1st pedestrian zone of intersection $i$. Each intersection has a total of 8 pedestrian zones in this study, reflecting the direction of pedestrian movements. 

Lastly, the signal state $G$ is encoded with one-hot embedding, as demonstrated by
\begin{equation}\label{eq5}
G=\left[ G_{1}, G_{2}, \cdots , G_{a} \right],
\end{equation}
and
\begin{equation}\label{eq6}
G_{i}=\left[\text{Phase A}, \text{Phase B}, \text{Phase C}, \text{Phase D}  \right],
\end{equation}
where each phase is represented by embedding a value of 1 when active and 0 otherwise. For instance, if phase C is currently activated at the $i^{\textrm{th}}$ intersection, then $G_{i}$ would be represented as $\left [ 0,0,1,0 \right ]$.

\subsection{Agent and action}
\label{action}
\begin{figure}[htbp]
\begin{center}
\includegraphics[width=0.80\textwidth]{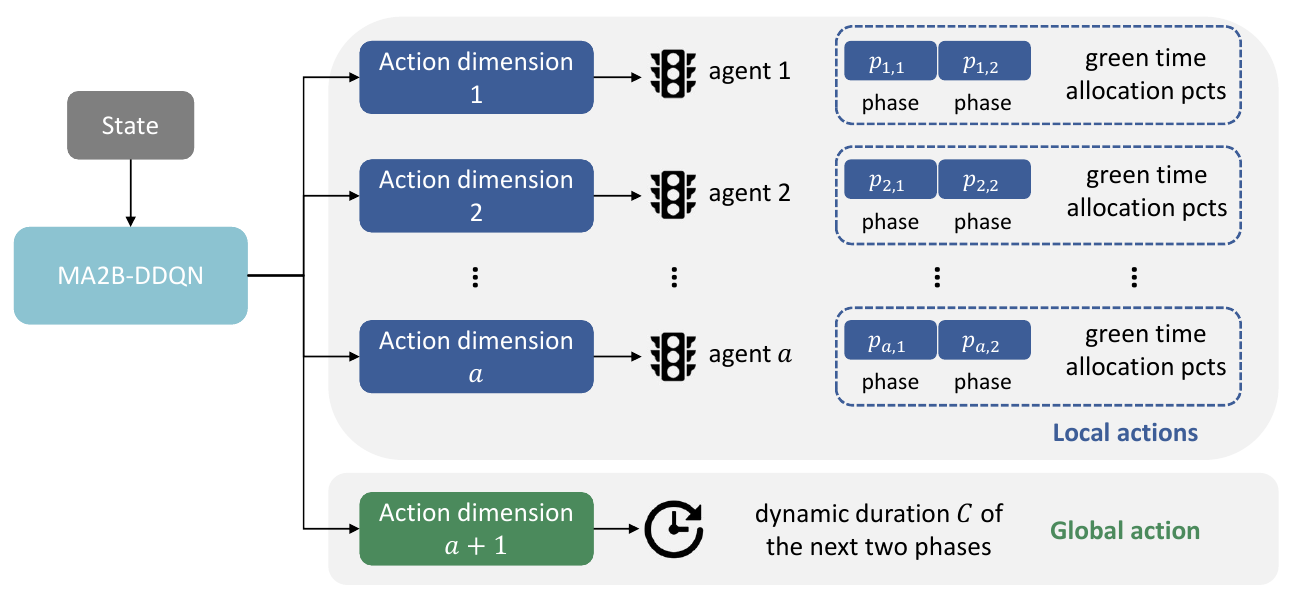}
\caption{Illustration of the action space via the action branching architecture.}
\label{faction}
\end{center}
\end{figure}

\begin{figure}[htbp]
\begin{center}
\includegraphics[width=0.70\textwidth]{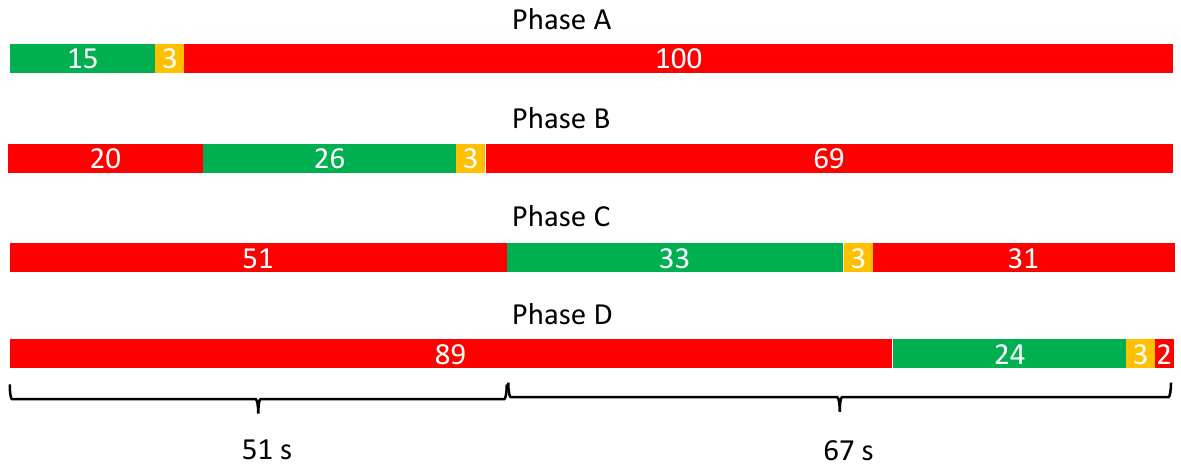}
\caption{An example of the signal plan for an agent. 51 s is the predicted optimal global action for the next two phases. 67 s is the next round of prediction. The green time for each phase (e.g., 15 s and 26 s) is also the predicted optimal green time.}
\label{signal}
\end{center}
\end{figure}

In the corridor for experiments, the traffic signals for vehicles follow a red-green-amber sequence, while pedestrian signals utilize a red-green-flashing red sequence, both of which are consistent with the actual traffic lights in Melbourne. The combined duration of green and flashing red phases for pedestrian signals overlaps with the green signal for both left-turning and through movements. We modeled the action space of this problem as a set of discrete actions. Our multi-agent strategy aims to predict the optimal signal plan for each agent for the upcoming two phases (half cycle), with the duration of these phases dynamically adjusted according to real-time traffic conditions. We focus on the next half cycle rather than a full cycle mainly based on the following two considerations: (1) the traffic condition is fast-changing and a full cycle time is too long to effectively be utilized for capturing the evolving trends; (2) more phases coming in could exponentially increase the complexity of the action space due to the discrete action space as well as multi-agent scenarios. To address this, we implement a system with a consistent number of two phases but introduce a dynamic global action to enhance the timing plan options for agents. Figure \ref{faction} outlines the action space and the action branching architecture of our approach. Assuming there are $a$ agents/intersections for TSC optimization, the action space is divided into $a+1$ dimensions. The first $a$ dimensions manage local actions for individual agents, and the final dimension, i.e., a global action, controls all agents. Each local dimension generates an action for its corresponding agent, as shown in Figure \ref{faction}. The $(a+1)^{\textrm{th}}$ dimension determines the total duration $C$ of the next two phases. For local actions, each agent's output involves the allocation of green time across the two phases. The distribution varies among agents and is tailored based on the traffic conditions. For simplicity, we maintain $p_{i,1}+p_{i,2}=1$ for each agent $i$, thus only $p_{i,1}$ requires optimization; $p_{i,2}$ can be directly derived as $1-p_{i,1}$. This significantly reduces the complexity of modeling the discrete action space in value-based DRL. Figure \ref{signal} displays an example of the optimal signal plan generated by our model. In this study, the amber, all-red clearance, and minimum green times are fixed at 3 seconds, 2 seconds, and 10 seconds, respectively. Both $p_{i,1}$ and $C$ are treated as discrete variables in this model. $p_{i,1}$ ranges from 0.02 to 0.98 in increments of 0.02, providing 49 options per variable. For the total duration $C$, it ranges from 30 to 70 seconds, with each second constituting a step, resulting in 41 distinct dimensions for this action space.

\subsection{Reward mechanism}
\label{rewards}
The reward is defined as the negative total of travelers impacted by intersection delays along the corridor, with impacts measured in seconds of delay per operation. These travelers include users of various transportation modalities such as buses, private vehicles, and pedestrians. The metric for impacted travelers is derived from the number of individuals experiencing delays, including both pedestrians and vehicle occupants (including those in buses and private vehicles). Delays for vehicle occupants and passengers are assessed by monitoring the speed of their vehicles. For vehicles, delays are considered when either a vehicle's speed or deceleration falls below a specific threshold, indicating being affected. Specifically, for occupants or passengers in vehicles, a delay is defined as occurring when
\begin{equation}\label{eq7}
v_{veh} < \theta_{v_{veh}},
\end{equation}
where $v_{veh}$ is the speed of the vehicle. $\theta_{v_{veh}}$ is the threshold which is set to 5 km/h in this study.
For pedestrians, delay is determined by the presence of pedestrians waiting at designated  crossing zones.
The reward at timestep $t$ is formulated as
\begin{equation}\label{eqreward}
r_{t} = -\left (w_{veh} \sum_{i=1}^{m}N_{i,t}^{veh} +w_{ped}\sum_{j=1}^{n}N_{j,t}^{ped} \right ).
\end{equation}
where $m$ is the total number of delayed vehicles among the corridor network, and $n$ is the number of intersections. $N_{i,t}^{veh}$ is the number of occupants in the $i^{\textrm{th}}$ delayed vehicle at timestep $t$, and $N_{j,t}^{ped}$ is the number of delayed pedestrians in the $j^{\textrm{th}}$ intersection at timestep $t$. $w_{veh}$ and $w_{ped}$ are weighting coefficients. In this study, we set $w_{veh} = w_{ped} = 1$.

\subsection{Network architecture}
\label{architecture}
The primary and target networks share the same branching Q-network architecture (Figure 7). Given the global state $S_{t}$, the network first applies a shared feature extractor consisting of two fully connected (dense) layers with ReLU activations, each followed by layer normalization, to obtain a latent representation. This shared representation is then fed into $n+1$ parallel output heads, one per action dimension in the action-branching design. Each head is a linear layer that outputs a vector of Q-values for its corresponding discrete action set (i.e., $Q^{(j)}(S_{t},.)$ for $j=1,\cdots ,n+1$). During execution, the action for each branch is selected via argmax over the branch Q-values, and the selected $n+1$ branch actions are combined to form the final joint action. In our experiments, each of the first $n$ (local) action branches has 49 discrete options, while the $(n+1)^{\textrm{th}}$ (global) action branch has 41 options, consistent with the action space definition in Section 3.3.

\begin{figure}[htbp]
\begin{center}
\includegraphics[width=0.65\textwidth]{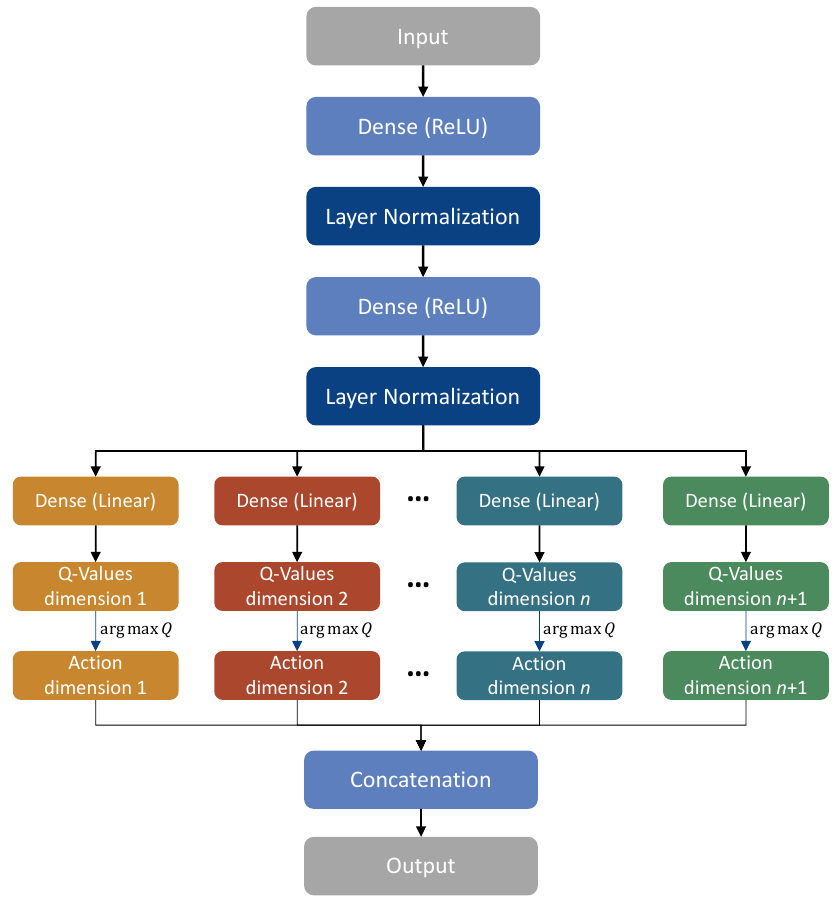}
\caption{Architecture of the primary and target networks in MA2B-DDQN.}
\label{structure}
\end{center}
\end{figure}

\subsection{Learning algorithm}
For arbitrary dimension $j$, $1\leqslant j\leqslant n+1$, we have its loss function, as defined by
\begin{equation}\label{eq9}
\mathcal{L}_{d_{j}} =\frac{1}{N}\sum_{i=1}^{N}\left ( Q\left ( S_{i},a_{i,d_{j}};\theta \right )-\left ( r_{i}+\gamma Q \left (S'_{i},\arg\max_{{a}'_{i,d_{j}}} Q\left ( {S}'_{i},{a}'_{i,d_{j}};\theta  \right );{\theta}' \right ) \right ) \right )^{2},
\end{equation}
where $r_{i}$ denotes the reward at timestep $i$, and $\gamma$ is the discount factor, which is set to $\gamma = 0.95$ in this study. $\theta $ and $\theta'$ are the parameters of the primary network and target network, respectively. $S_{i}$ is the state matrix at timestep $i$, and $S'_{i}$ is the next state matrix of $S_{i}$. $\textup{arg max}_{{a}'_{i,d_{j}}}Q\left ( {S}'_{i},{a}'_{i,d_{j}};\theta  \right )$ represents the action that maximizes the Q-value with the next state $S'_{i}$ according to the primary network. $a_{i,d_{j}}$ is the action at the $j^{\textrm{th}}$ dimension, and $a'_{i,d_{j}}$ is the next action. $N$ is the sample size.

The overall loss function is derived by averaging the losses across all dimensions of action branching. Considering that we have $n+1$ action dimensions, the overall loss function is defined as
\begin{equation}\label{eqloss}
\mathcal{L} = \frac{1}{n+1}\sum_{j=1}^{n+1}\mathcal{L}_{d_{j}}.
\end{equation}
The objective of training the primary network is to minimize the loss function $\mathcal{L}$. Optimal parameters $\theta$ are determined through the backpropagation algorithm, utilizing the Adam optimizer, a method analogous to that employed in basic neural network training \citep{zhang2024knowledge}.

The proposed MA2B-DDQN DRL model for optimizing multi-agent TSC is detailed in Algorithm \ref{algorithm_1}. In Steps 1 and 2, the weights for the primary and target networks are initialized, while Step 3 sets up the replay buffer. Steps 4 through 30 focus on training the entire network across each episode within the DRL framework. Step 5 initializes both the state and cumulative reward. In Steps 6 to 28, data is gathered for each time step in the episode for training purposes. Step 7 captures the state based on traffic conditions, pedestrian details, and signal status. Steps 8 to 15 involve generating actions based on the current state, with a warm-up phase (set to 5 episodes in this study) providing random actions at the initial stages. Steps 11 to 15 manage the action branching process. Step 16 executes the derived action, and Step 17 measures the reward, which is added to the cumulative reward in Step 18. Step 19 records the transition data, and Steps 20 to 22 conduct mini-batch training of the networks. Steps 23 to 27 check if the simulation episode has concluded, and Step 29 decides whether to save the primary network for this episode.
\begin{breakablealgorithm}
    \caption{MA2B-DDQN for multi-agent TSC optimization}
    \label{algorithm_1}
    \begin{algorithmic}[1]
        \State Initialize $\theta$ for primary network;
        \State Initialize $\theta'$ for target network with $\theta'\gets\theta$;
        \State Initialize replay buffer $R$;
        \For{episode = $1,\cdots ,M$}
            \State cumulative reward $\gets 0$
            \For{$t = 1,\cdots ,T$}
                \State $S_{t} \gets$ obtain state based on Eq. (\ref{eq1});
                \If{episode $\leqslant$  warm-up episode}
                    \State $a_{t} \gets $ random action;
                \Else{}
                    \For{$j = 1,\cdots ,n+1$}
                        \State $a_{t,d_{j}} \gets \arg\max_{a} \left ( Q\left ( S_{t},a;\theta  \right )+N_{t,j} \right )$;
                    \EndFor
                    \State $a_{t} \gets $ concat $\left\{a_{t,d_{j}}, 1\leqslant j\leqslant n+1\right\}$;
                \EndIf
                \State Execute action $a_{t}$ and obtain next state $S_{t+1}$;
                \State Get the human-centric reward $r_{t}$ based on Eq. (\ref{eqreward});
                \State cumulative reward $\gets$ cumulative reward + $r_{t}$;
                \State Store transition $\left ( S_{t},a_{t},r_{t},S_{t+1} \right )$ in $R$;
                \State Sample a minibatch from $R$;
                \State Update $\theta$ by minimizing $\mathcal{L}$ based on Eq. (\ref{eq9}) and Eq. (\ref{eqloss});
                \State Update $\theta'$ using $\theta'\gets\theta$;
                \If{meet termination condition}
                    \Break
                \Else{}
                    \Continue
                \EndIf
            \EndFor
            \State Determine whether to save the best $\theta$ based on cumulative reward;
        \EndFor
    \end{algorithmic}
\end{breakablealgorithm}

\section{Experiments and analyses}
\label{experiment}

\subsection{Experimental setup}
The proposed model was trained and tested across seven distinct scenarios. Scenario 1 represents a typical off-peak period with moderate traffic demand. Scenario 2 simulates rush hour with higher traffic volumes. Scenarios 3 and 4 replicate school start and end periods: Scenario 3 represents the morning school period with increased traffic reaching a specific area, while Scenario 4 describes the afternoon school period with increased traffic flow away from this area. Scenario 5 models heavy traffic nearing saturation levels. Scenario 6 captures the fluctuating traffic flow, evolving from normal to peak rush conditions. Scenarios 1 to 6 include pedestrian traffic, while Scenario 7 examines an intersection without pedestrian traffic. Traffic demand data for each scenario are accessible online\footnote[6]{\url{https://1drv.ms/f/s!AqOdc5BgGmiqjmuSpIpWLn01bY_V?e=6GOZwU}}. All models were built in Python using TensorFlow, with experiments conducted on a Windows system powered by an AMD EPYC 7702 CPU @ 1996 MHz with 64 cores.

\subsubsection{Evaluation metrics}
We present the metrics of aggregate individual delay (AID), per capita delay (PCD), average vehicle delay per second (AVDS), average pedestrian delay per second (APDS), average delay per vehicle (ADV), average waiting time per pedestrian (AWTP), and improvement in percentage (IMP), which are formulated by
\begin{equation}\label{eq10}
\textup{AID}= \sum_{i=1}^{T}n_{delay\_individual}^{i}, \qquad \textup{PCD}= \frac{\textup{AID}}{D},
\end{equation}
\begin{equation}\label{eq17_1}
\textup{AVDS}= \frac{\sum_{i=1}^{T}n_{delay\_veh}^{i}}{T}, \qquad \textup{APDS}= \frac{\sum_{i=1}^{T}n_{delay\_ped}^{i}}{T},
\end{equation}
\begin{equation}\label{eq17_3}
\textup{ADV}= \frac{\sum_{i=1}^{N_{v}}d_{veh}^{i}}{N_{v}}, \qquad \textup{AWTP}= \frac{\sum_{i=1}^{N_{p}}d_{ped}^{i}}{N_{p}},
\end{equation}
and
\begin{equation}\label{eq12}
\textup{IMP}=\frac{\textup{Baseline (AID)}-\textup{Our (AID)}}{\textup{Baseline (AID)}}\times 100\%,
\end{equation}
where $n_{delay\_individual}^{i}$ represents the total number of individuals experiencing delays within the corridor network at timestamp $i$, including pedestrians, occupants, and passengers, as described in Section \ref{rewards}.
$n_{delay\_veh}^{i}$ and $n_{delay\_ped}^{i}$ stand for the total number of delayed vehicles and pedestrians at timestamp $i$.
$D$ represents the total travel demands in terms of travelers and $T$ denotes the duration of the test in seconds. Additionally, $N_{v}$ and $N_{p}$ correspond to the total number of vehicles and pedestrians, respectively.
Furthermore, $d_{veh}^{i}$ and $d_{ped}^{i}$ quantify the total delay or waiting time for the $i^{\textrm{th}}$ vehicle or pedestrian, respectively.
"Baseline (AID)" refers to the AID of the baseline method, while "Our (AID)" denotes the AID metric of our approach. In our experiments, each test scenario was executed 20 times in the VISSIM simulator using different random seeds, and the results are reported as the mean ($\mu$) and standard deviation ($\sigma$).

\subsubsection{Benchmarking}
Nine baseline methods are developed for benchmarking performance with the proposed MA2B-DDQN, including
\begin{itemize}
  \item \textbf{Fixed Signal with Webster's Formula (FS-WF)}: FS-WF employs Webster's Formula to determine the optimal cycle time, followed by green splitting to allocate green time for each phase. Both the optimal cycle time estimation and green splitting are based on traffic demand. For optimal cycle time estimation, the saturation flow is set at 1,900 vehicles/hour/lane.
  \item \textbf{Fixed Signal with Green Wave control (FS-GW)}: FS-GW applies green wave control, beginning with a reference intersection that utilizes Webster's Formula and green splitting for signal configuration. Subsequent intersections then implement green wave control, calibrated according to their distance from the reference intersection and the design speed. Similarly, when estimating the optimal cycle time using Webster’s Formula, the saturation flow is defined as 1,900 vehicles/hour/lane.
  \item \textbf{MADQN} \citep{wei2018intellilight}: This baseline approach formulates the problem using a discrete action space, basically based on DQN. It employs a centralized learning and decentralized execution framework for the multi-agent environment. The key parameter settings are as follows: learning rate = 0.0001, batch size = 128, discount factor ($\gamma$) = 0.95, replay buffer size = 2,000,000, number of episodes = 500, and Gaussian noise ($\sigma$) = 0.2 for action exploration.
  \item \textbf{MADDQN}: This baseline method enhances MADQN by integrating the double deep Q-network (DDQN) framework as the base model to address the overestimation of Q-values in DQN. The study by \cite{yazdani2023intelligent} has utilized this approach to optimize traffic signals for single intersections. The primary hyperparameters are configured as follows: learning rate = 0.0001, batch size = 128, discount factor ($\gamma$) = 0.95, replay buffer size = 2,000,000, number of episodes = 500, and Gaussian noise ($\sigma$) = 0.2.
  \item \textbf{MADDPG} \citep{lowe2017multi}: In this approach, a continuous action space is utilized through MADDPG, fundamentally based on the DDPG framework. It uses a centralized training and decentralized execution framework to allow multiple agents to train simultaneously via a centralized process. The key hyperparameters are set as follows: actor learning rate = 0.00001, critic learning rate = 0.00005, batch size = 128, discount factor ($\gamma$) = 0.95, soft update parameter ($\tau$) = 0.005, replay buffer size = 2,000,000, number of episodes = 500, and Gaussian noise ($\sigma$) = 0.2.
  \item \textbf{CMRM} \citep{nie2025cmrm}: It employs a GAT-enhanced PPO framework for multi-agent traffic signal optimization under the CTDE paradigm. The main hyperparameters are: actor learning rate = 0.0001, critic learning rate = 0.0002, batch size = 64, discount factor ($\gamma$) = 0.95, clip ratio ($\epsilon $) = 0.2, and entropy coefficient = 0.05.
  \item \textbf{MASAC} \citep{wu2023multiagent}: This baseline is primarily built on the DRL framework of SAC, as applied by \cite{ge2021multi} for TSC optimization. It incorporates the concept of entropy to balance exploration with exploitation and uses a centralized learning and centralized execution approach. The key hyperparameters are configured as follows: actor learning rate = 0.00003, critic learning rate = 0.0003, temperature ($\alpha $) learning rate = 0.00003, batch size = 128, discount factor ($\gamma$) = 0.95, soft update parameter $\tau$ = 0.005, clip bound = [-10,2], target entropy = -1, replay buffer size = 2,000,000, and number of episodes = 500.
  \item \textbf{MAA2C} \citep{su2023emvlight}: MAA2C is fundamentally developed on the well-known A2C framework, which offers stability and efficiency for policy-based reinforcement learning tasks. It employs a centralized learning and decentralized execution framework for multi-agent environments. The key hyperparameters are configured as follows: learning rate = 0.0001, batch size = 32, discount factor ($\gamma$) = 0.95, advantage function = \textit{Q}-value - \textit{V}-value, number of episodes = 500, and Gaussian noise ($\sigma$) = 0.2 for action exploration.
  \item \textbf{MA2B-DQN}: This baseline approach defines the problem using a discrete action space, fundamentally relying on the DQN framework. It utilizes a centralized learning and decentralized execution strategy, serving as part of an ablation study to assess the impact of incorporating the DDQN framework into our model. It employs the same hyperparameters as MA2B-DDQN for consistency in comparison.
\end{itemize}

\subsection{Scenario 1}

\begin{table}[htbp]
\centering
\caption{Performance comparison for different models (scenario 1)}
\label{table_s1}
\begin{threeparttable}
\begin{tabular}{p{2.6cm}p{2.2cm}p{1.9cm}p{1.5cm}p{1.4cm}p{1.5cm}p{1.6cm}p{1.4cm}}
\hline
model & AID ($\mu$) & AID ($\sigma$) & PCD ($\mu$) & PCD ($\sigma$) & ADV (s) & AWTP (s) & IMP \\
\hline
FS-WF & 1,369,020.65 & 36,753.50 & 137.53 & 3.69 & 166.47 & 317.09 & 36.75\%\\
FS-GW & 1,102,796.65 & 29,581.45 & 110.79 & 2.94 & 139.72 & 315.67 & 21.48\%\\
MADQN & 1,027,639.40 & 86,911.21 & 103.24 & 8.73 & 117.07 & 75.82 & 15.74\%\\
MADDQN & 1,073,619.80 & 81,766.29 & 107.86 & 8.21 & 124.19 & 80.87 & 19.35\%\\
MADDPG & 1,002,162.95 & 97,396.30 & 100.68 & 9.78 & 116.97 & 99.84 & 13.60\%\\
CMRM & 1,251,988.70 & 75,473.73 & 125.78 & 7.58 & 156.34 & \textbf{45.81} & 30.84\%\\
MASAC & 985,492.65 & 53,883.69 & 99.00 & 5.41 & 109.71 & 77.36 & 12.14\%\\
MAA2C & 983,283.20 & 40,705.76 & 98.78 & 4.09 & 112.58 & 89.71 & 11.94\%\\
MA2B-DQN & 868,415.85 & 23,481.49 & 87.24 & 2.36 & 100.90 & 72.68 & 2.93\%\\
\textbf{MA2B-DDQN} & \textbf{865,874.70} & \textbf{22,953.49} & \textbf{86.99} & \textbf{2.31} & \textbf{98.37} & 77.23 & \textit{NA}\\
\hline
\end{tabular}
    \textit{Notes:} $\mu$ and $\sigma $ stand for mean and standard deviation. The same for the following tables.
\end{threeparttable}
\end{table}

Table \ref{table_s1} provides a comparative analysis of MA2B-DDQN against other baseline methods. MA2B-DDQN demonstrated the lowest AID, scoring 865,874.70, and the best PCD at 86.99, while also exhibiting the smallest standard deviations ($\sigma$) across both metrics. Additionally, it recorded the lowest ADV at 98.37 seconds per vehicle. However, its APDS is 77.23 seconds—relatively low, but slightly higher than that of certain baseline models, such as MA2B-DQN (72.68 seconds per pedestrian) and MADQN (75.82 seconds per pedestrian). These results underscore the stability and robustness of MA2B-DDQN. The incorporation of the double DQN architecture into MA2B-DDQN enhanced its performance by 2.93\% compared to the standard DQN, signifying a notable improvement. CMRM achieves the lowest AWTP (45.81 s). However, it also yields a very high vehicle delay (ADV = 156.34 s) and a high AID, indicating an imbalanced prioritization across modalities.
Among fixed-signal methods, the green wave control (FS-GW) approach performed notably better (21.48\% vs. 36.75\%) than the Webster's Formula-based fixed signal (FS-WF), which had the poorest results. Among DRL-based methods, MAA2C and MASAC emerged as the closest competitors to MA2B-DDQN, showing performance gaps of 11.94\% and 12.14\%, respectively. These findings reveal the great potential of actor-critic structured DRL with exploration-exploitation balance for multi-agent TSC optimization. Both MADQN and MADDQN exhibited similar performance, indicating limitations in handling the complex, high-dimensional action space of multi-agent environments. MADDPG performed poorly in this scenario, probably due to its sensitivity or inappropriateness for this context, as suggested by \cite{li2021network}.

\begin{figure}[htbp]
\begin{center}
\includegraphics[width=0.95\textwidth]{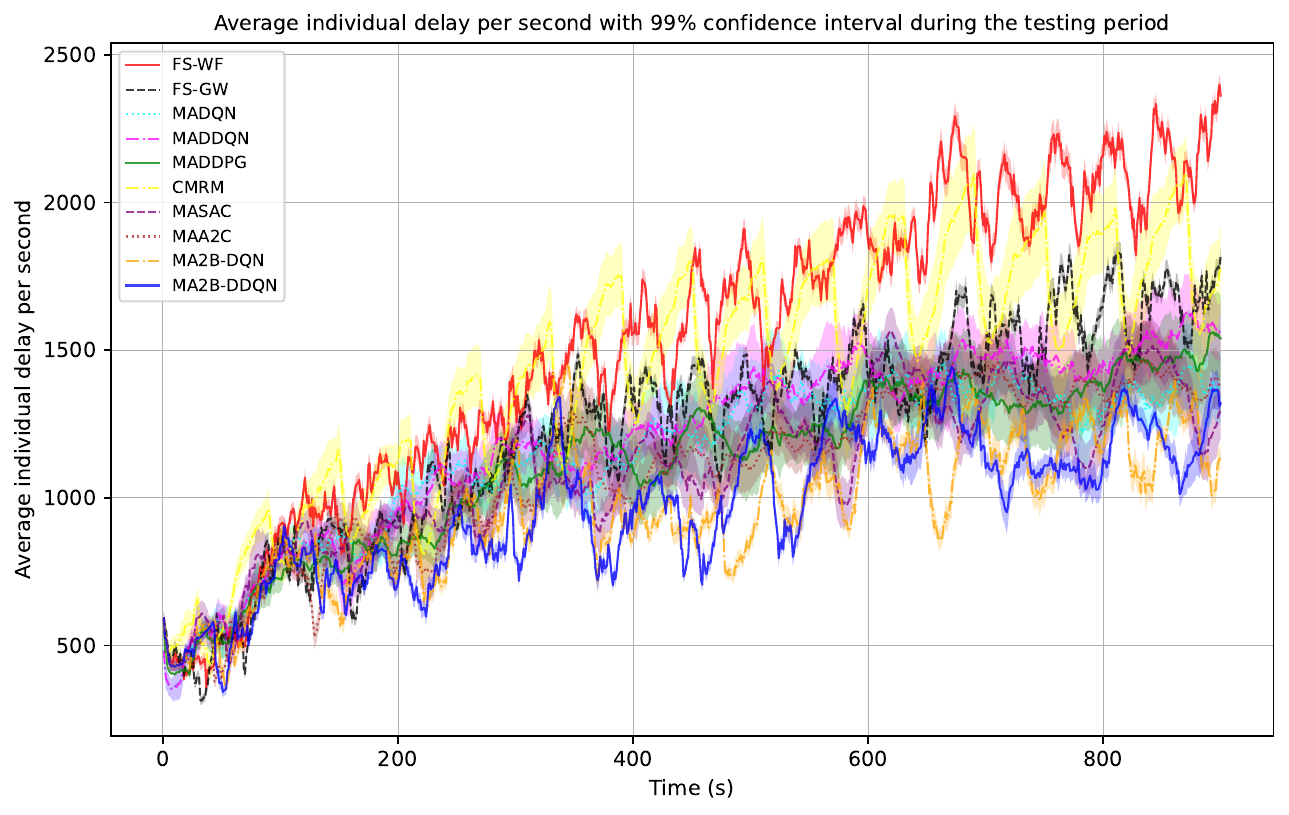}
\caption{Comparison of the individual delay and 99\% confidence interval throughout the testing period for scenario 1.}
\label{cost_comp_1}
\end{center}
\end{figure}

\begin{figure}[htbp]
\begin{center}
\subfloat[]{
    \includegraphics[width=0.85\textwidth]{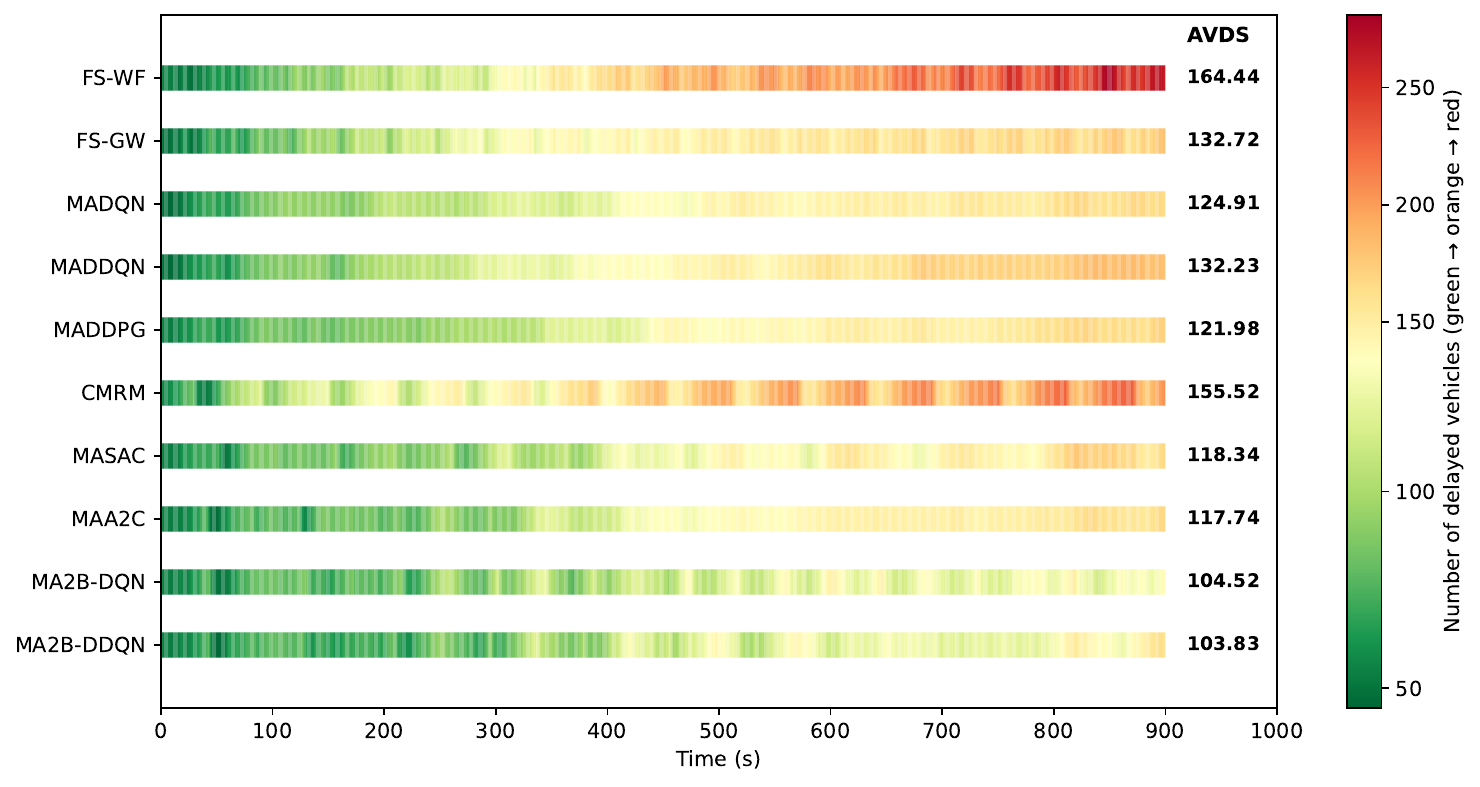}
    }
    
\subfloat[]{
    \includegraphics[width=0.85\textwidth]{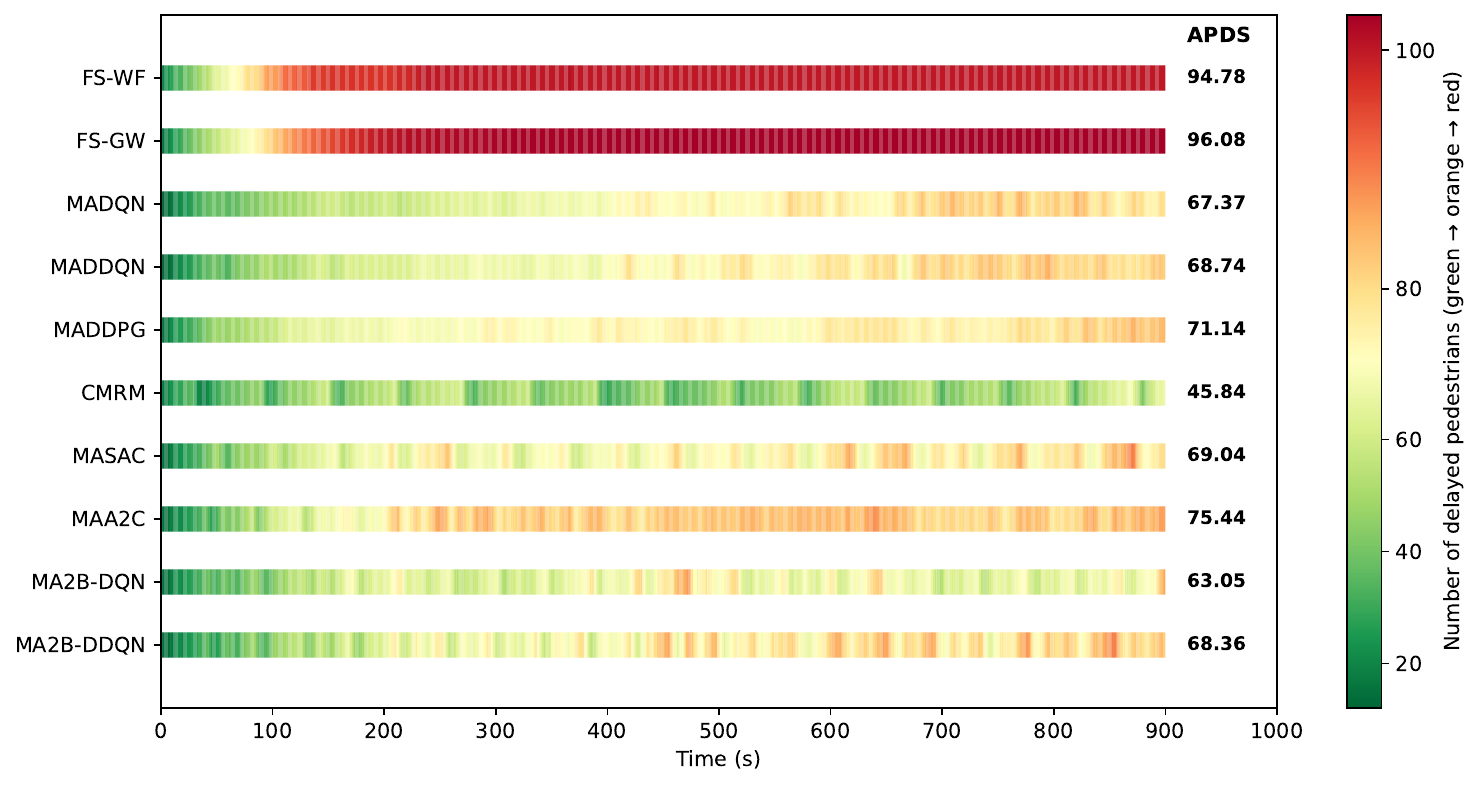}
    }
\caption{Analysis of different models in Scenario 1. (a) Vehicle delay over time (unit: vehicles per second). (b) Pedestrian delay over time (unit: pedestrians per second).}
\label{veh_ped_comp_1}
\end{center}
\end{figure}

Figure \ref{cost_comp_1} shows the average individual delay with a 99\% confidence interval over the independent testing period. As observed in the figure, MA2B-DDQN consistently demonstrates stable and reliable performance, maintaining a relatively low value throughout the period. The local peaks and troughs in the curve are due to the accumulation of travelers experiencing delay before or after each action execution (i.e., signal change).

\begin{figure}[htbp]
\centering
\subfloat[FS-WF: 200 s]{
    \includegraphics[width=0.24\textwidth]{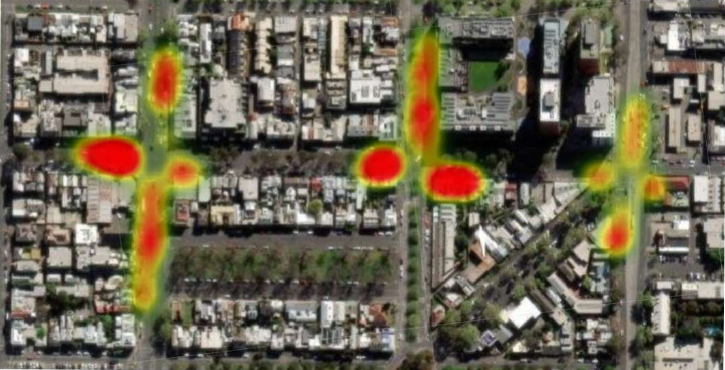}
}
\subfloat[FS-WF: 400 s]{
    \includegraphics[width=0.24\textwidth]{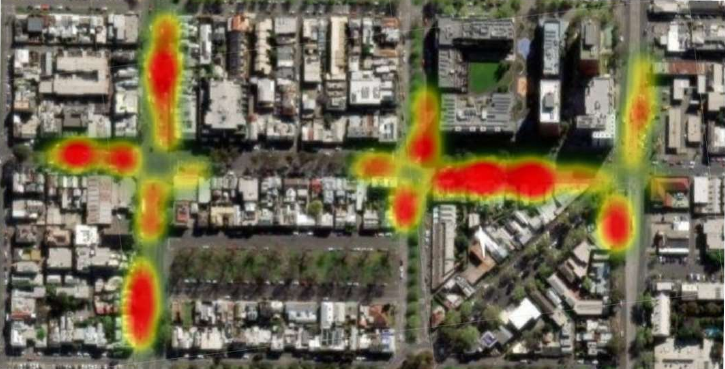}
}
\subfloat[FS-WF: 600 s]{
    \includegraphics[width=0.24\textwidth]{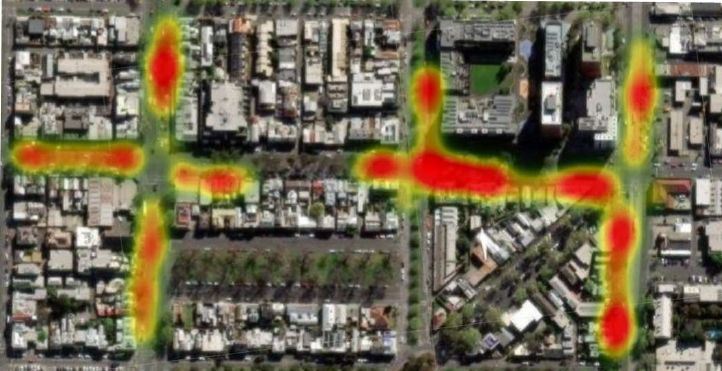}
}
\subfloat[FS-WF: 800 s]{
    \includegraphics[width=0.24\textwidth]{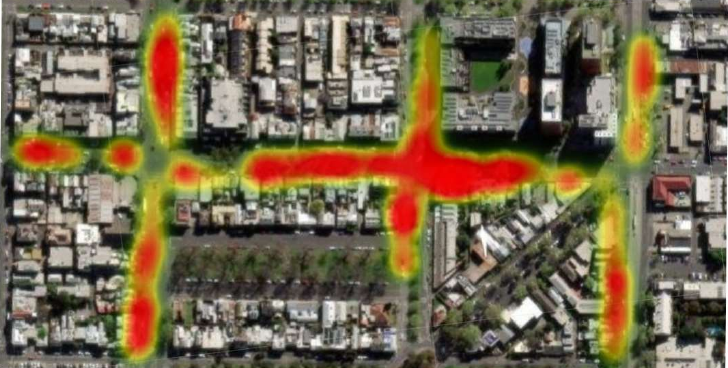}
}

\subfloat[MADDQN: 200 s]{
    \includegraphics[width=0.24\textwidth]{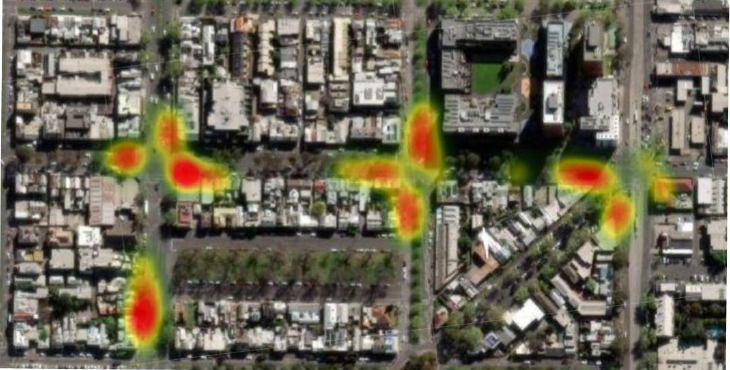}
}
\subfloat[MADDQN: 400 s]{
    \includegraphics[width=0.24\textwidth]{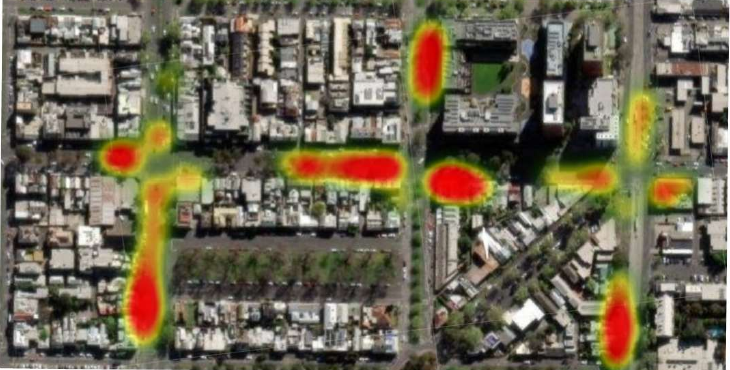}
}
\subfloat[MADDQN: 600 s]{
    \includegraphics[width=0.24\textwidth]{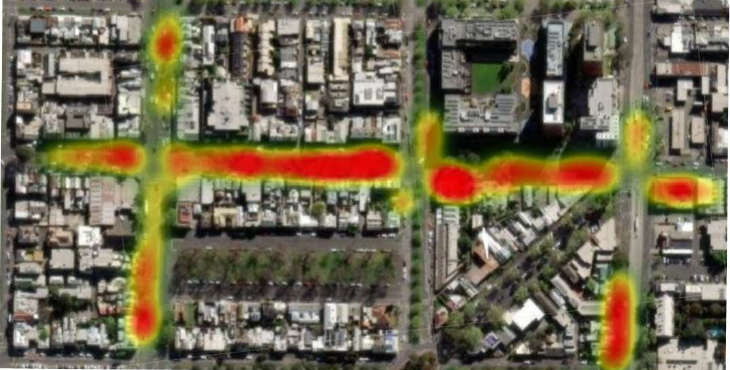}
}
\subfloat[MADDQN: 800 s]{
    \includegraphics[width=0.24\textwidth]{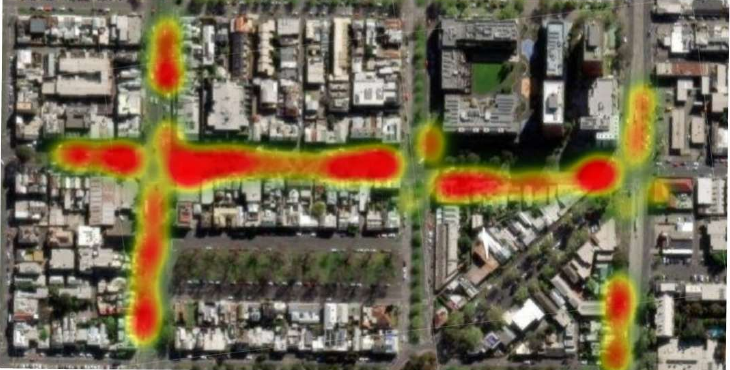}
}

\subfloat[MADDPG: 200 s]{
    \includegraphics[width=0.24\textwidth]{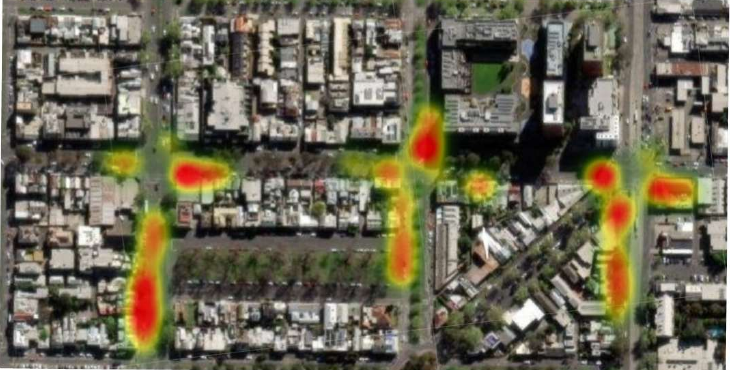}
}
\subfloat[MADDPG: 400 s]{
    \includegraphics[width=0.24\textwidth]{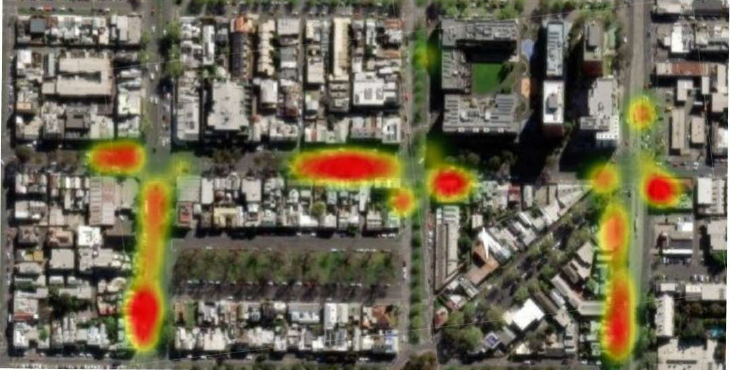}
}
\subfloat[MADDPG: 600 s]{
    \includegraphics[width=0.24\textwidth]{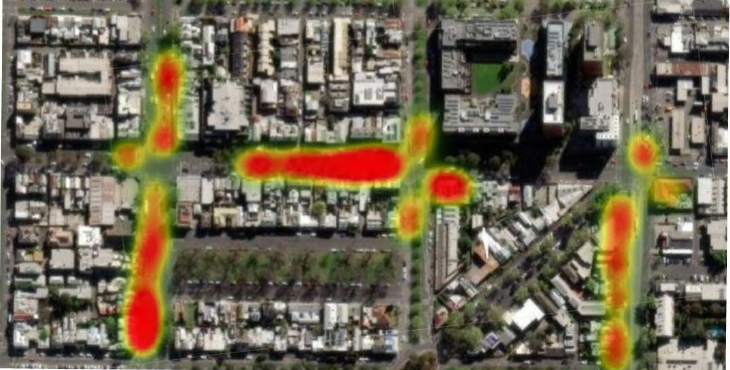}
}
\subfloat[MADDPG: 800 s]{
    \includegraphics[width=0.24\textwidth]{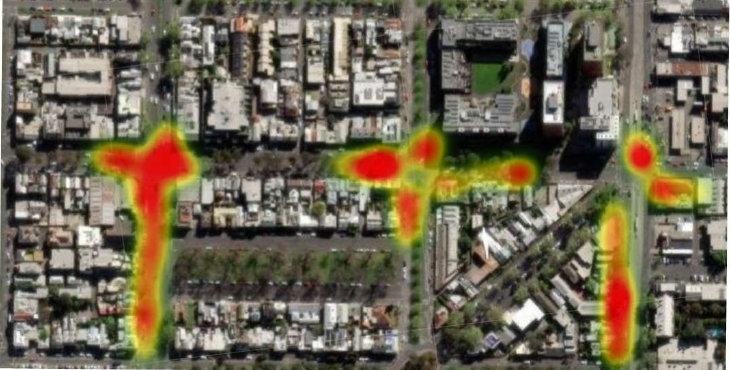}
}

\subfloat[MAA2C: 200 s]{
    \includegraphics[width=0.24\textwidth]{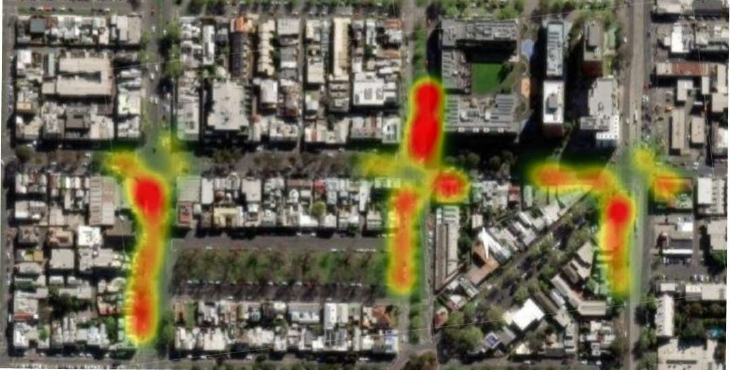}
}
\subfloat[MAA2C: 400 s]{
    \includegraphics[width=0.24\textwidth]{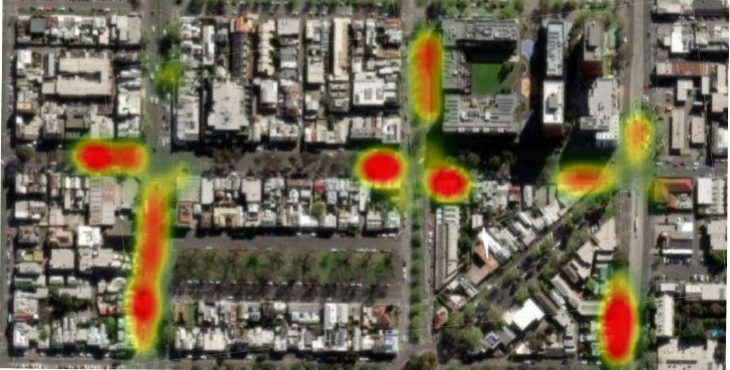}
}
\subfloat[MAA2C: 600 s]{
    \includegraphics[width=0.24\textwidth]{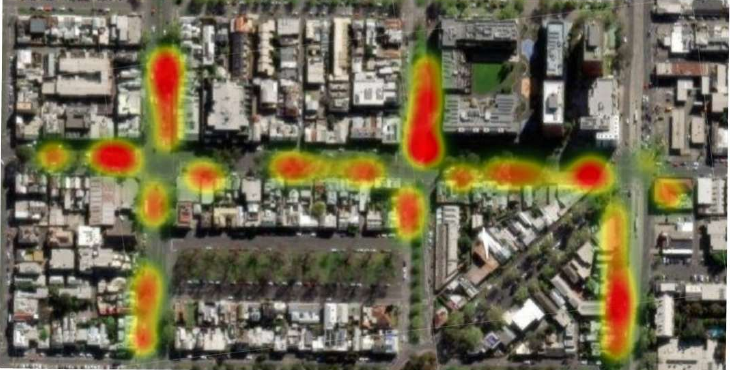}
}
\subfloat[MAA2C: 800 s]{
    \includegraphics[width=0.24\textwidth]{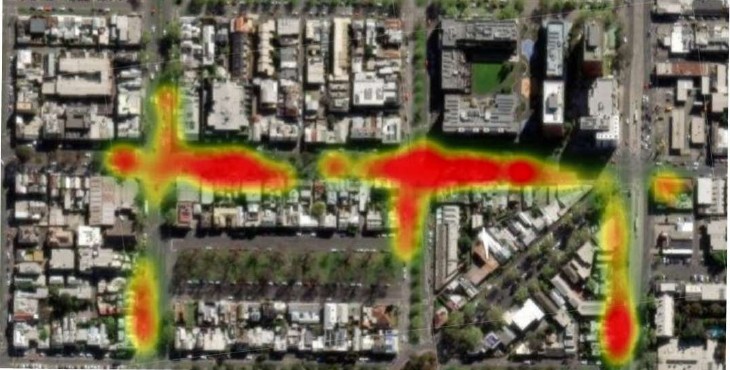}
}

\subfloat[MA2B-DQN: 200 s]{
    \includegraphics[width=0.24\textwidth]{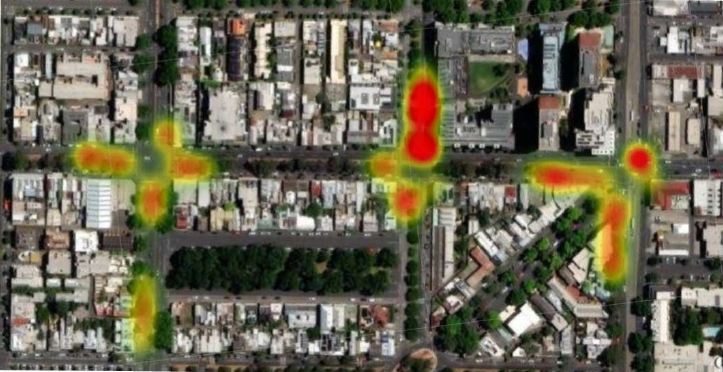}
}
\subfloat[MA2B-DQN: 400 s]{
    \includegraphics[width=0.24\textwidth]{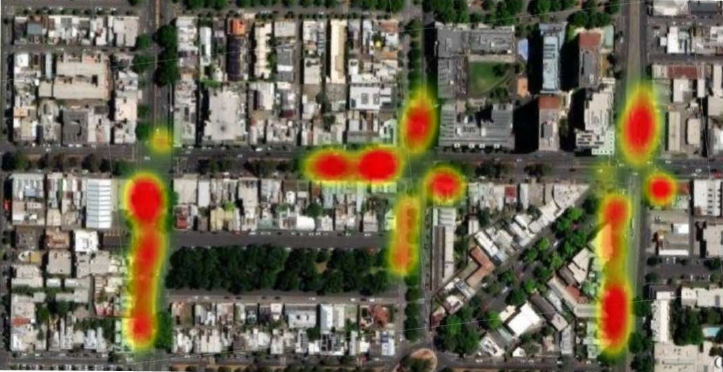}
}
\subfloat[MA2B-DQN: 600 s]{
    \includegraphics[width=0.24\textwidth]{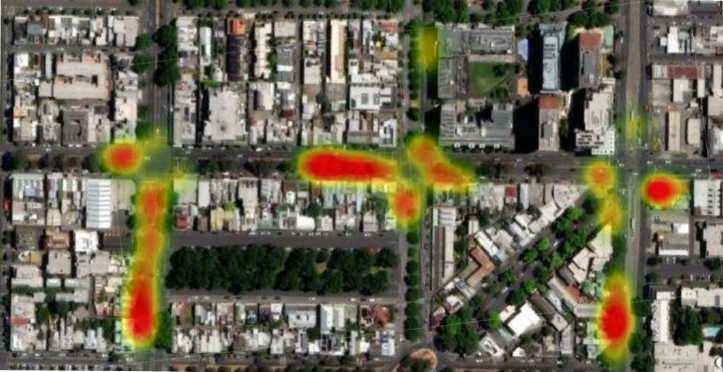}
}
\subfloat[MA2B-DQN: 800 s]{
    \includegraphics[width=0.24\textwidth]{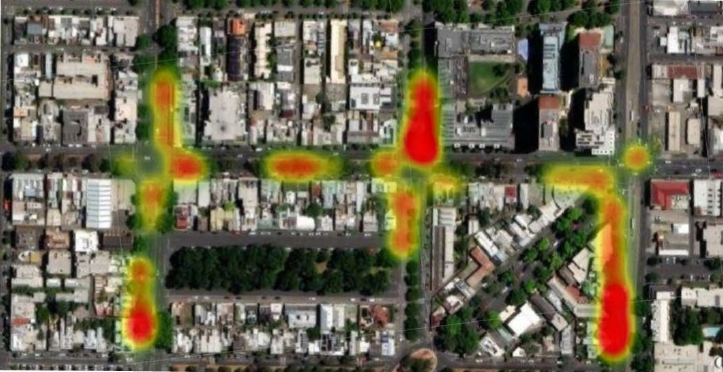}
}

\subfloat[MA2B-DDQN: 200 s]{
    \includegraphics[width=0.24\textwidth]{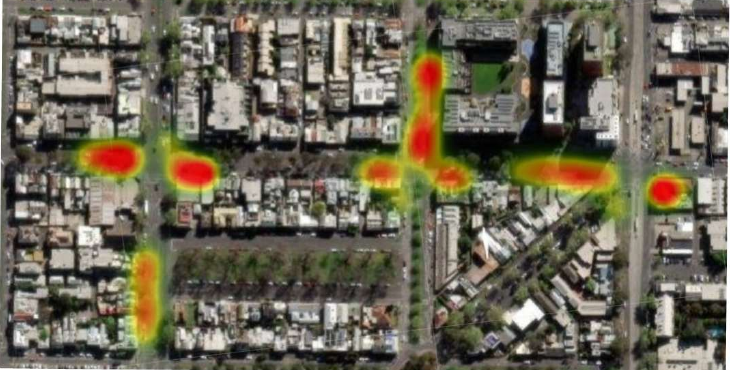}
}
\subfloat[MA2B-DDQN: 400 s]{
    \includegraphics[width=0.24\textwidth]{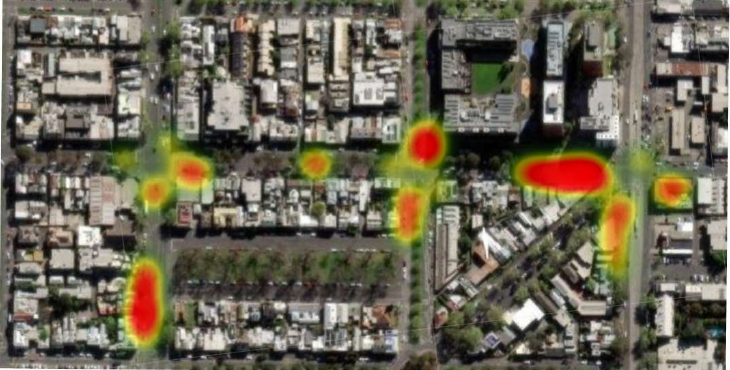}
}
\subfloat[MA2B-DDQN: 600 s]{
    \includegraphics[width=0.24\textwidth]{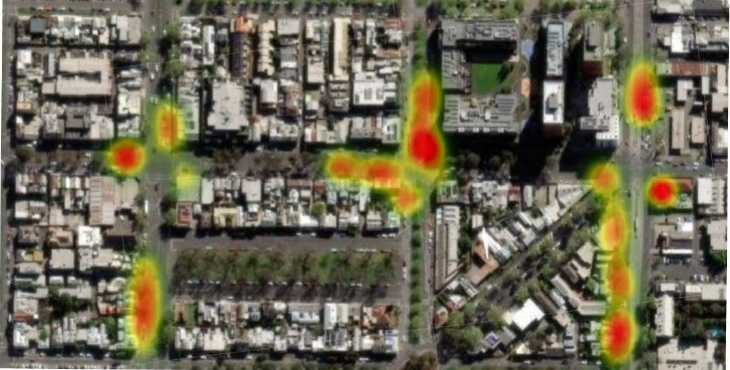}
}
\subfloat[MA2B-DDQN: 800 s]{
    \includegraphics[width=0.24\textwidth]{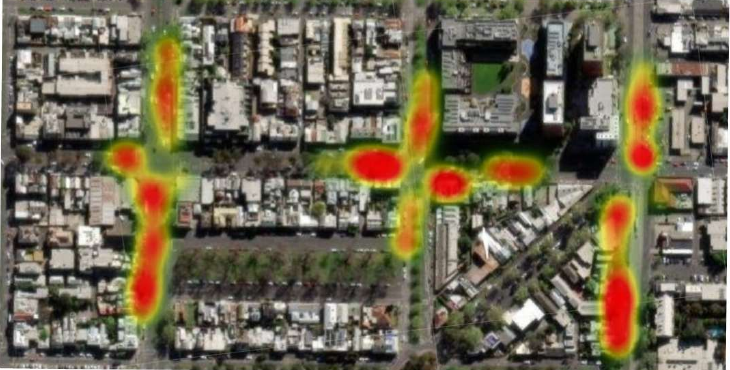}
}
\caption{Heatmaps of delayed vehicles across the corridor network for MA2B-DDQN and baseline models at timestamps 200 s, 400 s, 600 s, and 800 s.}
\label{heatmap_s1}
\end{figure}

Figure \ref{veh_ped_comp_1} illustrates the vehicle and pedestrian delay over time for this scenario across different models. MA2B-DDQN achieved the lowest AVDS at 103.83 vehicles per second, outperforming all baseline models. However, its performance in minimizing pedestrian delay was not the best, recording an APDS of 68.36 pedestrians per second, which was surpassed by MA2B-DQN (63.05 pedestrians/second), MADQN (67.37 pedestrians/second), and CMRM (45.84 pedestrians/second). This discrepancy is likely due to the significantly higher number of vehicle users compared to pedestrians, leading MA2B-DDQN to prioritize vehicle flow optimization. In contrast, fixed-signal methods, including FS-WF and FS-GW, performed poorly in managing pedestrian movements, as Webster’s Formula and green wave control solely focus on vehicle volume while neglecting pedestrian considerations. Figure \ref{heatmap_s1} presents heatmaps of delayed vehicles across the corridor network for different methods. Overall, MA2B-DDQN exhibits the lowest density of delayed vehicles at timestamps 200 s, 400 s, 600 s, and 800 s during the testing period. In contrast, FS-WF and MADDQN show the highest densities of delayed vehicles. These findings align with the AVDS results in Figure \ref{veh_ped_comp_1}(a), where FS-WF and MADDQN record AVDS values of 164.44 and 132.23 vehicles per second, respectively—significantly higher than MA2B-DDQN's AVDS of 103.83 vehicles per second.

\begin{figure}[htbp]
\begin{center}
\includegraphics[width=1.0\textwidth]{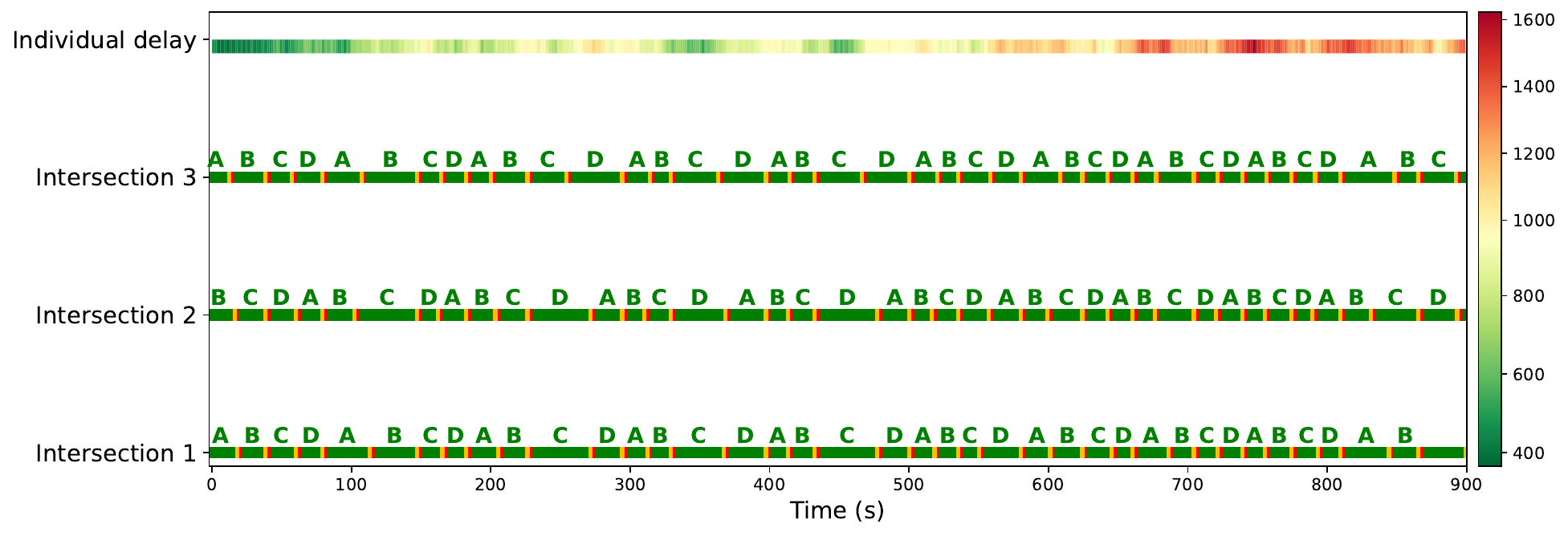}
\caption{The optimal signal state (green/amber/red) for each intersection derived by MA2B-DDQN, along with the number of delayed individuals throughout the testing period for Scenario 1. Intersection 1 is the westernmost intersection of the selected network corridor, while Intersection 3 is the easternmost. The activated phase (phases A, B, C, or D) for each green period is labeled at the top. The number of delayed individuals is visualized using a color-coding system on the right side of the figure, where red indicates a higher number and green represents a lower number.}
\label{signal_state}
\end{center}
\end{figure}

Figure \ref{signal_state} illustrates the optimized signal state derived by MA2B-DDQN for each intersection, along with the corresponding number of delayed individuals throughout the testing period. The adaptive cycle times (total duration of four phases) observed during the testing period are [87 s, 103 s, 112 s, 103 s, 103 s, 80 s, 80 s, 79 s, 70 s, 110 s]. The offsets for intersections 2 and 3 are consistently estimated as 16.2 s and 14.3 s, respectively. The maximum bandwidths are calculated as [62 s, 78 s, 87 s, 78 s, 78 s, 55 s, 55 s, 54 s, 45 s, 85 s]. Notably, the results in Figure \ref{signal_state} do not indicate any extreme green time durations (i.e., excessively long or short green times) for specific phases. Additionally, there is no substantial variation in green time allocation between consecutive cycles, indicating that the proposed model effectively maintains stability and consistency in traffic signal control while optimizing overall traffic flow.

\subsection{Scenario 2}

\begin{table}[htbp]
\centering
\caption{Performance comparison for different models (scenario 2)}
\label{table_s2}
\begin{tabular}{p{2.6cm}p{2.2cm}p{1.9cm}p{1.5cm}p{1.4cm}p{1.5cm}p{1.6cm}p{1.4cm}}
\hline
model & AID ($\mu$) & AID ($\sigma$) & PCD ($\mu$) & PCD ($\sigma$) & ADV (s) & AWTP (s) & IMP \\
\hline
FS-WF & 4,412,134.20 & 76,902.72 & 160.45 & 2.80 & 311.20 & 303.97 & 37.16\%\\
FS-GW & 4,119,712.45 & 68,660.90 & 149.81 & 2.50 & 258.45 & 309.26 & 32.70\%\\
MADQN & 3,211,821.75 & 120,885.61 & 116.80 & 4.40 & 170.49 & 86.39 & 13.67\%\\
MADDQN & 3,249,719.95 & 108,452.97 & 118.18 & 3.94 & 176.38 & 89.58 & 14.68\%\\
MADDPG & 3,332,324.95 & 256,520.40 & 121.18 & 9.33 & 172.37 & 101.92 & 16.80\%\\
CMRM & 3,922,039.90 & 108,185.07 & 142.63 & 3.93 & 238.35 & \textbf{56.71} & 29.31\%\\
MASAC & 3,711,994.40 & 136,199.06 & 134.99 & 4.95 & 198.75 & 70.44 & 25.31\%\\
MAA2C & 3,141,047.95 & 88,431.15 & 114.22 & 3.22 & 198.75 & 78.70 & 11.73\%\\
MA2B-DQN & 3,072,463.95 & 55,324.00 & 111.73 & \textbf{1.83} & 154.13 & 81.20 & 9.76\%\\
\textbf{MA2B-DDQN} & \textbf{2,772,647.95} & \textbf{53,629.81} & \textbf{100.83} & 1.95 & \textbf{133.71} & 71.99 & \textit{NA}\\
\hline
\end{tabular}
\end{table}

\begin{figure}[htbp]
\begin{center}
\includegraphics[width=0.95\textwidth]{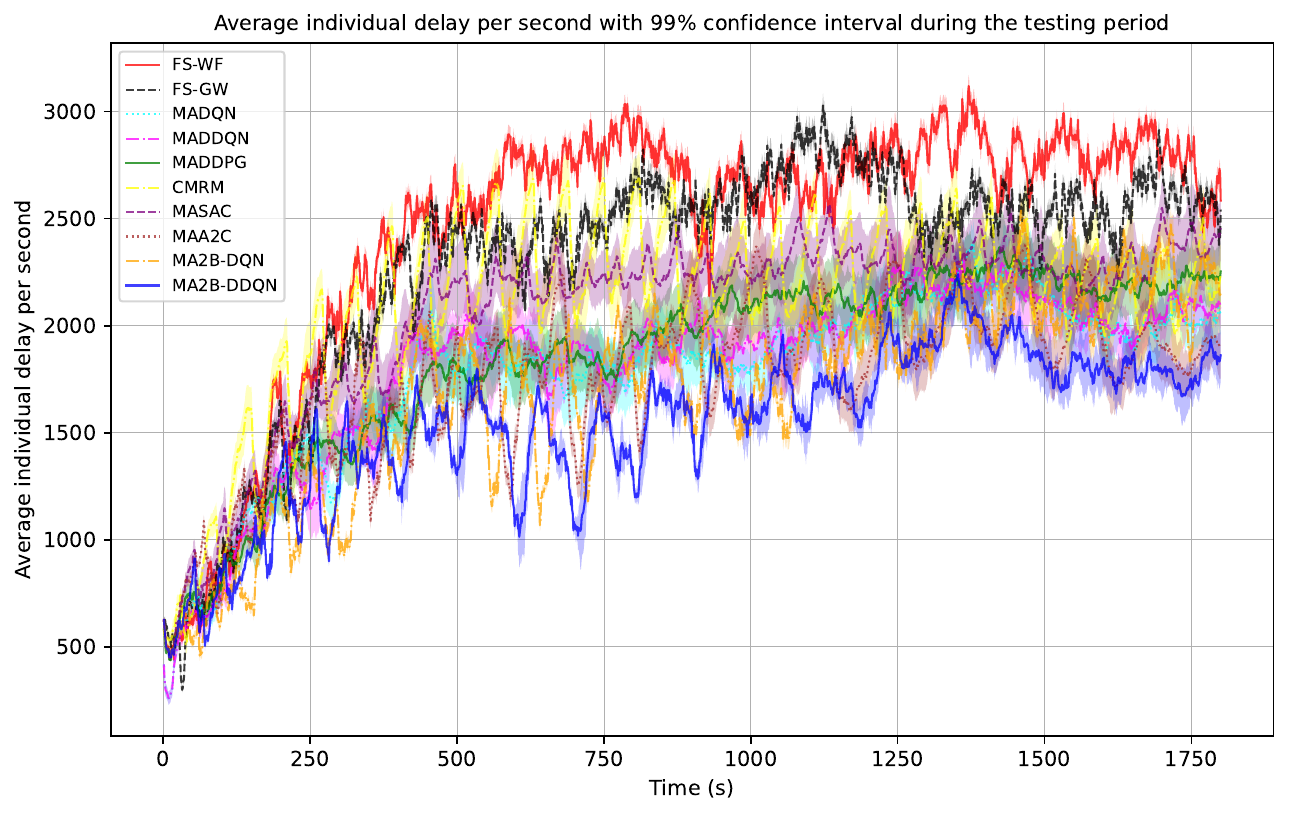}
\caption{Comparison of the individual delay and 99\% confidence interval throughout the testing period for scenario 2.}
\label{cost_comp_2}
\end{center}
\end{figure}

Table \ref{table_s2} provides the results for Scenario 2, describing rush hour traffic demands. In this scenario, MA2B-DDQN achieves the top performance with an AID of 2,772,647.95 impacted travelers, outperforming all other DRL and fixed signal models by at least 11.73\% (compared to MAA2C). 
The average delay per vehicle is also the lowest for MA2B-DDQN, with 133.71 seconds per vehicle, significantly outperforming the next-best baseline model, which records 154.13 seconds per vehicle by MA2B-DQN. However, its average pedestrian waiting time is slightly higher than that of the top-performing baseline, MASAC, at 71.99 seconds compared to MASAC’s 70.44 seconds.
Additionally, MA2B-DDQN improves performance by 9.76\% over MA2B-DQN in this scenario. Although MA2B-DQN shows slightly more robust PCD across various test cases, both MA2B-DDQN and MA2B-DQN demonstrate highly stable performance, indicated by their considerably lower standard deviations (1.83 vs. 1.95 in terms of PCD ($\sigma$)) compared to other baselines. FS-GW shows a substantial improvement over FS-WF (32.70\% vs. 37.16\%), likely due to FS-WF's lack of coordination between intersections. Similarly, CMRM records the lowest AWTP (56.71 s), but it also incurs substantial vehicle delay and a high AID, suggesting an uneven prioritization between modalities. In this scenario, MAA2C is the most competitive baseline model, followed by MADQN and MADDQN. All DRL models in this scenario yield lower AID and PCD than the fixed signal models (FS-WF and FS-GW), emphasizing the effectiveness and potential of DRL in adaptive traffic signal control.

Figure \ref{cost_comp_2} depicts the average individual delay across the entire testing period. As illustrated, MA2B-DDQN maintains stable and reliable performance, with consistently low values throughout the whole period. The 99\% confidence interval for MADDPG is the widest, reflecting the high standard deviations of MADDQN shown in Table \ref{table_s2}, further confirming MADDPG’s sensitivity in multi-agent TSC optimization problems.

\begin{figure}[htbp]
\begin{center}
\subfloat[]{
    \includegraphics[width=0.85\textwidth]{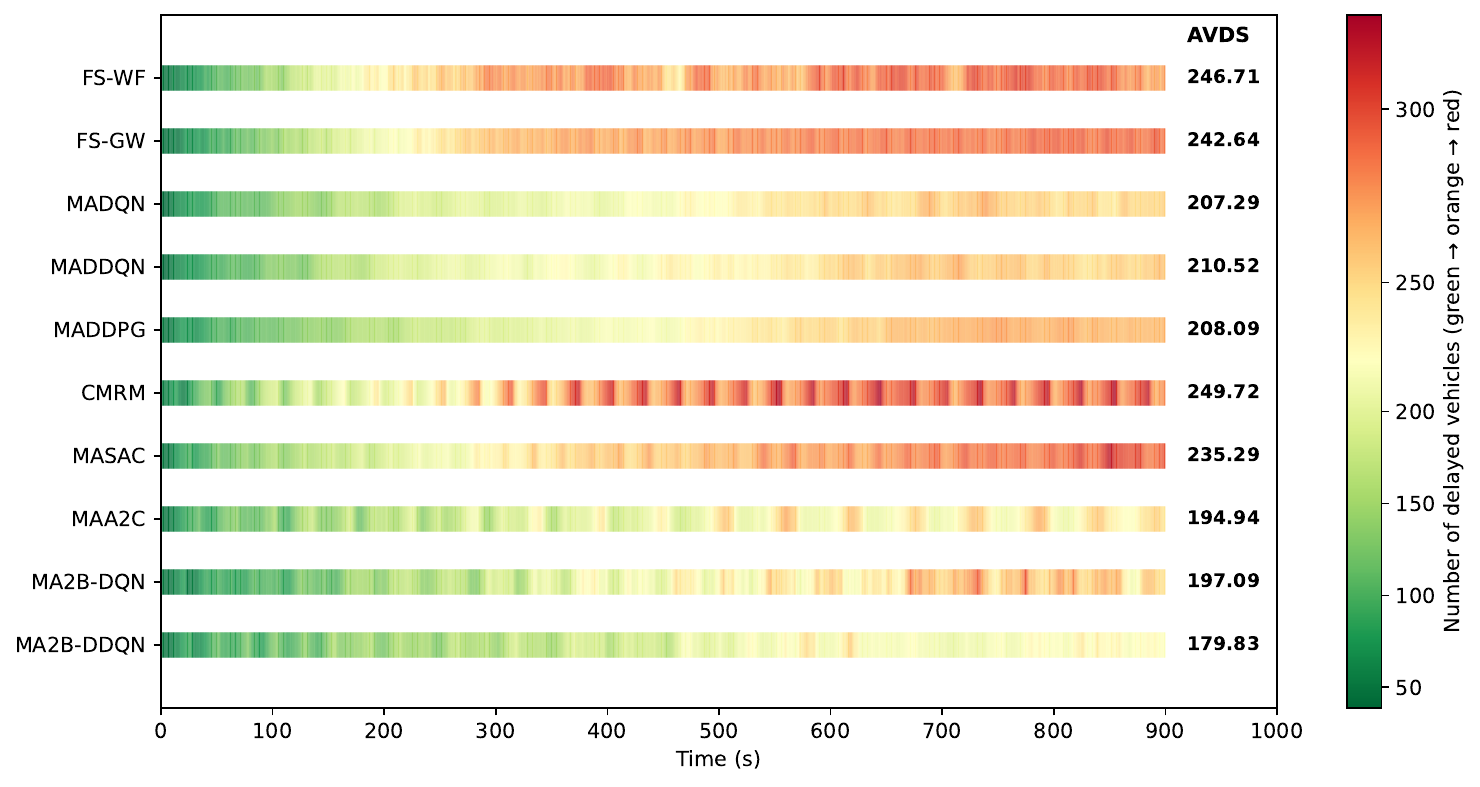}
    }
    
\subfloat[]{
    \includegraphics[width=0.85\textwidth]{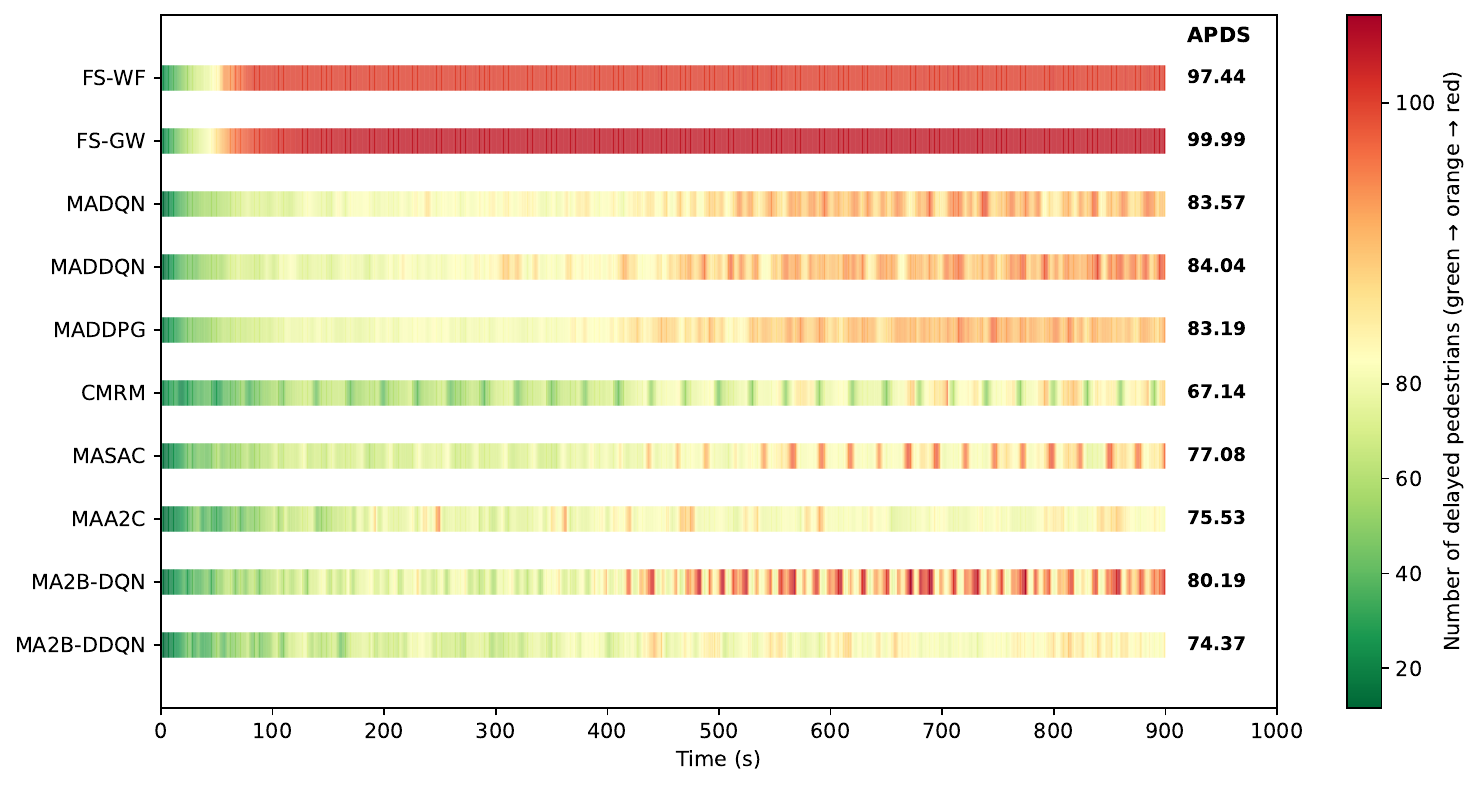}
    }
\caption{Analysis of different models in Scenario 2. (a) Vehicle delay over time (unit: vehicles per second). (b) Pedestrian delay over time (unit: pedestrians per second).}
\label{veh_ped_comp_2}
\end{center}
\end{figure}

Figure \ref{veh_ped_comp_2} illustrates the temporal delay distribution for vehicles (Figure \ref{veh_ped_comp_2}(a)) and pedestrians (Figure \ref{veh_ped_comp_2}(b)). MA2B-DDQN achieves an AVDS of 179.83 and an APDS of 74.37, outperforming all other baseline methods; the only exception is CMRM, which attains a lower APDS of 67.14.
In contrast, FS-WF and FS-GW record higher APDS values of 97.44 and 99.99, respectively, as both prioritize vehicle volume over pedestrian volume when optimizing signal plans. Among the competing DRL-based methods, MAA2C ranks as the second-best, achieving an AVDS of 194.94 and an APDS of 75.53, demonstrating strong performance in both vehicle and pedestrian optimization.

\subsection{Scenario 3}

\begin{table}[htbp]
\centering
\caption{Performance comparison for different models (scenario 3)}
\label{table_s3}
\begin{tabular}{p{2.6cm}p{2.2cm}p{1.9cm}p{1.5cm}p{1.4cm}p{1.5cm}p{1.6cm}p{1.4cm}}
\hline
model & AID ($\mu$) & AID ($\sigma$) & PCD ($\mu$) & PCD ($\sigma$) & ADV (s) & AWTP (s) & IMP \\
\hline
FS-WF & 2,926,135.05 & 40,907.89 & 150.14 & 2.10 & 211.92 & 205.13 & 37.06\%\\
FS-GW & 2,502,719.30 & 36,587.61 &	128.42 & 1.88 & 173.33 & 351.28 & 26.41\%\\
MADQN & 2,359,815.25 & 90,749.30 &	121.08 & 4.66 & 150.86 & 91.81 & 21.96\%\\
MADDQN & 2,415,213.60 & 94,783.20 & 123.93 & 4.86 & 157.95 & 92.47 & 23.75\%\\
MADDPG & 2,643,811.80 & 387,170.66 & 135.66 & 19.87 & 182.26 & 115.65 & 30.34\%\\
CMRM & 2,942,060.90 & 103,845.67 & 150.96 & 5.33 & 206.74 & \textbf{55.99} & 37.40\%\\
MASAC & 2,793,317.25 & 116,759.26 & 143.33 & 5.99 & 179.51 & 71.93 & 34.07\%\\
MAA2C & 2,139,522.40 & 71,304.60 & 109.78 & 3.66 & 131.73 & 78.38 & 13.92\%\\
MA2B-DQN & 2,203,075.70 & 39,715.09 & 113.04 & 2.04 & 132.23 & 89.31 & 16.40\%\\
\textbf{MA2B-DDQN} & \textbf{1,841,687.10} & \textbf{29,881.33} & \textbf{94.50} & \textbf{1.53} & \textbf{111.44} & 78.06 & \textit{NA}\\
\hline
\end{tabular}
\end{table}

\begin{figure}[htbp]
\begin{center}
\includegraphics[width=0.95\textwidth]{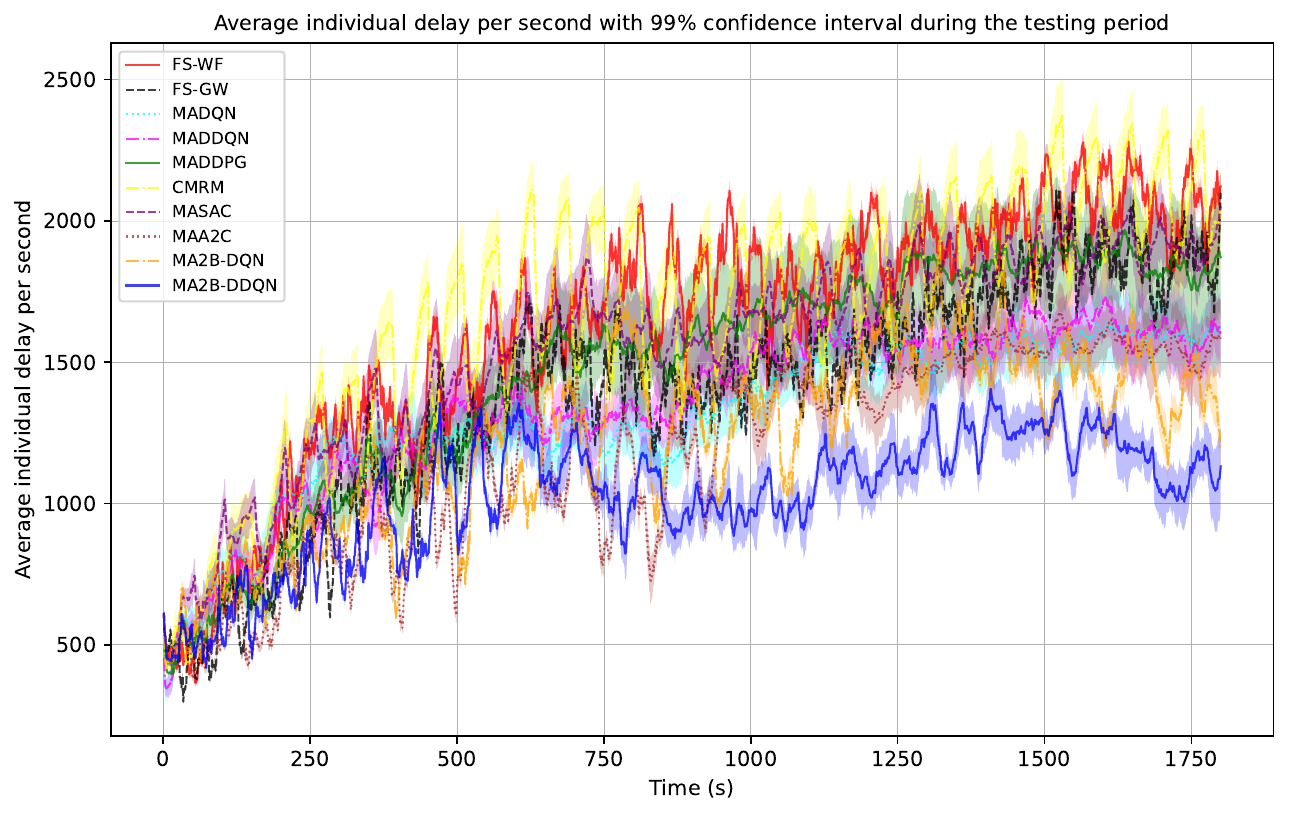}
\caption{Comparison of the individual delay and 99\% confidence interval throughout the testing period for scenario 3.}
\label{cost_comp_3}
\end{center}
\end{figure}

\begin{figure}[htbp]
\begin{center}
\subfloat[]{
    \includegraphics[width=0.85\textwidth]{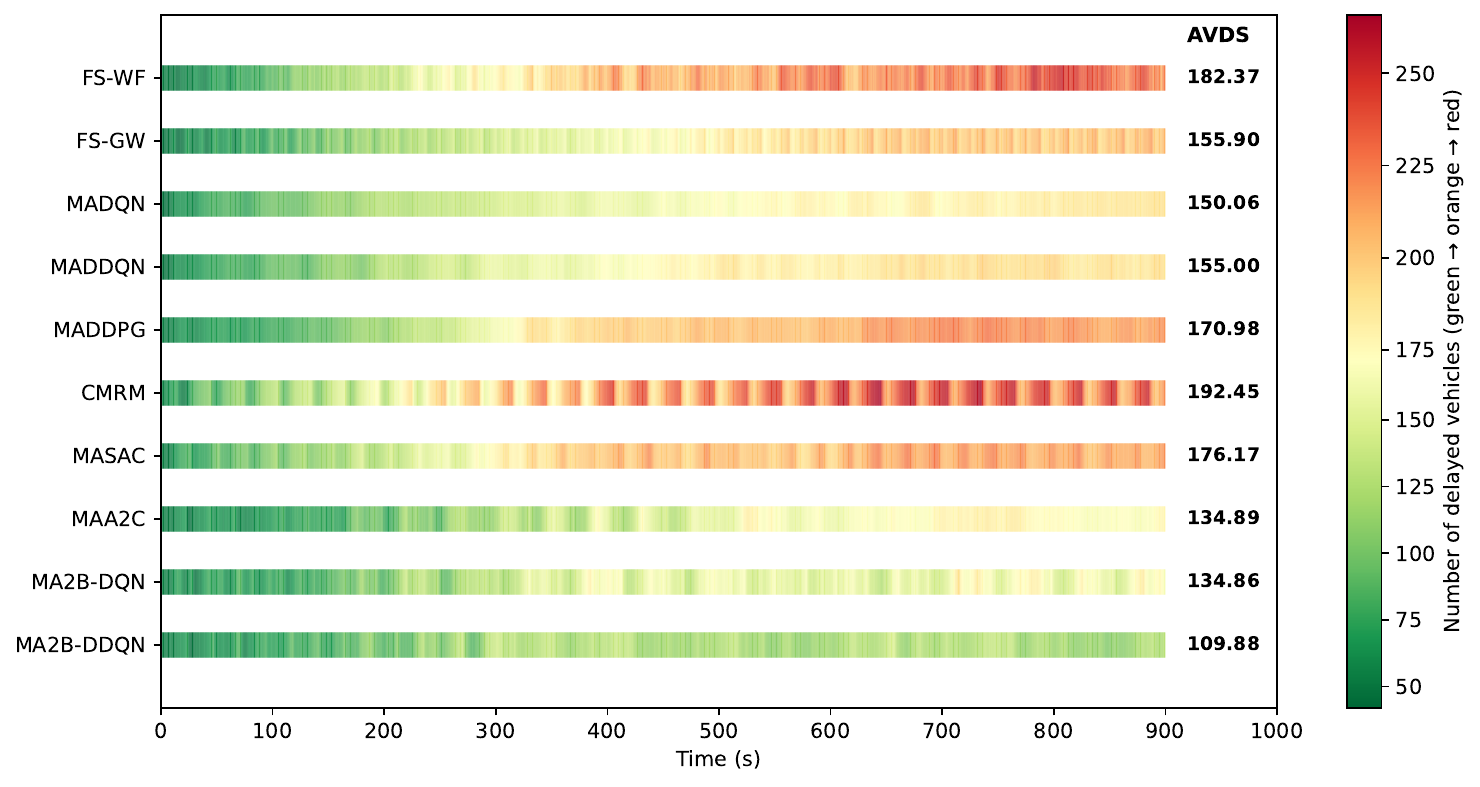}
    }
    
\subfloat[]{
    \includegraphics[width=0.85\textwidth]{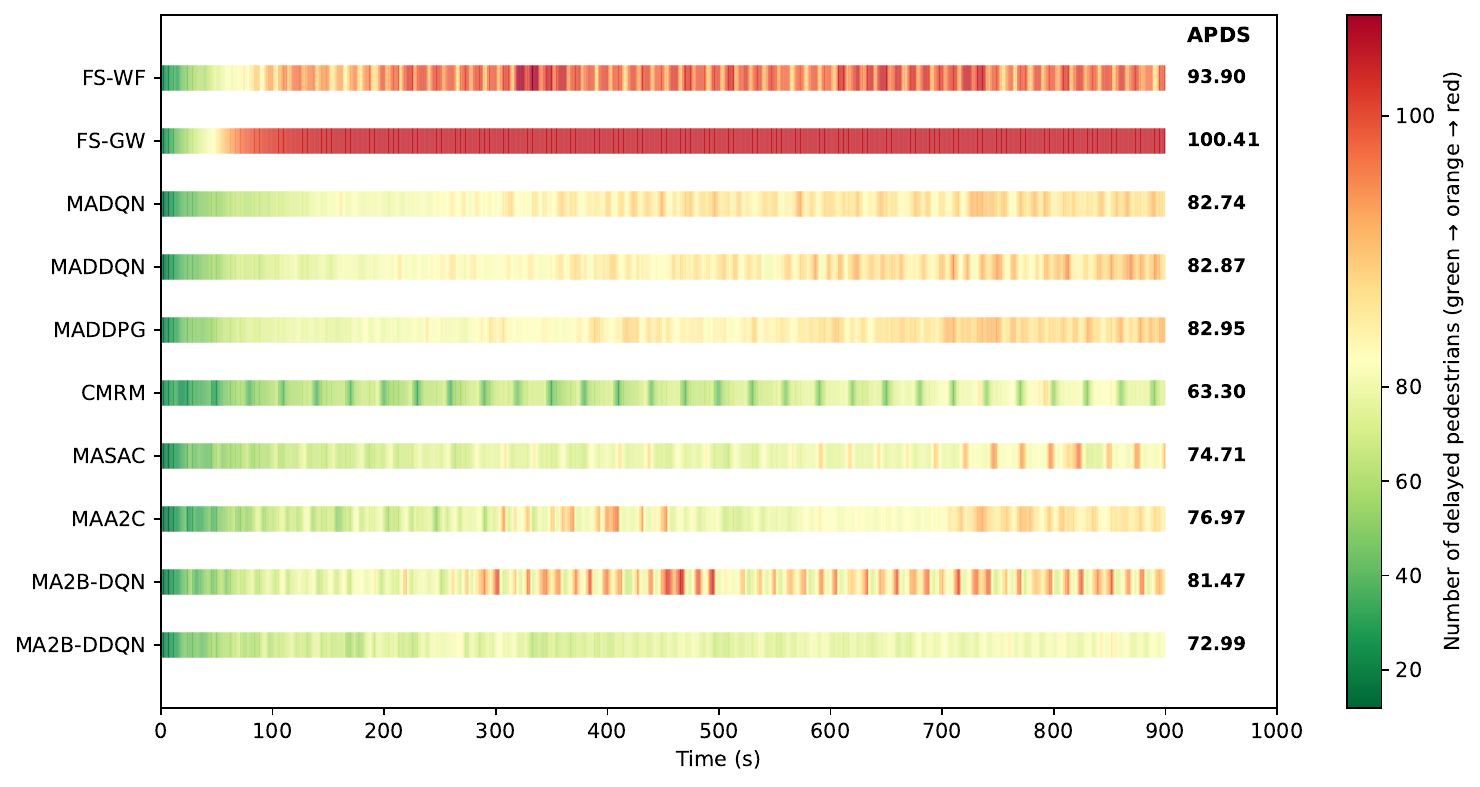}
    }
\caption{Analysis of different models in Scenario 3. (a) Vehicle delay over time (unit: vehicles per second). (b) Pedestrian delay over time (unit: pedestrians per second).}
\label{veh_ped_comp_3}
\end{center}
\end{figure}

Scenario 3 simulates morning school hours with relatively heavy traffic, where increasing traffic demand is directed toward a certain zone within the road network. Table \ref{table_s3} presents the performance comparison for this scenario, indicating that MA2B-DDQN outperforms all baseline methods in both AID and PCD, as well as in standard deviation metrics. Additionally, MA2B-DDQN records the shortest average delay per vehicle, achieving an ADV of 111.44 seconds per vehicle. Similar to Scenario 2, its average waiting time per pedestrian is slightly higher than that of MASAC (78.06 vs. 71.93 seconds), reflecting an imbalance in optimization between vehicles and pedestrians. This imbalance is even more pronounced for CMRM.
In this scenario, MA2B-DDQN shows a substantial improvement over other DRL baselines, with at least a 13.92\% enhancement. Incorporating DDQN into MA2B-DQN further significantly boosts its performance by 16.40\%.

Figure \ref{cost_comp_3} illustrates the average individual delay of various methods over the independent testing period. MA2B-DDQN achieves the lowest individual delay in the later stages of testing, while both MA2B-DDQN and MA2B-DQN maintain very low values initially. Notably, MASAC performs unfavorably at the start of the testing period, with its values then dropping significantly, resulting in relatively weaker overall performance for MASAC. This may be due to SAC's sensitivity to initial network conditions or state distributions \citep{haarnoja2018soft}.

Figure \ref{veh_ped_comp_3} compares the vehicle delay and pedestrian delay over the entire testing period. MA2B-DDQN achieves a low AVDS of 109.88, indicating that, on average, 109.88 vehicles experience delays per operational second. This surpasses its closest competitor, MA2B-DQN, which records an AVDS of 134.86 vehicles per second. MA2B-DDQN also attains the second-lowest APDS at 72.99, meaning an average of 72.99 pedestrians are delayed per second, although it trails the best-performing baseline, CMRM, which achieves an APDS of 63.30 pedestrians per second. FS-GW is not capable of managing pedestrian movements, as it places too much emphasis on coordinating vehicle flow. These results further highlight MA2B-DDQN’s effectiveness in optimizing both vehicle and pedestrian traffic flow.

\subsection{Scenario 4}

\begin{table}[htbp]
\centering
\caption{Performance comparison for different models (scenario 4)}
\label{table_s4}
\begin{tabular}{p{2.6cm}p{2.2cm}p{1.9cm}p{1.5cm}p{1.4cm}p{1.5cm}p{1.6cm}p{1.4cm}}
\hline
model & AID ($\mu$) & AID ($\sigma$) & PCD ($\mu$) & PCD ($\sigma$) & ADV (s) & AWTP (s) & IMP \\
\hline
FS-WF & 3,440,078.70 & 65,761.93 & 184.38 & 3.52 & 257.13 & 348.96 & 61.30\%\\
FS-GW & 2,903,712.95 & 60,871.72 & 155.63 & 3.26 & 199.26 & 354.92 & 54.15\%\\
MADQN & 1,762,628.80 & 173,550.88 & 94.47 & 9.30 & 119.02 & 87.83 & 24.47\%\\
MADDQN & 1,809,260.20 & 179,820.75 & 96.97 & 9.64 & 122.12 & 91.61 & 26.42\%\\
MADDPG & 1,778,341.45 & 254,832.57 & 95.31 & 13.66 & 117.14 & 104.75 & 25.14\%\\
CMRM & 2,393,935.80 & 234,388.20 & 128.31 & 12.56 & 165.73 & \textbf{55.81} & 44.39\%\\
MASAC & 2,055,677.20 & 189,304.65 & 110.18 & 10.15 & 134.93 & 73.22 & 35.24\%\\
MAA2C & 1,609,647.60 & 58,088.72 & 86.27 & 3.11 & 102.37 & 97.01 & 17.30\%\\
MA2B-DQN & 1,370,579.60 & 23,325.78 & 73.46 & 1.25 & 85.43 & 124.71 & 2.87\%\\
\textbf{MA2B-DDQN} & \textbf{1,331,248.75} & \textbf{16,270.70} & \textbf{71.35} & \textbf{0.87} & \textbf{79.38} & 85.82 & \textit{NA}\\
\hline
\end{tabular}
\end{table}

\begin{figure}[htbp]
\begin{center}
\includegraphics[width=0.95\textwidth]{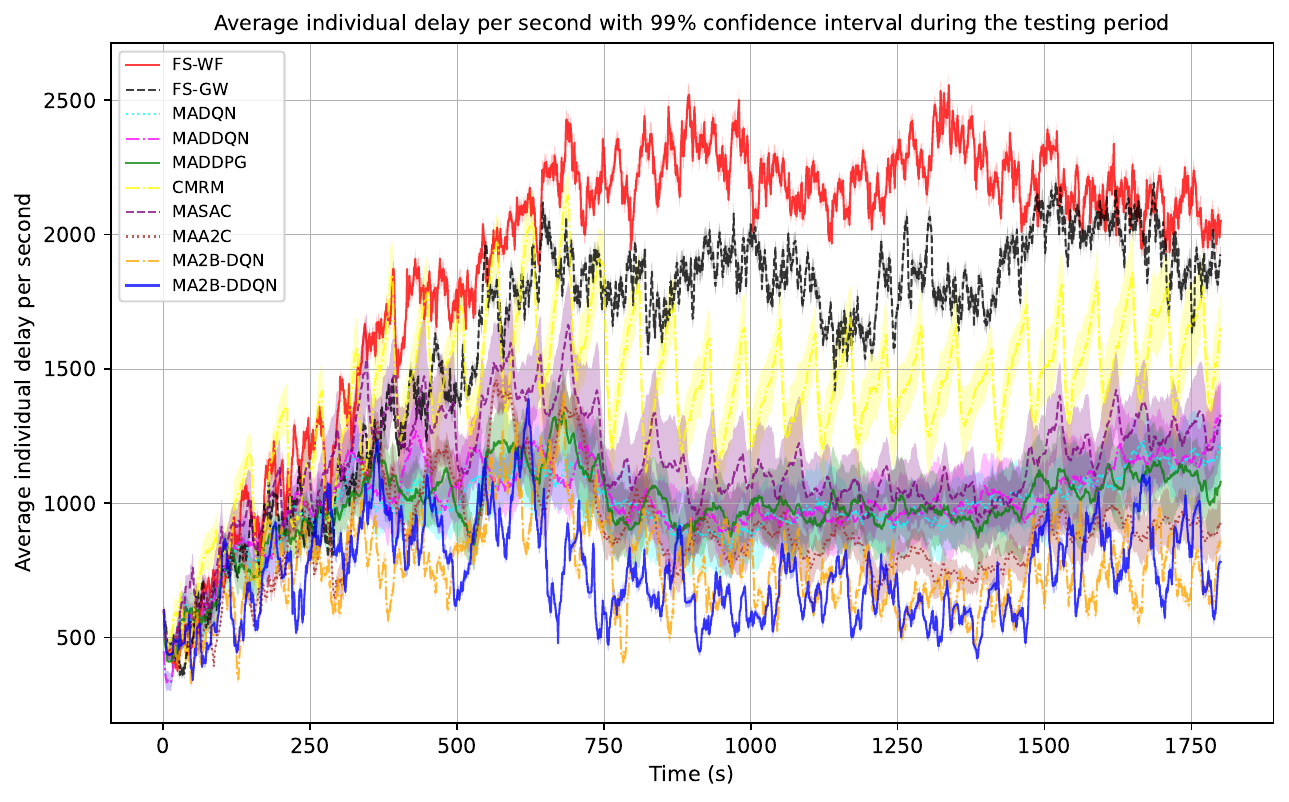}
\caption{Comparison of the individual delay and 99\% confidence interval throughout the testing period for scenario 4.}
\label{cost_comp_4}
\end{center}
\end{figure}

\begin{figure}[htbp]
\begin{center}
\subfloat[]{
    \includegraphics[width=0.85\textwidth]{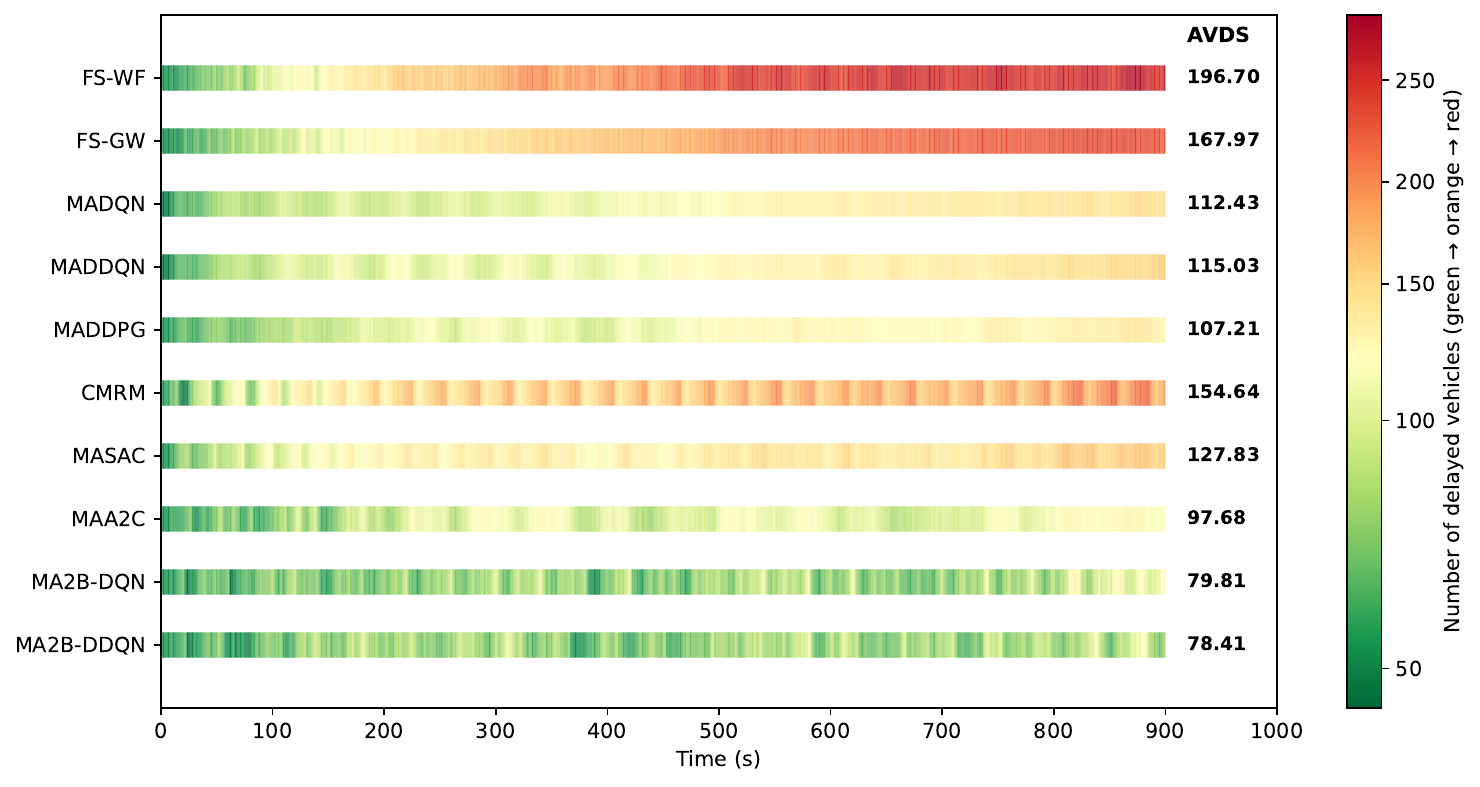}
    }
    
\subfloat[]{
    \includegraphics[width=0.85\textwidth]{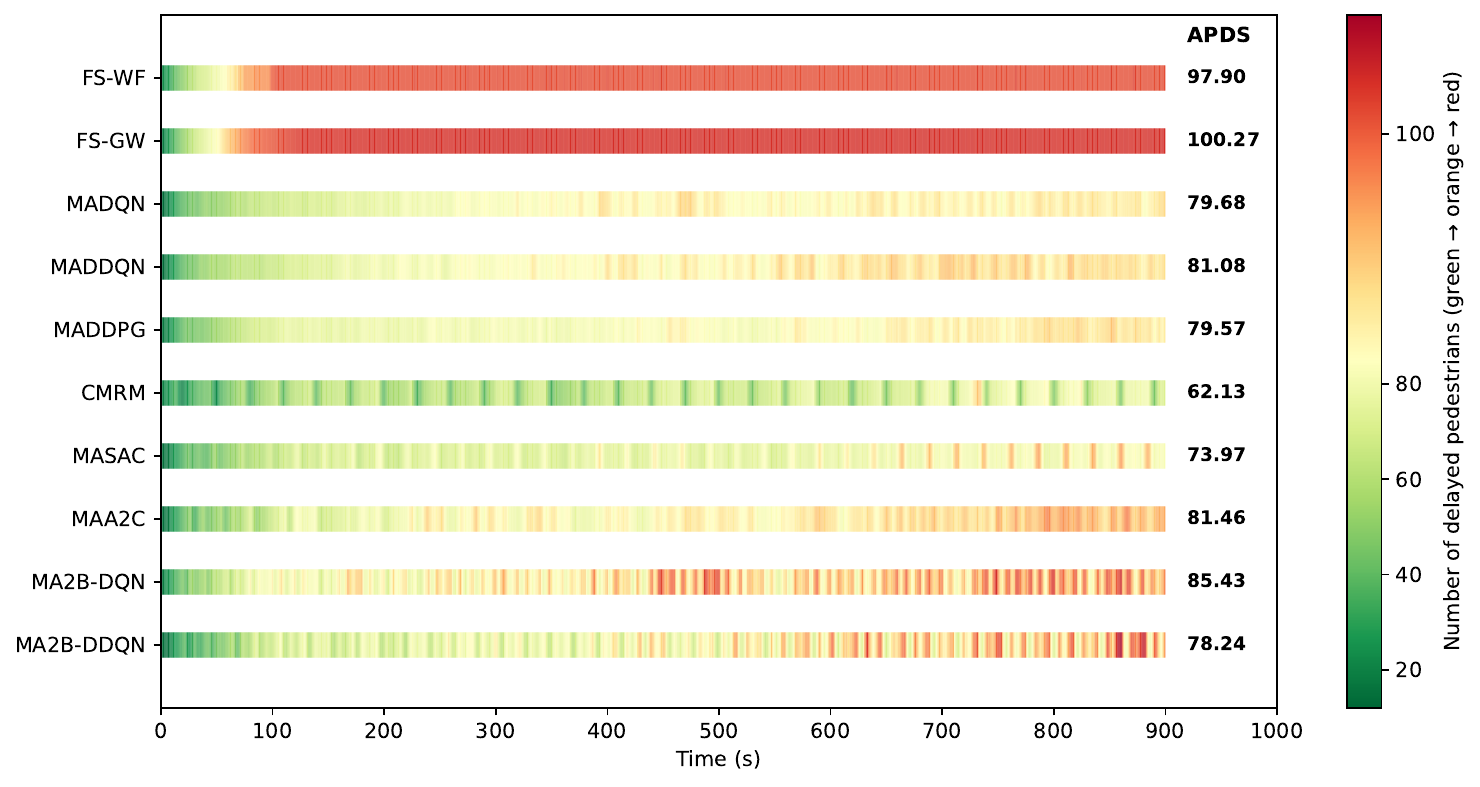}
    }
\caption{Analysis of different models in Scenario 4. (a) Vehicle delay over time (unit: vehicles per second). (b) Pedestrian delay over time (unit: pedestrians per second).}
\label{veh_ped_comp_4}
\end{center}
\end{figure}

In Scenario 4, we model afternoon school hours with moderate traffic conditions. Table \ref{table_s4} displays the performance of all methods. MA2B-DDQN achieves an AID of 1,331,248.75 individuals, a PCD of 71.35, an ADV of 79.38 seconds, and an AWTP of 85.82 seconds. As in Scenario 3, MA2B-DDQN attains the lowest mean values as well as standard deviations, indicating its superiority and stability in multi-agent TSC optimization, with a standard deviation of 0.87 for PCD. FS-WF and FS-GW are the poorest-performing methods in this scenario; however, integrating green wave control technology improves the fixed signal’s performance by over 7\%. MASAC achieves the shortest pedestrian delay, demonstrating its emphasis on pedestrian flow management. However, this comes at the cost of significantly higher vehicle delays, with an average of 134.93 seconds per vehicle, highlighting an imbalance between vehicle and pedestrian optimization.

The overall individual delays of different methods over the testing period for this scenario have been illustrated in Figure \ref{cost_comp_4}. It can be observed that FS-WF and FS-GW incur the highest value of individual delay once the simulation time exceeds approximately 400 seconds. In contrast, MA2B-DDQN maintains a consistently minimal delay throughout the entire testing period.

Figure \ref{veh_ped_comp_4} illustrates the delay trends over time for both vehicles and pedestrians. Throughout the testing period, vehicle delay per second remains consistently low for both MA2B-DQN and MA2B-DDQN, demonstrating the effectiveness of the action branching mechanism in DRL. For pedestrian delays, CMRM and MASAC achieve better performance than MA2B-DDQN, with an APDS of 73.97 and 62.13 pedestrians per second compared to 78.24 pedestrians per second. However, MASAC and CMRM’s AVDS are significantly higher at 127.83 and 154.64 vehicles per second, whereas MA2B-DDQN maintains a significantly lower AVDS of 78.41 vehicles/second. As a result, MA2B-DDQN achieves a substantially lower aggregate individual delay of 1,331,248.75 compared to 2,055,677.20 achieved by MASAC and 2,393,935.80 achieved by CMRM, reflecting a 35.24\% and a 44.39\% improvement in performance.

\subsection{Scenario 5}

\begin{table}[htbp]
\centering
\caption{Performance comparison for different models (scenario 5)}
\label{table_s5}
\begin{tabular}{p{2.6cm}p{2.2cm}p{1.9cm}p{1.5cm}p{1.4cm}p{1.5cm}p{1.6cm}p{1.4cm}}
\hline
model & AID ($\mu$) & AID ($\sigma$) & PCD ($\mu$) & PCD ($\sigma$) & ADV (s) & AWTP (s) & IMP \\
\hline
FS-WF & 4,042,961.60 & 80,113.63 & 123.22 & 2.44 & 335.25 & 272.04 & 18.41\%\\
FS-GW & 3,972,120.15 & 76,212.32 & 121.06 & 2.32 & 306.02 & 603.59 & 16.95\%\\
MADQN & 3,676,373.70 & 146,196.11 & 112.04 & 4.46 & 225.42 & 73.98 & 10.27\%\\
MADDQN & 3,719,111.55 & 146,862.75 & 113.35 & 4.48 & 230.46 & 67.01 & 11.30\%\\
CMRM & 3,700,549.35 & 161,539.85 & 112.78 & 4.92 & 252.91 & 43.57 & 10.86\%\\
MADDPG & 3,616,426.50 & 245,287.80 & 110.22 & 7.48 & 224.09 & 110.23 & 8.78\%\\
MASAC & 3,548,403.30 & 84,515.78 & 108.14 & 2.58 & 222.14 & 67.42 & 7.03\%\\
MAA2C & 3,392,125.65 & 55,836.71 & 103.38 & 1.70 & 217.63 & \textbf{42.01} & 2.75\%\\
MA2B-DQN & 3,592,312.60 & 87,996.72 & 109.48 & 2.68 & 223.76 & 267.49 & 8.17\%\\
\textbf{MA2B-DDQN} & \textbf{3,298,780.95} & \textbf{43,617.94} & \textbf{100.54} & \textbf{1.33} & \textbf{191.84} & 63.41 & \textit{NA}\\
\hline
\end{tabular}
\end{table}

\begin{figure}[htbp]
\begin{center}
\includegraphics[width=0.95\textwidth]{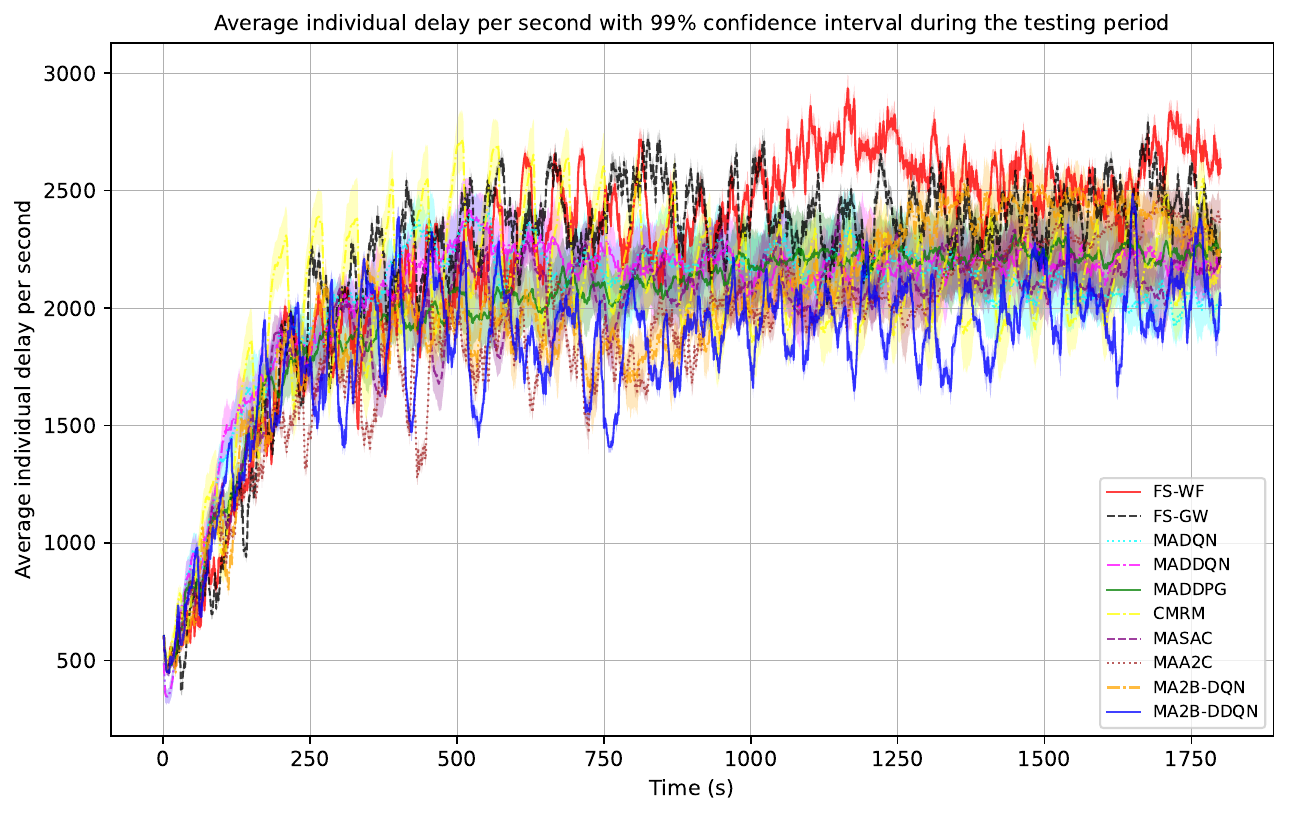}
\caption{Comparison of the individual delay and 99\% confidence interval throughout the testing period for scenario 5.}
\label{cost_comp_5}
\end{center}
\end{figure}

\begin{figure}[htbp]
\begin{center}
\subfloat[]{
    \includegraphics[width=0.85\textwidth]{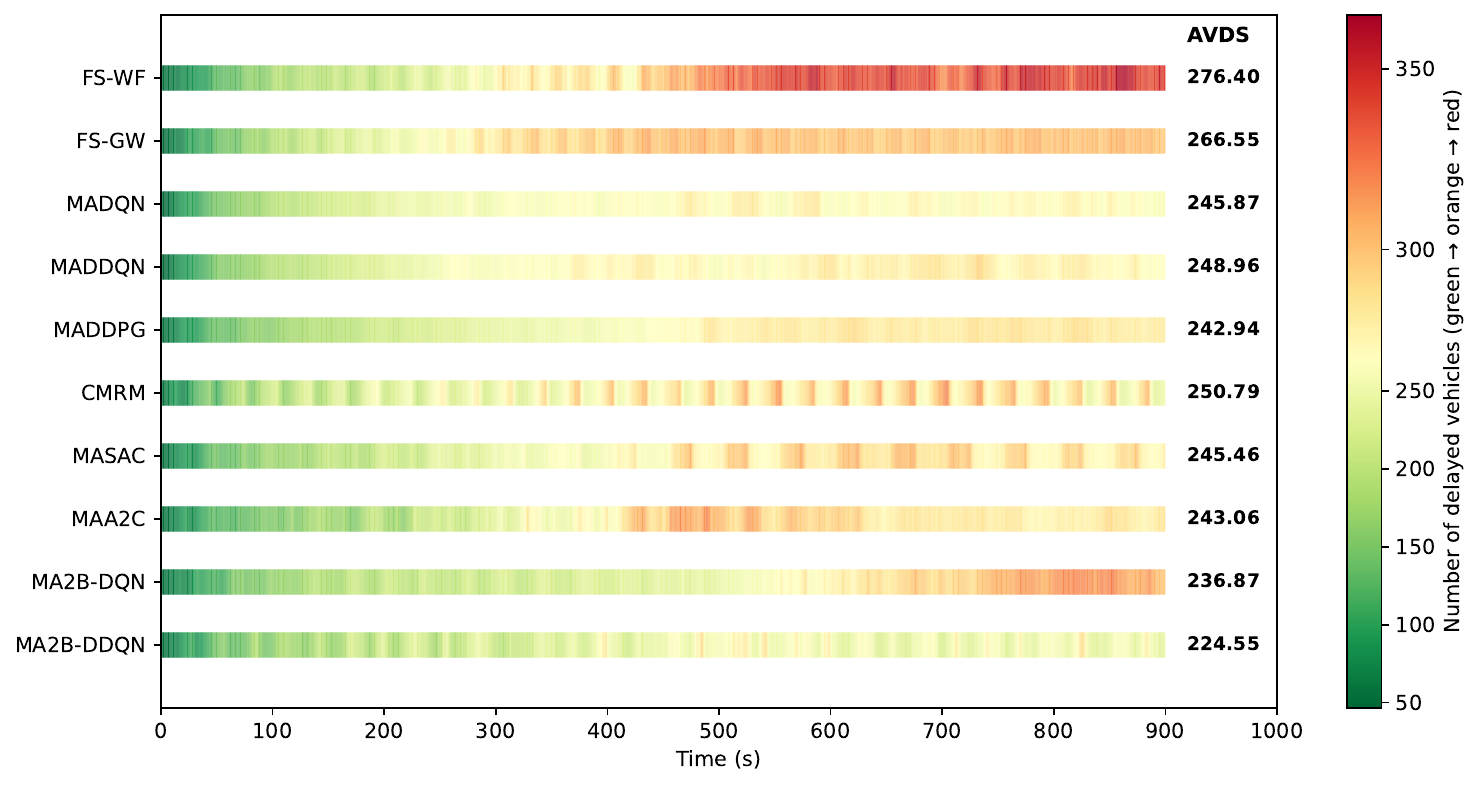}
    }
    
\subfloat[]{
    \includegraphics[width=0.85\textwidth]{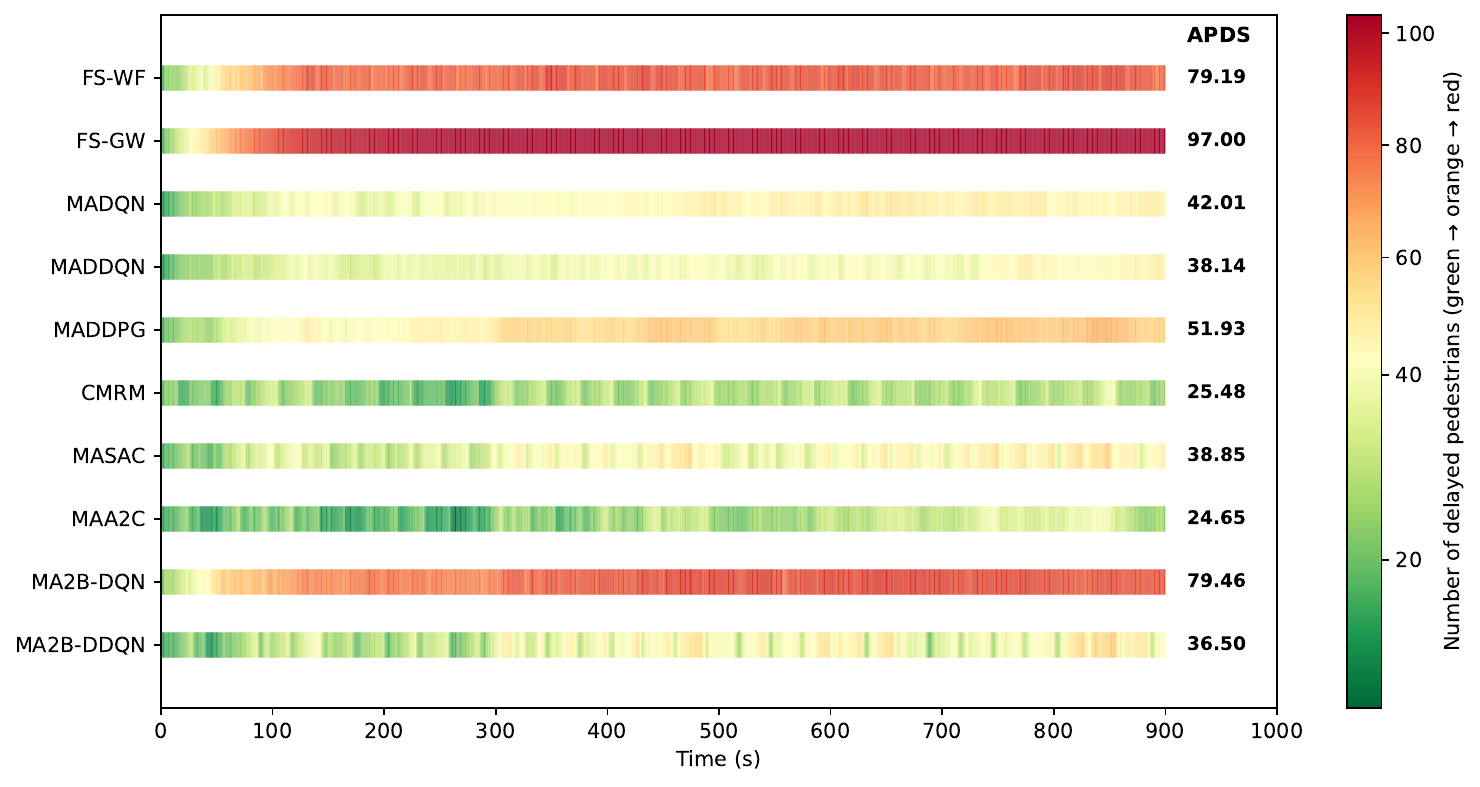}
    }
\caption{Analysis of different models in Scenario 5. (a) Vehicle delay over time (unit: vehicles per second). (b) Pedestrian delay over time (unit: pedestrians per second).}
\label{veh_ped_comp_5}
\end{center}
\end{figure}

In this scenario, traffic demand is increased to approach the saturation flow of the entire corridor network, providing a rigorous test of our method’s performance under near-saturation conditions. Table \ref{table_s5} presents a comparison of MA2B-DDQN against baseline methods. As observed in previous scenarios, MA2B-DDQN achieves the lowest average values and standard deviations across most metrics, except for AWTP. However, the performance differences between MA2B-DDQN and the other baseline methods are less pronounced in this scenario. 
This is because, once the approach is near saturation, the saturation flow caps how many vehicles can be discharged. In that regime, signal coordination has limited leverage, and physical lane capacity becomes the dominant constraint rather than signal timing.
MA2B-DDQN outperforms MAA2C by a modest margin of 2.75\%, while MADDPG, MASAC, and MA2B-DQN exhibit similar performance, with differences remaining under 9\%.
Among the baselines, MAA2C records an ADV of 217.63 seconds per vehicle and achieves the lowest AWTP at 42.01 seconds per pedestrian, demonstrating its strength in pedestrian delay management.

Figure \ref{cost_comp_5} displays the individual delays of different methods over the testing period for this scenario. Overall, MA2B-DDQN maintains a consistently low number throughout the entire period. However, there are specific intervals where other methods achieve lower values. For example, from timestep 1300 to 1600, MADQN has the lowest value, and in the initial phase, MAA2C sometimes achieves the lowest value.

Figure \ref{veh_ped_comp_5} illustrates the temporal trends of vehicle and pedestrian delays throughout the testing period. MA2B-DDQN consistently maintains low vehicle delays, highlighting its robustness in handling near-saturation traffic conditions.
For pedestrian delays, MA2B-DQN records a high APDS of 79.46 pedestrians per second, indicating unstable performance in multimodal optimization. By comparison, MA2B-DDQN substantially reduces APDS to 36.50 pedestrians per second, reflecting improved pedestrian flow management. Nonetheless, in this scenario it is still outperformed by MAA2C and CMRM, which achieve lower APDS values of 24.65 and 25.48 pedestrians per second, respectively.

\subsection{Scenario 6}

\begin{table}[htbp]
\centering
\caption{Performance comparison for different models (scenario 6)}
\label{table_s6}
\begin{tabular}{p{2.6cm}p{2.2cm}p{1.9cm}p{1.5cm}p{1.4cm}p{1.5cm}p{1.6cm}p{1.4cm}}
\hline
model & AID ($\mu$) & AID ($\sigma$) & PCD ($\mu$) & PCD ($\sigma$) & ADV (s) & AWTP (s) & IMP \\
\hline
FS-WF & 8,509,615.30 & 122,465.15 & 163.07 & 2.35 & 436.66 & 251.56 & 30.08\%\\
FS-GW & 7,326,417.85 & 106,852.40 & 140.39 & 2.05 & 298.13 & 501.47 & 18.79\%\\
MADQN & 6,678,458.90 & 168,412.67 & 127.98 & 3.23 & 230.38 & 96.28 & 10.91\%\\
MADDQN & 7,004,117.60 & 174,211.20 & 134.22 & 3.34 & 246.38 & 88.53 & 15.05\%\\
MADDPG & 6,830,951.20 & 534,581.05 & 130.90 & 10.24 & 251.45 & 118.32 & 12.90\%\\
CMRM & 7,389,267.55 & 221,605.67 & 141.60 & 4.25 & 298.98 & \textbf{47.74} & 19.48\%\\
MASAC & 7,333,523.95 & 196,241.58 & 140.53 & 3.76 & 261.85 & 73.08 & 18.87\%\\
MAA2C & 6,626,140.20 & 137,520.58 & 126.97 & 2.64 & 217.79 & 90.84 & 10.20\%\\
MA2B-DQN & 6,148,863.70 & 67,732.67 & 117.83 & 1.30 & 207.47 & 69.03 & 3.23\%\\
\textbf{MA2B-DDQN} & \textbf{5,950,030.20} & \textbf{49,227.07} & \textbf{114.02} & \textbf{0.94} & \textbf{196.56} & 56.18 & \textit{NA}\\
\hline
\end{tabular}
\end{table}

\begin{figure}[htbp]
\begin{center}
\includegraphics[width=0.95\textwidth]{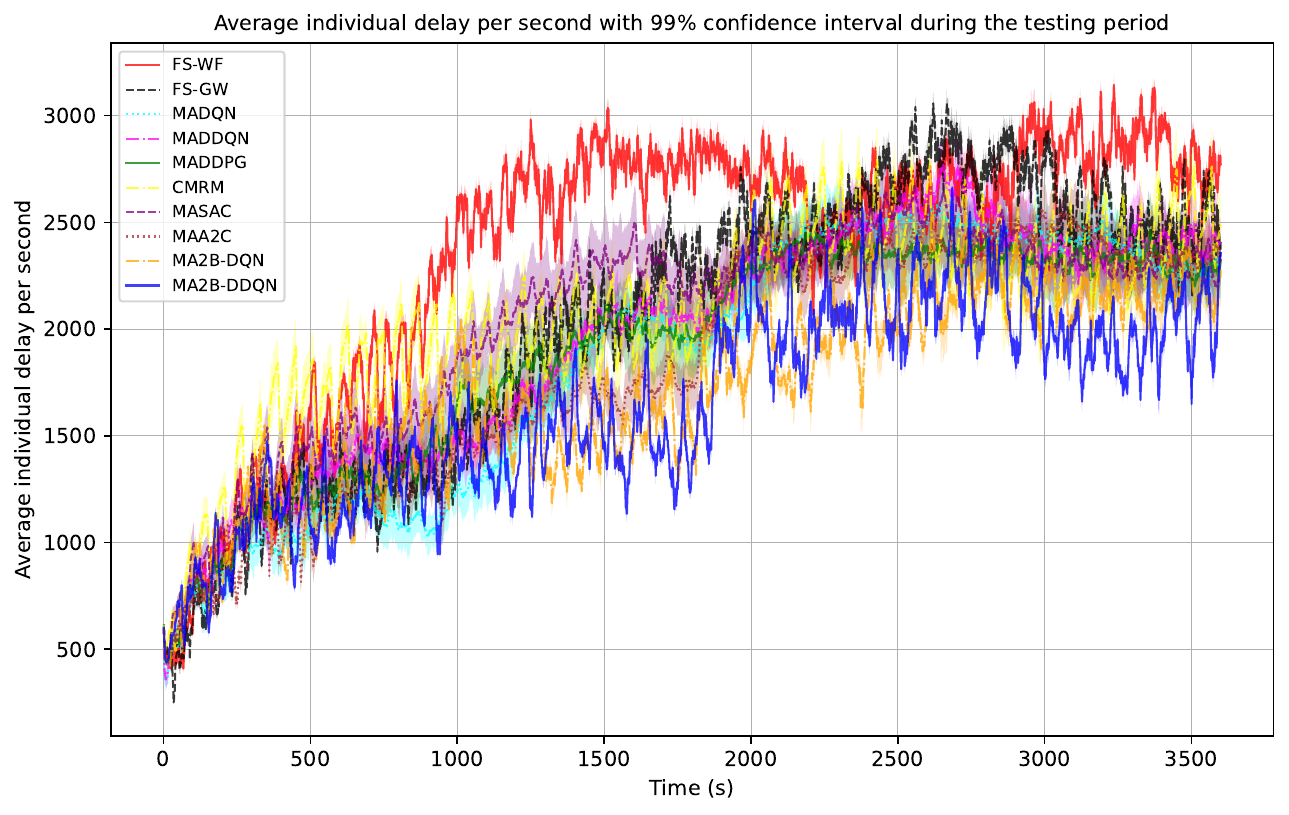}
\caption{Comparison of the individual delay and 99\% confidence interval throughout the testing period for scenario 6.}
\label{cost_comp_6}
\end{center}
\end{figure}

\begin{figure}[htbp]
\begin{center}
\subfloat[]{
    \includegraphics[width=0.85\textwidth]{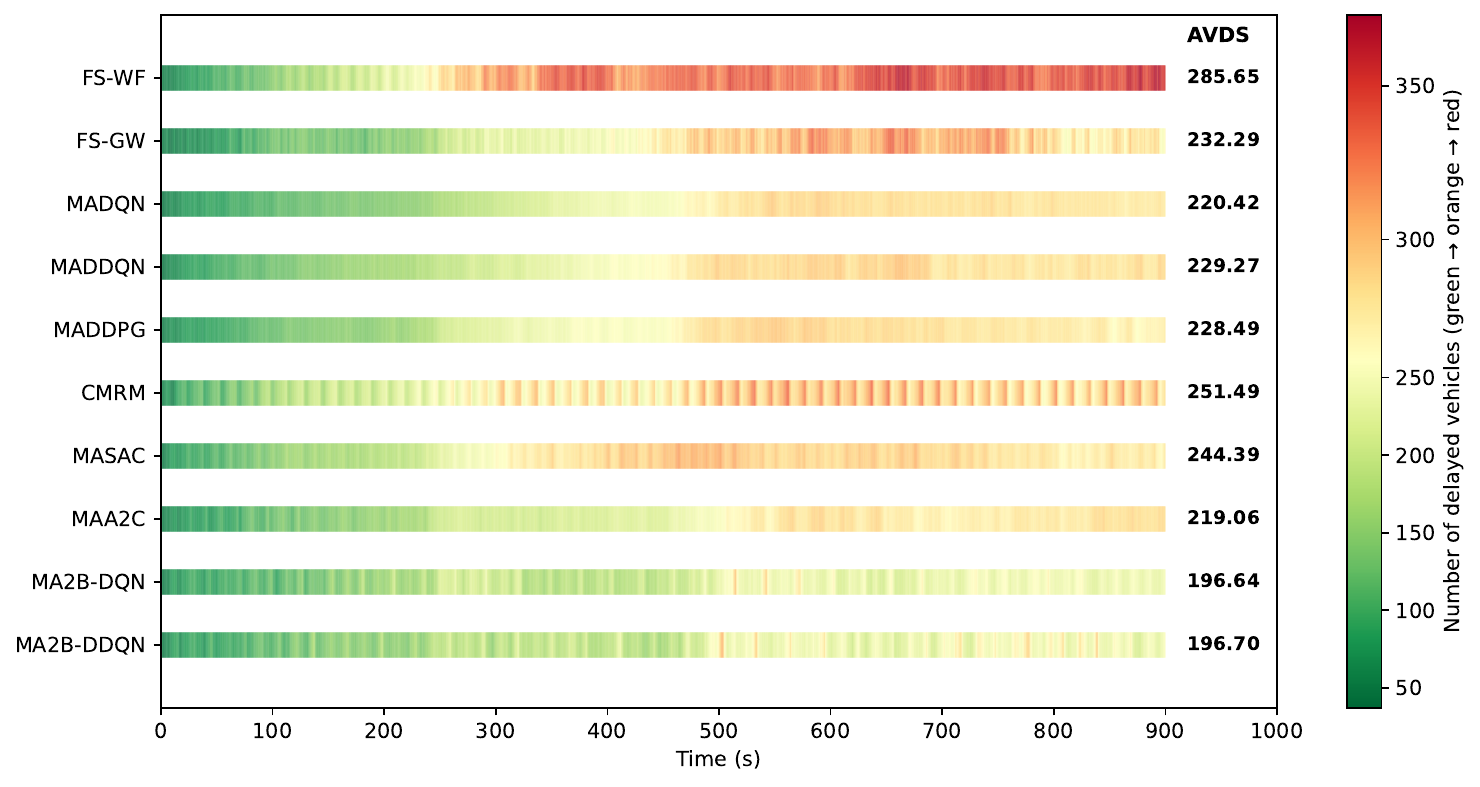}
    }
    
\subfloat[]{
    \includegraphics[width=0.85\textwidth]{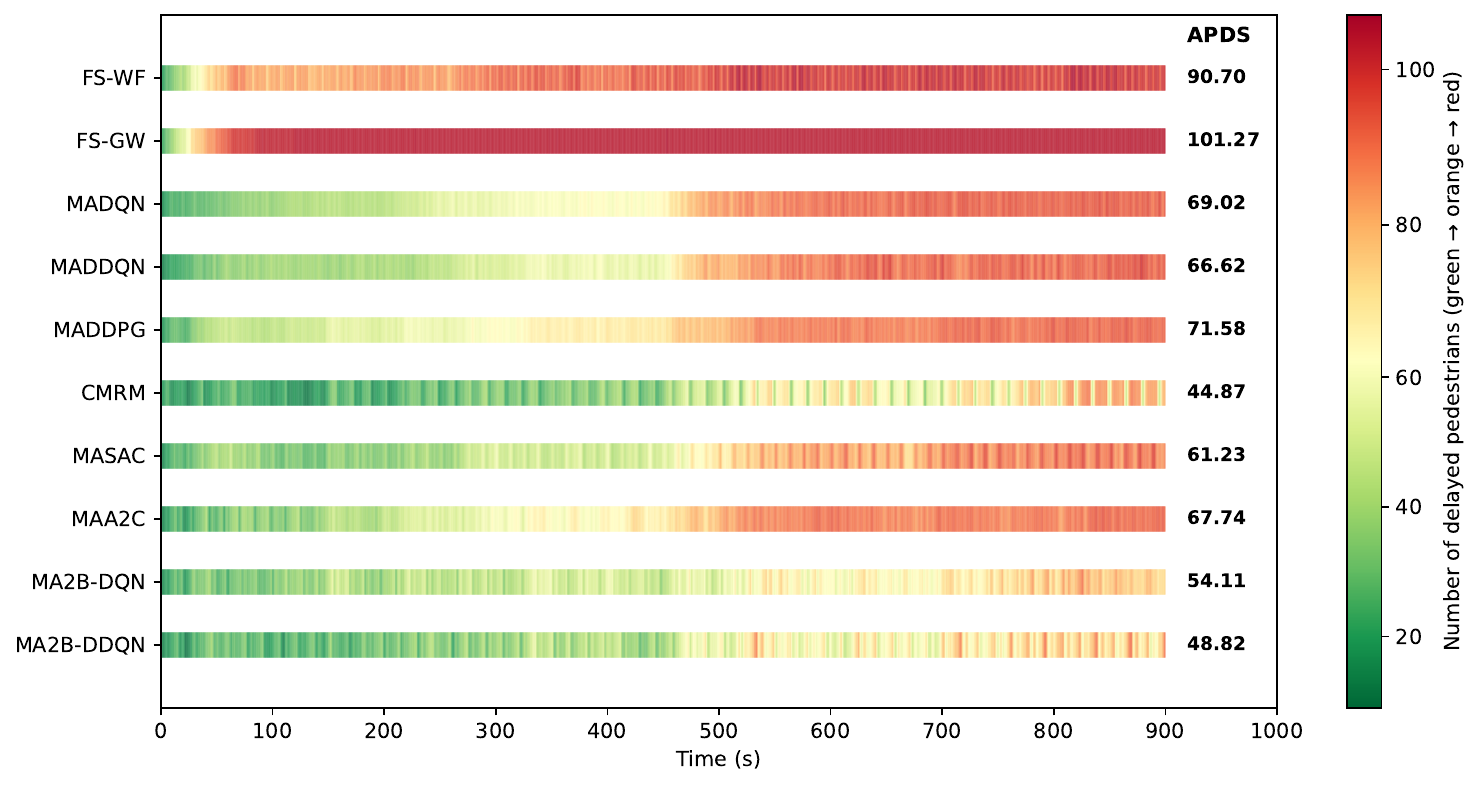}
    }
\caption{Analysis of different models in Scenario 6. (a) Vehicle delay over time (unit: vehicles per second). (b) Pedestrian delay over time (unit: pedestrians per second).}
\label{veh_ped_comp_6}
\end{center}
\end{figure}

This scenario simulates dynamically changing traffic demand over a one-hour testing period. During the first 30 minutes, traffic demand is moderate, but increases to heavy levels in the latter second half hour. This setup aims to test the ability of the method to adapt to rapidly changing traffic conditions. Table \ref{table_s6} shows the results for MA2B-DDQN and baseline methods, with MA2B-DDQN achieving the best performance with a PCD of 114.02. It surpasses MA2B-DQN by 3.23\%. The top-performing DRL baseline is MAA2C as well, with a comparatively high PCD of 126.97, and MA2B-DDQN surpasses it by 10.20\% (114.02 vs. 126.97 in terms of PCD ($\mu$)).
Considering average delay for both vehicles and pedestrians, MA2B-DDQN achieves consistently low values, highlighting its strong ability to manage traffic for both modalities. While CMRM attains the lowest AWTP (47.74 s), it performs poorly on AID, PCD, and ADV. The integration of the action branching architecture proves highly effective in multi-agent settings, as evidenced by the strong performance of both MA2B-DDQN and MA2B-DQN, which leverage this mechanism to outperform other DRL models and fixed-signal control methods.

Figure \ref{cost_comp_6} illustrates the individual delays of various methods throughout the testing period. FS-WF incurs the highest delay over the simulation period, indicating that such approaches are less suitable for this scenario due to limited coordination between intersections. In contrast, MA2B-DDQN consistently maintains very low delays throughout the entire period.

Figure \ref{veh_ped_comp_6} demonstrates the temporal distribution of vehicle and pedestrian delays throughout the entire testing period. MA2B-DDQN and MA2B-DQN achieve nearly identical AVDS of 196.70 and 196.64 vehicles per second, respectively, outperforming all other baseline methods. The best-performing baseline in this category, MAA2C, records an AVDS of 219.06 vehicles per second.
Regarding pedestrian delay, MA2B-DDQN achieves the second-lowest APDS at 48.82 pedestrians per second, outperforming MA2B-DQN (54.11 pedestrians per second) but trailing CMRM, which attains the lowest APDS of 44.87 pedestrians per second. These results reinforce the effectiveness of MA2B-DDQN in managing both vehicle and pedestrian traffic under dynamically changing conditions.

\subsection{Scenario 7}

\begin{table}[htbp]
\centering
\caption{Performance comparison for different models (scenario 7)}
\label{table_s7}
\begin{tabular}{p{2.7cm}p{2.5cm}p{2.0cm}p{1.5cm}p{1.5cm}p{1.5cm}p{1.5cm}}
\hline
model & AID ($\mu$) & AID ($\sigma$) & PCD ($\mu$) & PCD ($\sigma$) & ADV (s)& IMP \\
\hline
FS-WF & 3,321,165.85 & 86,559.28 & 211.40 & 5.51 & 268.56 & 66.54\%\\
FS-GW & 2,592,152.30 & 59,957.85 & 165.00 & 3.82 & 203.23 & 57.13\%\\
MADQN & 1,735,091.05 & 194,783.17 & 110.45 & 12.40 & 118.46 & 35.95\%\\
MADDQN & 1,597,786.40 & 169,355.70 & 101.71 & 10.78 & 107.18 & 30.45\%\\
MADDPG & 1,710,539.25 & 393,860.23 & 108.88 & 25.07 & 118.13 & 35.03\%\\
CMRM & 2,456,338.05 & 263,011.41 & 156.36 & 16.74 & 176.56 & 54.76\%\\
MASAC & 2,095,276.10 & 207,671.19 & 133.37 & 13.22 & 142.29 & 46.96\%\\
MAA2C & 1,381,279.75 & 90,859.14 & 87.92 & 5.78 & 92.1 & 19.55\%\\
MA2B-DQN & 1,203,887.35 & \textbf{21,982.32} & 76.63 & \textbf{1.40} & 78.26 & 7.69\%\\
\textbf{MA2B-DDQN} & \textbf{1,111,271.70} & 22,219.49 & \textbf{70.74} & 1.41 & \textbf{76.36} & \textit{NA}\\
\hline
\end{tabular}
\end{table}

\begin{figure}[htbp]
\begin{center}
\includegraphics[width=0.95\textwidth]{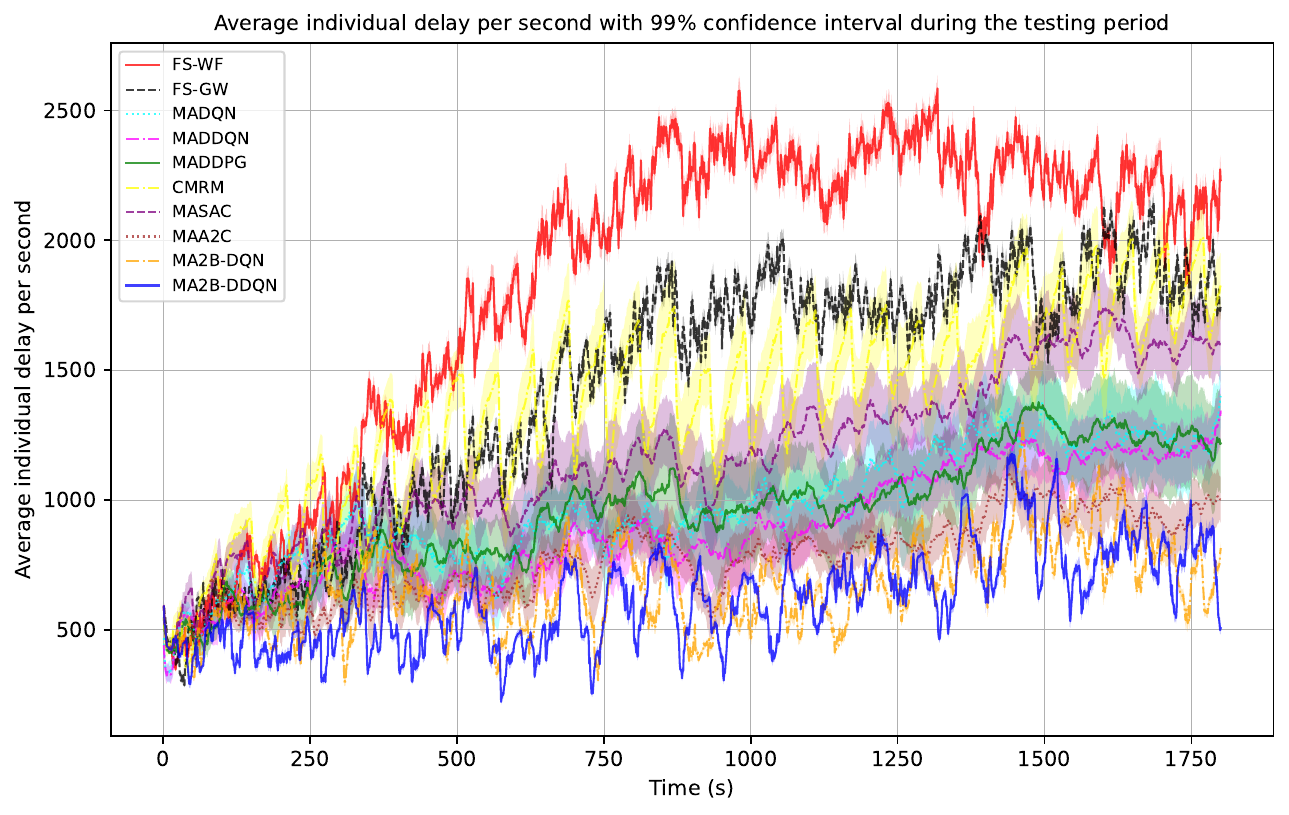}
\caption{Comparison of the individual delay and 99\% confidence interval throughout the testing period for scenario 7.}
\label{cost_comp_7}
\end{center}
\end{figure}

\begin{figure}[htbp]
\begin{center}
    \includegraphics[width=0.85\textwidth]{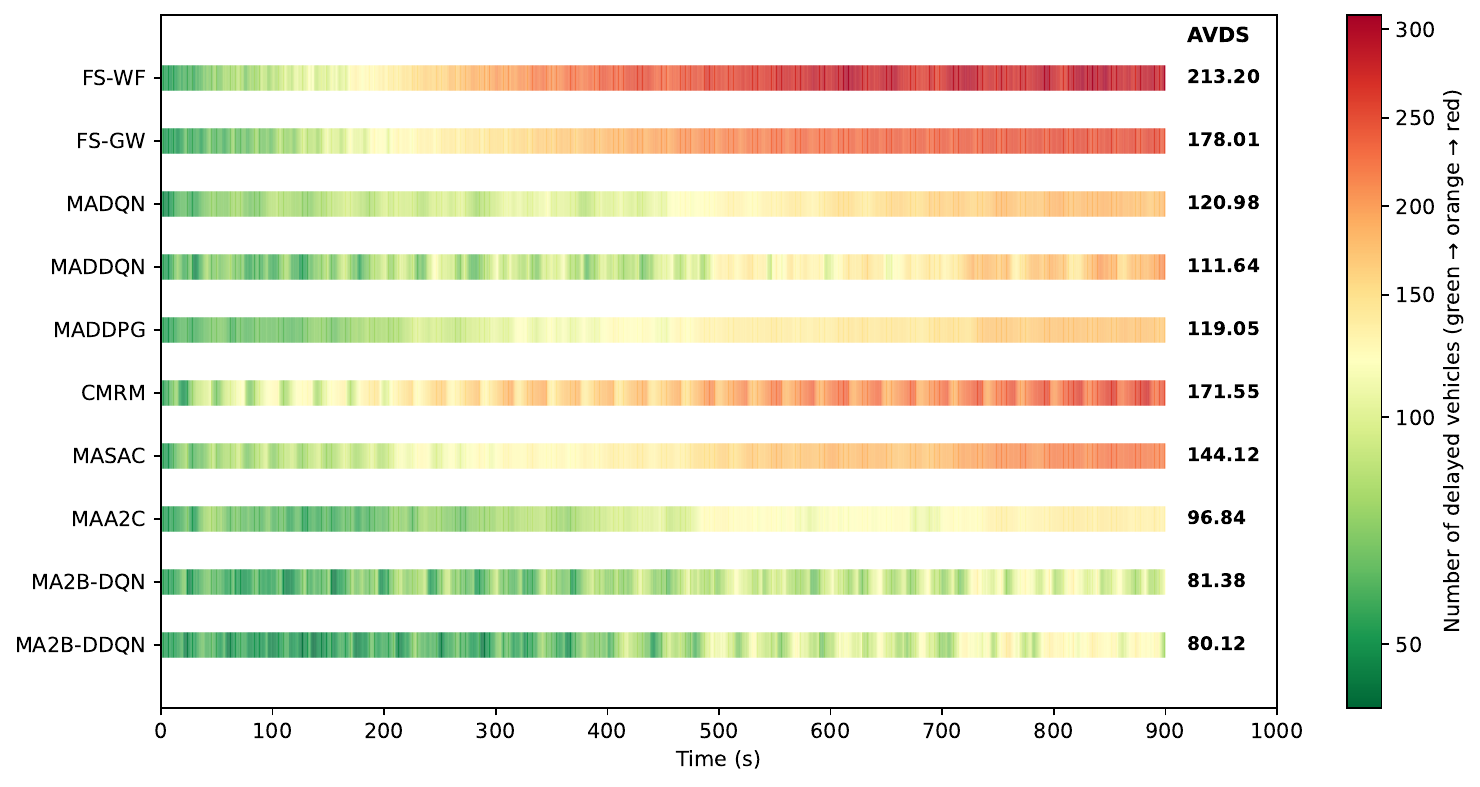}
\caption{Analysis of different models in Scenario 7. Vehicle delay over time (unit: vehicles per second).}
\label{veh_ped_comp_7}
\end{center}
\end{figure}

In this scenario, pedestrians are excluded from the entire corridor network to simulate intersections without pedestrian crossings. The results are shown in Table \ref{table_s7}. MA2B-DDQN achieves the best average performance, with a low AID of 1,111,271.70 individuals, a PCD of 70.74, and a low ADV of 76.36 seconds. Since pedestrians are not part of this scenario, AWTP is not applicable in Table \ref{table_s7}. The exclusion of pedestrians results in lower PCD compared with other scenarios. In a similar way, MA2B-DDQN outperforms all DRL and fixed signal baselines. However, MA2B-DQN shows greater robustness, evidenced by its lower standard deviations in AID and PCD, indicating that while MA2B-DDQN achieves higher average performance, MA2B-DQN is more robust. Specifically, MA2B-DDQN improves average performance by 7.69\% over MA2B-DQN. The best-performing DRL baseline is MAA2C, with a PCD of 87.92, which MA2B-DDQN significantly surpasses with an improvement (IMP) of 19.55\%.

Figure \ref{cost_comp_7} illustrates the individual delays of different methods over the testing period for this scenario. Similarly, FS-WF and FS-GW obtain the highest numbers over the simulation period, suggesting that these types of approaches are not well-suited for this problem. In contrast, MA2B-DDQN consistently maintains a very low number throughout the entire testing period.

Figure \ref{veh_ped_comp_7} presents the delay for vehicles over time. In this scenario, MA2B-DDQN and MA2B-DQN obtain similar AVDS, with 80.12 and 81.38 vehicles per second, respectively, where MA2B-DDQN exhibits a slight advantage. Both models significantly outperform the best baseline competitor, MAA2C, which records an AVDS of 96.84 vehicles per second.
This performance gap highlights the effectiveness of the action branching architecture and the DQN-based DRL approach in optimizing multimodal TSC. The results further confirm the superiority of MA2B-DDQN in efficiently managing vehicle flow under scenarios without pedestrian influence.

\subsection{Discussion}
\begin{figure}[htbp]
\begin{center}
\includegraphics[width=0.63\textwidth]{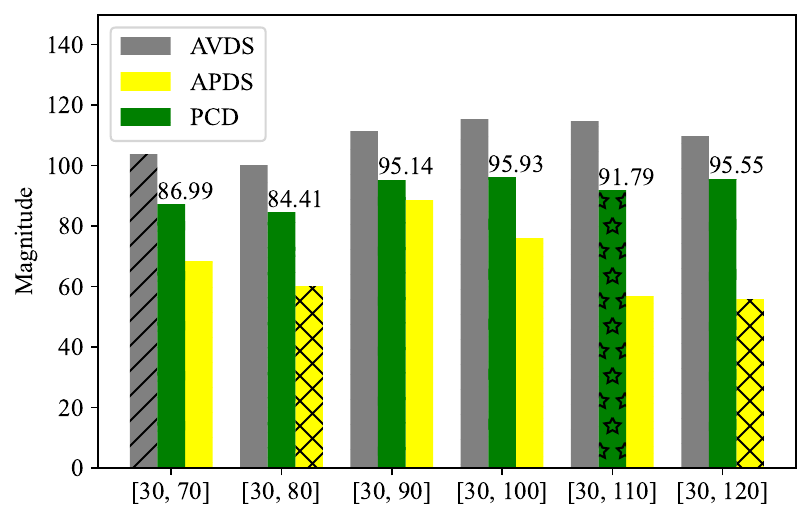}
\caption{Sensitivity analysis of MA2B-DDQN for the total duration parameter $C$ (Scenario 1).}
\label{sensitivity}
\end{center}
\end{figure}

The total duration parameter, $C$, is a critical factor that directly determines the green time allocated to each phase. MA2B-DDQN optimizes this value within a predefined range, currently set at [30, 70]. To understand the sensitivity of this parameter on overall TSC optimization performance, we conducted a sensitivity analysis across multiple duration ranges: [30, 70], [30, 80], [30, 90], [30, 100], [30, 110], and [30, 120]. Since the lower bound is fixed at 30 to comply with the minimum green time requirement, ensuring safety and preventing excessively short green phases, the analysis focuses on increasing the upper bound in 10-second increments, extending it from 70 to 120 seconds. Figure \ref{sensitivity} presents the results using PCD, AVDS, and APDS metrics for each range configuration. Overall, the performance of our method exhibits low sensitivity to variations in $C$, as fluctuations in results remain minimal with increasing upper bounds. Specifically, expanding the range from [30, 70] to [30, 80] leads to a slight reduction in PCD, from 86.99 to 84.41, respectively. However, beyond an upper bound of 90 (90/100/110/120), performance slightly deteriorates further, with PCD increasing to 95.14, 95.93, 91.79, and 95.55, respectively. AVDS follows a pattern similar to PCD, demonstrating limited variation across different ranges, with values spanning from 100.08 (at [30, 80]) to 115.23 (at [30, 100]). In contrast, APDS exhibits greater sensitivity, fluctuating more significantly between 55.70 pedestrians per second (at [30, 120]) and 88.57 pedestrians per second (at [30, 90]), indicating that pedestrian performance is more affected by changes in $C$ compared to vehicle flow.

Table \ref{effect} presents the results of an ablation study to evaluate the impact of the proposed action branching mechanism in MA2B-DDQN for TSC problems. The study compares three scenarios (Scenarios 1, 2, and 3), with their respective results detailed in Table \ref{effect}. The findings highlight that action branching plays a crucial role in enhancing the learning efficiency of MA2B-DDQN. It significantly improves overall individual-based traffic performance, as measured by AID and PCD, and enhances vehicle flow, as measured by AVDS and ADV, across all scenarios. Notably, the improvement in AID is substantial, ranging from 16.09\% in Scenario 1 to 46.78\% in Scenario 3. However, the mechanism does not yield better outcomes for pedestrian-related metrics, such as APDS and AWTP. Instead, it leads to an increase in both values across all three scenarios. This finding is likely due to the significantly higher number of vehicle users compared to pedestrians. Nevertheless, the observed discrepancy remains moderate, indicating a slight imbalance in prioritizing pedestrian flow relative to vehicle movement. For example, in Scenario 2, incorporating the action branching mechanism increases APDS from 60.95 to 74.37 pedestrians per second.

\begin{table}[htbp]
\centering
\caption{The effect of action branching mechanism in MA2B-DDQN}
\label{effect}
\begin{threeparttable}
\begin{tabular}{p{1.3cm}p{2.3cm}p{2.2cm}p{1.3cm}p{1.3cm}p{1.3cm}p{1.3cm}p{1.3cm}p{1.3cm}}
\hline
Scenario & Model & AID & PCD & AVDS & APDS & ADV & AWTP & IMP\\
\hline
\multirow{2}{*}{1} & W/o AB & 1,031,932.05 & 103.67 & 124.29 & \textbf{48.65} & 129.77 & \textbf{51.17} & 16.09\%\\
 & MA2B-DDQN & \textbf{865,874.70} & \textbf{86.99} & \textbf{103.83} & 68.36 & \textbf{98.37} & 77.23 & \textit{NA}\\
\hline
\multirow{2}{*}{2} & W/o AB & 3,952,766.60 & 143.75 & 234.77 & \textbf{60.95} & 280.77 & \textbf{54.87} & 29.86\%\\
 & MA2B-DDQN & \textbf{2,772,647.95} & \textbf{100.83} & \textbf{179.83} & 74.37 & \textbf{133.71} & 71.99 & \textit{NA}\\
\hline
\multirow{2}{*}{3} & W/o AB & 3,460,280.90 & 177.55 & 213.69 & \textbf{60.26} & 233.66 & \textbf{59.67} & 46.78\%\\
 & MA2B-DDQN & \textbf{1,841,687.10} & \textbf{94.50} & \textbf{109.88} & 72.99 & \textbf{111.44} & 78.06 & \textit{NA}\\
\hline
\end{tabular}
\textit{Notes:} W/o AB denotes MA2B-DDQN without action branching mechanism. Units for ADV and AWTP are in seconds.
\end{threeparttable}
\end{table}

Tables \ref{sensitivity_gamma} and \ref{sensitivity_lr} present the sensitivity analysis for the discount factor and the learning rate in MA2B-DDQN. Overall, $\gamma$ influences MA2B-DDQN’s performance, but the model is not highly sensitive within the tested range. As shown in Table \ref{sensitivity_gamma}, $\gamma=0.99$ yields the lowest mean AID (768,230.35) and PCD (77.18), while keeping variability relatively small ($\sigma_{\text{AID}}=43,420.72$, $\sigma_{\text{PCD}}=4.36$). By comparison, intermediate settings such as $\gamma=0.98$ lead to higher AID/PCD and noticeably larger standard deviations, indicating less stable learning. This suggests that a larger discount factor is preferable in Scenario 1 for better long-horizon credit assignment and more consistent outcomes.
For the learning rate, Table \ref{sensitivity_lr} indicates that MA2B-DDQN is moderately sensitive to $\text{lr}$, where overly small step sizes can hinder learning. In particular, $\text{lr}=1\times10^{-5}$ produces the largest AID (970,545.90) and PCD (97.50), together with high variability ($\sigma_{\text{AID}}=85,091.91$, $\sigma_{\text{PCD}}=8.55$), suggesting slow and unstable convergence. Increasing the learning rate to $\text{lr}=5\times10^{-4}$ yields the best overall performance, achieving the lowest mean AID (784,257.15) and a low mean PCD (78.79) with moderate deviations. Although $\text{lr}=5\times10^{-5}$ attains a comparable PCD (81.32), its AID remains higher and the results are less stable.

\begin{table}[htbp]
\centering
\caption{Sensitivity analysis for the discount factor $\gamma$ in MA2B-DDQN (Scenario 1)}
\label{sensitivity_gamma}
\begin{tabular}{p{1.3cm}p{2.0cm}p{2.0cm}p{1.6cm}p{1.6cm}p{1.6cm}p{1.6cm}}
\hline
$\gamma$ & AID ($\mu$) & AID ($\sigma$) & PCD ($\mu$) & PCD ($\sigma$) & ADV (s) & AWTP (s)\\
\hline
0.95 & 865,874.70 & 22,953.49 & 86.99 & 2.31 & 98.37 & 77.23\\
0.96 & 790,623.10 & 54,049.50 & 79.43 &	5.43 & 93.19 & 106.16\\
0.97 & 878,617.65 & 70,869.79 & 88.27 & 7.12 & 102.68 & 76.93\\
0.98 & 897,581.00 & 90,222.77 & 90.17 &	9.06 & 106.12 & 119.88\\
0.99 & 768,230.35 & 43,420.72 & 77.18 &	4.36 & 92.70 &	76.98\\
\hline
\end{tabular}
\end{table}

\begin{table}[htbp]
\centering
\caption{Sensitivity analysis for the learning rate in MA2B-DDQN (Scenario 1)}
\label{sensitivity_lr}
\begin{tabular}{p{1.3cm}p{2.0cm}p{2.0cm}p{1.6cm}p{1.6cm}p{1.6cm}p{1.6cm}}
\hline
lr & AID ($\mu$) & AID ($\sigma$) & PCD ($\mu$) & PCD ($\sigma$) & ADV (s) & AWTP (s)\\
\hline
0.00001 & 970,545.90 & 85,091.91 & 97.50	& 8.55 & 113.25 & 85.31\\
0.00005 & 809,431.30 & 58,826.08 & 81.32 & 5.91 & 96.53 & 73.48\\
0.0001 & 865,874.70 & 22,953.49 & 86.99 & 2.31 & 98.37 & 77.23\\
0.0005 & 784,257.15 & 51,821.93 & 78.79	& 5.21 & 94.12 & 78.68\\
0.001 & 849,217.70 & 69,156.02 & 85.31	& 6.95 &	100.80 & 49.01\\
\hline
\end{tabular}
\end{table}

\begin{figure}[htbp]
\begin{center}
\includegraphics[width=0.93\textwidth]{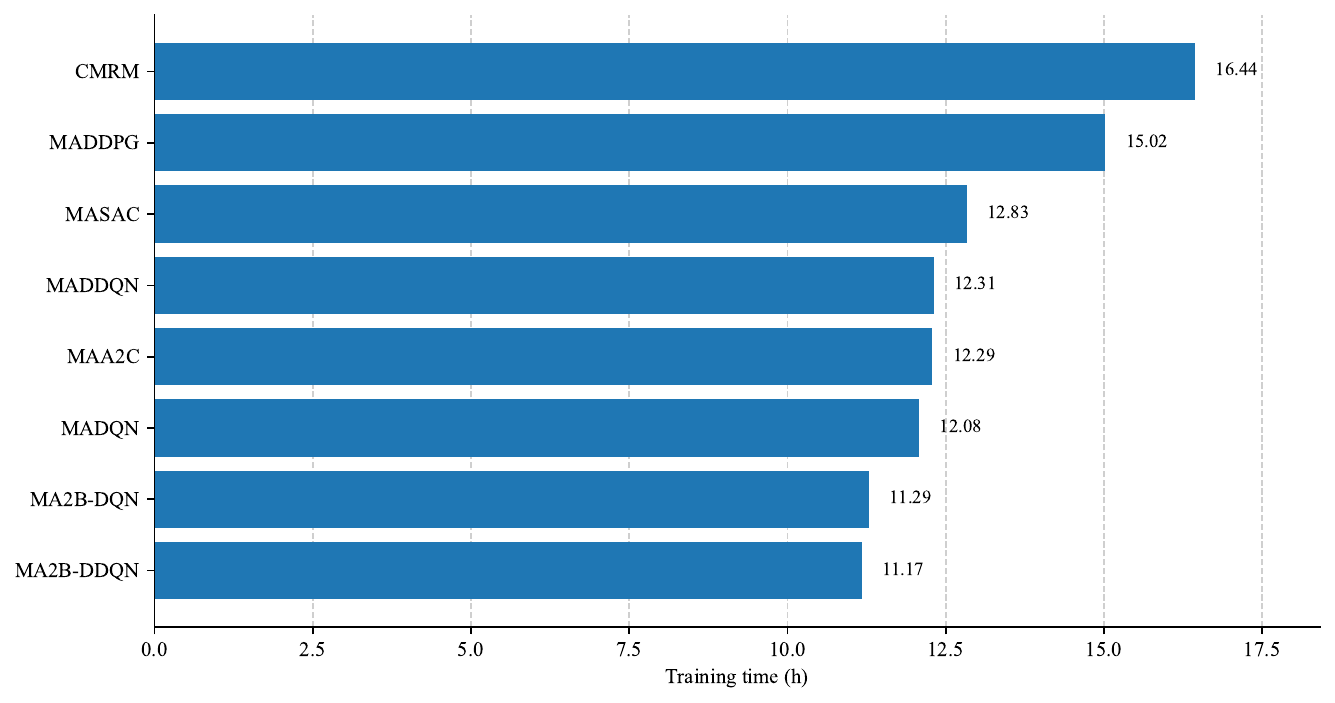}
\caption{Comparison of training time across different DRL models (Scenario 1).}
\label{computationaltime}
\end{center}
\end{figure}

Figure \ref{computationaltime} compares the training time of all evaluated models in Scenario 1 and shows that MA2B-based methods are among the most computationally efficient. In particular, MA2B-DDQN (11.17 h) and MA2B-DQN (11.29 h) achieve the shortest training times, followed by a middle cluster including MADQN (12.08 h), MAA2C (12.29 h), MADDQN (12.31 h), and MASAC (12.83 h). By contrast, MADDPG (15.02 h) and CMRM (16.44 h) require substantially longer training, indicating a higher computational burden. Overall, the results suggest that MA2B-DDQN not only delivers strong performance, but also offers a practical advantage in training efficiency compared with heavier baselines such as MADDPG and CMRM.

Based on the analysis of performance metrics and outcomes across the seven testing scenarios described above, several key insights can inform and improve future research or practical applications. These findings are summarized as follows:
\begin{itemize}
\item Overall, applying DRL for adaptive traffic signal control typically outperforms fixed-signal methods that rely solely on traffic volume. This advantage stems from the fact that traffic volume can vary significantly, making fixed-signal approaches less scalable and effective. Additionally, fixed-signal methods focus primarily on vehicle demand, often leading to suboptimal performance for pedestrians.

\item For managing multiple intersections within a corridor network, signal coordination between intersections generally results in better performance than modeling each intersection independently. This is supported by the superior performance of green wave control, which enables coordination, compared to agents managing signals independently without coordination.

\item In multi-agent TSC problems, it remains unclear whether continuous or discrete action spaces are more suitable for modeling, as baseline models using both types of action spaces yield very similar performance in certain scenarios. For instance, MADDQN and MADDPG perform comparably in scenarios 4, 5, and 7.

\item Incorporating the proposed action branching architecture for this problem can greatly improve the performance of DRL methods when operating within a discrete action space. This is evidenced by the results of MA2B-DDQN and MA2B-DQN, both of which achieve significantly better performance than other DRL methods across most testing scenarios.

\item DRL methods like MADDPG, MADQN, and MADDQN may be less suitable for traffic signal control in multi-agent environments. MADDPG appears sensitive in multi-agent TSC optimization, exhibiting high variance in performance. Additionally, both MADQN and MADDQN struggle to handle the high-dimensional and complex action spaces inherent to multi-agent environments.

\item MAA2C is a strong baseline for benchmarking in traffic signal control, consistently ranking as the best-performing DRL baseline. This may be due to the advantage function in A2C, which guides the algorithm to focus training on actions that significantly enhance performance \citep{heo2024optimal}.

\item When the demand of the corridor network approaches the saturation flow, it will limit the effectiveness of different optimization methods, including advanced DRL approaches. The effectiveness of TSC optimization methods can plateau. This is probably because the physical constraints of the intersection (like the number of lanes and the maximum throughput they can handle) become the limiting factors, rather than the efficiency of the signal timings themselves.
\end{itemize}

\section{Conclusions and future work}
\label{conclusion}
This paper proposes a novel cooperative multi-agent action branching DRL framework for traffic signal control at the medium-scale corridor network level. This intelligent traffic signal control system prioritizes fairness and equality by treating all users across the transportation network equally. First, it constructs a discrete action space that spans the next two phases, integrating both adaptive intervals and green time distributions. Second, an action branching architecture with both local and global actions is introduced, enabling effective modeling within a discrete action space and multi-agent control context. Third, a reward function is designed to ensure fairness and equality across multimodal transportation. Finally, a double deep Q-network is incorporated to address Q-value overestimation within this DRL framework. The proposed DRL method, termed MA2B-DDQN, is comprehensively tested through experiments across seven typical traffic scenarios on a real-world corridor network in Melbourne, demonstrating superior performance in multi-agent traffic signal control. MA2B-DDQN consistently outperforms baseline methods, achieving at least 2.75\% improvement in near-saturated traffic scenarios and at least 10.20\% in the remaining scenarios in terms of overall delay. This includes outperforming fixed signals based on Webster’s Formula, green wave control, and advanced DRL models such as MADQN, MADDQN, MADDPG, CMRM, MASAC, and MAA2C. Additionally, MA2B-DDQN is confirmed to be the most robust and stable method, demonstrating the lowest variance across different testing conditions. Comprehensive comparisons further show that incorporating double deep Q-network within the MA2B-DDQN framework yields significant performance improvements, with gains ranging from 2.87\% to 16.40\%. In addition, the ablation study confirms that the proposed action-branching mechanism is a key driver of MA2B-DDQN’s gains, improving learning efficiency and traffic performance across scenarios.

The proposed MA2B-DDQN offers a scalable learning-based controller for multi-intersection and multimodal traffic signal optimization. Beyond performance gains, the outcomes of this study support practical applications such as digital-twin/simulation-based benchmarking and offline validation under varying demand conditions, with repeated-seed testing providing an indicator of performance stability.

In future work, we will enhance MA2B-DDQN in two directions. 
First, we will incorporate safety and environmental objectives (e.g., safety surrogates and emissions proxies) into a multi-objective or constrained RL formulation. Second, we will consider dynamic priorities for emergency vehicles by incorporating pre-emption and priority requests into the state and action constraints, enabling the controller to create green waves or temporary phase extensions while maintaining fairness impacts on general traffic within bounded limits.

\section*{CRediT authorship contribution statement}
\textbf{Xiaocai Zhang: }Conceptualization, Data curation, Methodology, Software, Writing - original draft.
\textbf{Neema Nassir: }Conceptualization, Methodology, Project administration, Writing - review \& editing.
\textbf{Lok Sang Chan: }Data curation, Methodology, Software, Writing - review \& editing.
\textbf{Milad Haghani: }Methodology, Writing - review \& editing.

\section*{Declaration of competing interest}
The authors declare that they have no known competing financial interests or personal relationships that could have appeared to influence the work reported in this paper.

\section*{Acknowledgment}
This work was funded by ARC LP200301389, Kapsch TrafficCom Australia, RACQ, and iMOVE CRC, the Cooperative Research Centres program, an Australian Government initiative. 

\bibliography{mybibfile}

\end{document}